\documentclass[preprints,article,accept,pdftex,moreauthors]{Definitions/mdpi} 
\firstpage{1} 
\pubvolume{1}
\issuenum{1}
\articlenumber{0}
\pubyear{2025}
\copyrightyear{2025}
\datereceived{ } 
\daterevised{ } % Comment out if no revised date
\dateaccepted{ } 
\datepublished{ } 
\hreflink{https://doi.org/} % If needed use \linebreak
\pdfoutput=1 % Uncommented for upload to arXiv.org
\Title{Few-Shot Optimization for Sensor Data Using Large Language Models: A Case Study on Fatigue Detection}

\TitleCitation{Title}

\Author{Elsen Ronando$^{1,2,\dagger,\ddagger,*}$\orcidA{} and Sozo Inoue $^{1,\ddagger}$\orcidB{}}

\AuthorNames{Elsen Ronando and Sozo Inoue}

\isAPAStyle{%
       \AuthorCitation{Ronando, E. \& Inoue, S.}
         }{%
        \isChicagoStyle{%
        \AuthorCitation{Ronando, Elsen and Sozo Inoue.}
        }{
        \AuthorCitation{Ronando, E.; Inoue, S.}
        }
}

\address{%
$^{1}$ \quad Graduate School of Life Science and Systems Engineering, Kyushu Institute of Technology, 2-4 Hibikino, Wakamatsu Ward, Kitakyushu 808-0135, Japan\\
$^{2}$ \quad Department of Informatics, Universitas 17 Agustus 1945 Surabaya, Semolowaru No. 45, Kota Surabaya 60118, Indonesia}

\corres{Correspondence: \href{mailto:ronando.elsen840@mail.kyutech.jp}{ronando.elsen840@mail.kyutech.jp}}

\firstnote{Current address: Graduate School of Life Science and Systems Engineering, Kyushu Institute of Technology, 2-4 Hibikino, Wakamatsu Ward, Kitakyushu 808-0135, Japan}  % Current address should not be the same as any items in the Affiliation section.
\secondnote{These authors contributed equally to this work.}
\abstract{In this paper, we propose a novel few-shot optimization with HED-LM (Hybrid Euclidean Distance with Large Language Models) to improve example selection for sensor-based classification tasks. While few-shot prompting enables efficient inference with limited labeled data, its performance largely depends on the quality of selected examples. HED-LM addresses this challenge through a hybrid selection pipeline that filters candidate examples based on Euclidean distance and re-ranks them using contextual relevance scored by large language models (LLMs). To validate its effectiveness, we apply HED-LM to a fatigue detection task using accelerometer data characterized by overlapping patterns and high inter-subject variability. Unlike simpler tasks such as activity recognition, fatigue detection demands more nuanced example selection due to subtle differences in physiological signals. Our experiments show that HED-LM achieves a mean macro F1-score of 69.13$\pm$10.71\%, outperforming both random selection (59.30$\pm$10.13\%) and distance-only filtering (67.61$\pm$11.39\%). These represent relative improvements of 16.6\% and 2.3\%, respectively. The results confirm that combining numerical similarity with contextual relevance improves the robustness of few-shot prompting. Overall, HED-LM offers a practical solution to improve performance in real-world sensor-based learning tasks and shows potential for broader applications in healthcare monitoring, human activity recognition, and industrial safety scenarios.}


\begin{document}

%%%%%%%%%%%%%%%%%%%%%%%%%%%%%%%%%%%%%%%%%%
%\setcounter{section}{-1} %% Remove this when starting to work on the template.
%\section{How to Use this Template}

%The template details the sections that can be used in a manuscript. Note that the order and names of article sections may differ from the requirements of the journal (e.g., the positioning of the Materials and Methods section). Please check the instructions on the authors' page of the journal to verify the correct order and names. For any questions, please contact the editorial office of the journal or support@mdpi.com. For LaTeX-related questions please contact latex@mdpi.com.%\endnote{This is an endnote.} % To use endnotes, please un-comment \printendnotes below (before References). Only journal Laws uses \footnote.

% The order of the section titles is different for some journals. Please refer to the "Instructions for Authors” on the journal homepage.

\section{Introduction}
Few-shot prompting with large language models (LLMs) has attracted considerable interest as a groundbreaking method for tackling new tasks with limited training data. In contrast to conventional machine learning models, which typically depend on vast datasets for training, few-shot prompting enables LLMs to adapt to new tasks by leveraging only a small number of in-context examples. For example, GPT-3 has exhibited exceptional abilities in producing contextually appropriate and precise responses based on just a handful of instance prompts, emphasizing few-shot prompting's potential to lessen the resource requirements of extensive training \citep{brown2020languagemodelsfewshotlearners}. Key attributes of this approach include its flexibility across diverse tasks \citep{izacard2022atlasfewshotlearningretrieval}, its multilingual processing capabilities \citep{Lin2021FewshotLW}, and its superior reasoning performance in complex scenarios \citep{cao2021conceptlearnersfewshotlearning}. However, these advantages are counterbalanced by challenges such as limited generalization to unfamiliar domains \citep{Qin2023Bi-Level}, substantial computational resource requirements \citep{izacard2022atlasfewshotlearningretrieval}, and a pronounced dependence on effective prompt design \citep{gao2021makingpretrainedlanguagemodels, sclar2024quantifying}.

The effectiveness of few-shot prompting largely hinges on selecting examples incorporated into the prompt. Studies have shown that using inappropriate examples, whether randomly selected or overly specific, can impede the model's ability to generalize across tasks \citep{adiga-etal-2024-designing}. Moreover, thoughtful example selection is critical for optimizing resource utilization and reducing computational overhead while maintaining high performance \citep{9996364}. Notably, prompts that integrate examples closely aligned with the target task have been observed to enhance model accuracy by as much as 30\% compared to prompts with generic or mismatched examples \citep{gao2021makingpretrainedlanguagemodels}. These findings underscore the need for robust strategies in example selection to mitigate biases and improve generalization, especially in applications where data variability and complexity are high.

\begin{figure}[ht]
\begin{center}
 \includegraphics[width=1\textwidth]{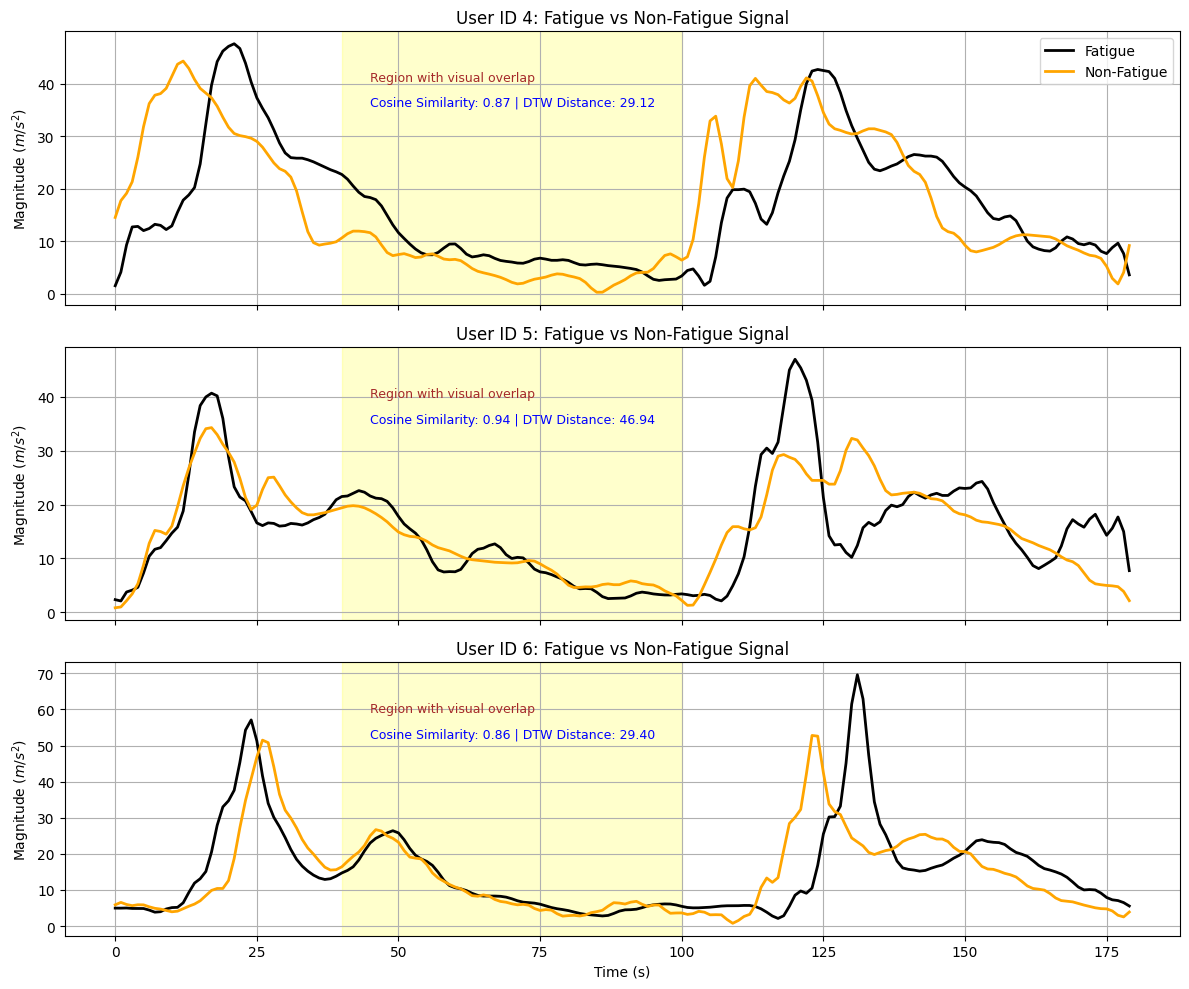}
\end{center}
\vspace{-1em}
\caption{\textbf{Comparison of fatigue and non-fatigue accelerometer signals from three different users}. Each subplot shows one representative sample per class, extracted from running sessions. The yellow-highlighted region (time index 40–100) indicates a temporal window where the waveform patterns between fatigue and non-fatigue signals exhibit visual similarity. To quantitatively support this observation, we report Cosine Similarity and Dynamic Time Warping (DTW) distance between the two signals. Cosine Similarity captures directional similarity (1.0 = identical), while DTW accounts for temporal misalignment (lower values = more similar). These metrics further emphasize the challenge of signal overlap, motivating the need for context-aware example selection in few-shot prompting.}
\label{fig:chal}
\vspace{-2mm}
\end{figure}

Real-world applications of few-shot prompting, particularly in domains involving sensor data, present additional complexities due to the inherent variability and noise in the data. Sensor data, such as accelerometer readings, often exhibit intricate patterns that are challenging to represent effectively in a few-shot prompting framework, as illustrated in Fig.~\ref{fig:chal}. Misrepresenting such data can significantly reduce task performance, emphasizing the importance of selecting representative and contextually relevant examples for the target task. For instance, \citet{feng2019fewshotlearningbasedhumanactivity} proposed a parameter transfer method, for example, selection in human activity detection using sensor data, achieving a 15\% improvement in generalization. However, their approach was highly sensitive to data quality, with noisy inputs causing a performance drop of over 20\%. Similarly, \citet{weko_239518_1} utilized a randomized sample selection method, transforming accelerometer data into graph representations for few-shot prompting. While this method improved task-specific performance by 22\%, it lacked robustness in handling complex datasets compared to zero-shot learning setups.

These limitations highlight the necessity for structured and practical strategies, such as selection for few-shot prompting, mainly when dealing with high-variability and high-dimensional sensor data. Proximity-based measures have been explored as a potential solution, evaluating the similarity between examples and the target task. Two primary approaches are commonly considered: leveraging the internal reasoning capabilities of LLMs or employing distance-based metrics such as Euclidean distance. While LLMs excel in contextual evaluations, they struggle with numerical data, such as sensor readings, due to their text-centric architectures \citep{weko_239518_1, fan-etal-2024-ctyun, Spathis2023The, Hota_2024, li2024sensorllmaligninglargelanguage}. Conversely, distance-based methods provide straightforward numerical similarity assessments but fail to capture the contextual depth required for effective prompting.

Despite various approaches to few-shot prompting with sensor data, existing methods remain limited in adapting to complex and noisy real-world signals. Traditional machine learning models, such as Random Forests, rely on handcrafted features and require extensive labeled data for training. These models often struggle with intra-subject variability and overlapping patterns commonly found in fatigue detection scenarios \citep{10.1145/3594806.3594825}. While computationally simple, random selection frequently introduces irrelevant or misleading examples, especially when data complexity is high \citep{zhao2021calibrateuseimprovingfewshot}. Distance-based approaches provide more structured numeric similarity but suffer from semantic ambiguity; examples may be numerically close yet contextually inappropriate, resulting in misclassification \citep{1360862102144993152}. These shortcomings demonstrate the need for a hybrid selection mechanism combining quantitative precision and semantic relevance to support more robust and adaptable few-shot prompting.

As further illustrated in Fig.~\ref{fig:chal}, accelerometer signals from different users often exhibit visually overlapping patterns between fatigue and non-fatigue classes. Even when the true labels differ, these signals' temporal structures and magnitudes may appear strikingly similar. This phenomenon exacerbates the abovementioned limitations, making it difficult for random sampling and distance-based methods to select representative examples reliably. These visual overlaps highlight the practical need for a more context-aware selection strategy to address semantic and numeric ambiguity in few-shot prompting tasks.

We adopt fatigue detection from accelerometer data as our case study to provide a rigorous benchmark for evaluating example selection strategies under real-world signal ambiguity. Fatigue detection is characterized by high intra-class variability, overlapping signal patterns, and subtle class boundaries, making it a suitable yet challenging scenario for evaluating the robustness of few-shot prompting. However, we emphasize that improving fatigue detection is not this research's primary goal. Rather, fatigue detection serves as a representative task to assess the adaptability and effectiveness of our proposed selection framework in realistic sensor-based learning environments.

To overcome these challenges, we propose the Hybrid Euclidean Distance with Large Language Models (HED-LM) framework, which integrates numerical similarity filtering with LLM-based contextual evaluation to optimize example selection in sensor-based few-shot prompting. Our approach systematically integrates multiple stages to enhance the selection process. First, it employs numerical preprocessing to transform raw sensor signals into structured feature representations, ensuring that relevant features are retained while reducing noise. Next, Euclidean distance filtering is applied to identify the most numerically similar examples to the target task, serving as an initial selection mechanism. However, recognizing that numerical similarity alone is insufficient, HED-LM then incorporates LLM-based contextual relevance scoring, which refines the selected examples by evaluating their domain-specific semantic alignment. Finally, an optimized few-shot prompting strategy is applied to construct an effective in-context learning setup, ensuring that the chosen examples contribute to better model generalization.

This study aims to address the following research questions (RQs):
\begin{itemize}
    \item \textbf{RQ1:} How can the performance of few-shot prompting be improved for sensor-based classification tasks?\\
    \textbf{Contribution 1:} We explore the impact of optimized example selection strategies in few-shot prompting using wearable sensor data, highlighting challenges in generalization, variability, and signal ambiguity with case study on fatigue detection.
    \item \textbf{RQ2:} How can the proposed HED-LM framework enhance example selection in few-shot prompting for sensor data?\\
    \textbf{Contribution 2:} We introduce HED-LM, a novel hybrid framework that integrates Euclidean distance filtering and LLM-based contextual scoring to select semantically and numerically relevant examples.
    \item \textbf{RQ3:} How does HED-LM compare to conventional example selection approaches such as random sampling and distance-based filtering in few-shot prompting?\\
    \textbf{Contribution 3:} We show that HED-LM improves few-shot prompting performance over conventional methods. Compared to random and distance-based selection, HED-LM achieves relative macro F1-score improvements of 16.6\% and 2.3\%, respectively, demonstrating the advantage of combining numerical similarity and contextual relevance in selecting examples.
\end{itemize}

This study introduces a novel strategy for few-shot prompting that leverages numerical proximity and domain-aware contextual reasoning using large language models (LLMs). While fatigue detection is adopted as a representative case due to its inherent signal ambiguity, the central contribution lies in developing a generalizable selection mechanism, Hybrid Euclidean Distance with LLM (HED-LM), that integrates distance-based filtering and LLM-driven relevance scoring. This dual-selection strategy addresses core limitations in existing few-shot approaches for sensor data, offering a scalable and adaptable prompting framework applicable beyond the fatigue domain, including other physiological, safety-critical, or activity recognition scenarios.

The remainder of this paper is structured as follows: Section~\ref{related_work} reviews existing literature, identifying gaps this study seeks to address. Section~\ref{methods} details the methodology and the HED-LM framework. Section~\ref{experiments} presents experimental results and discusses their implications. Section~\ref{discussion} explores the broader impacts and limitations of this research. Finally, Section~\ref{conclusion} concludes with insights and potential directions for future work.

%%%%%%%%%%%%%%%%%%%%%%%%%%%%%%%%%%%%%%%%%%
\section{Related Work}
\label{related_work}
Optimizing example selection in few-shot prompting remains a critical challenge, particularly in sensor-based data applications. This section reviews the literature on few-shot optimization and its broader application to sensor data, highlighting key challenges and opportunities. We employ physical fatigue detection as a representative test case to systematically evaluate these challenges. However, it is important to note that fatigue detection is not the primary goal of this research; instead, it serves as a benchmark to validate the effectiveness of HED-LM in optimizing example selection for high-variability sensor data. The insights gained from this case study are expected to generalize to other sensor-based prompting applications.

\subsection{Few-Shot Optimization}
\label{subsec:rel-1}
Few-shot prompting is a method that allows large language models to generalize from a small set of labeled examples by leveraging in-context learning without additional training. This approach has gained increasing attention due to its efficiency across a wide range of domains, including action recognition \citep{shi2022knowledgepromptingfewshotaction}, vision-language tasks \citep{jin2022goodpromptworthmillions}, and text classification \citep{min2022noisychannellanguagemodel}. 

Although prompting-based models are highly flexible, their performance is heavily influenced by the quality of examples selected for inclusion in the prompt, especially when dealing with complex or specialized data. Several strategies have been proposed to optimize this process. \citet{aguirre2023selectingshotsdemographicfairness} emphasized the importance of contextual evaluation in improving model performance, while \citet{cegin2024userandomselectionnow} examined the trade-offs between random and structured selection. Although the random selection is computationally efficient and does not require additional heuristics, \citet{weko_239518_1} demonstrated that it often leads to significant performance degradation, mainly when applied to high-dimensional sensor data.

Alternative approaches have sought to impose more structure on example selection. \citet{Perez2021TrueFL} introduced a method combining cross-validation with a minimum description length criterion but found that it performed inconsistently across tasks. Similarly, \citet{Chang2021OnTI} employed a K-means clustering approach to improve selection, yet this method relied on unlabeled data, limiting its applicability. A feature distribution analysis technique was proposed by \citet{9996364}, which identified informative samples based on data distribution properties, though it required extensive labeling and lacked adaptability across different prompting tasks.

Dynamic and similarity-based approaches have also been explored. \citet{margatina2023activelearningprinciplesincontext} proposed an active learning-based approach that effectively identified high-value examples but was limited by its single-iteration process, missing opportunities for iterative refinement. Similarly, \citet{yao2024samplespromptsexploringeffective} introduced in-context sampling (ICS), which aligned prompts based on data similarity but required substantial domain expertise. Methods like ACSESS \citep{pecher2024automaticcombinationsampleselection} and IDS \citep{qin-etal-2024-context} provided more consistent improvements but faced scalability and computational constraints. Skill-KNN \citep{an-etal-2023-skill} further emphasized the difficulties in balancing domain generalization with robust example selection.

While these studies have made significant progress in example selection, there remains a persistent gap when applying these strategies to sensor-based few-shot prompting. This is especially critical in domains such as accelerometer analysis, where data variability, noise, and high dimensionality complicate similarity evaluation and example construction.

\subsection{Few-shot Prompting on Sensor Data}
\label{subsec:rel-2}
Applying few-shot prompting to sensor data presents additional complexities due to the fundamental differences between structured numerical signals and the text-centric architectures of large language models (LLMs). Unlike textual inputs, sensor readings are high-dimensional, noise-prone, and exhibit significant temporal dependencies, making them difficult to interpret using few-shot prompting frameworks for natural language tasks.

\subsubsection{General Challenges in Sensor Data for Few-shot Prompting}
\label{subsubsec:rel-1}
Sensor data inherently possess high dimensionality, significant variability, and noise. These complexities pose unique challenges, such as selection for few-shot prompting. The high dimensionality and temporal dependencies in sensor signals complicate the representation and interpretation process, leading to potential misclassification if examples are not carefully selected. Additionally, noise and variability, common in real-world sensor data, further exacerbate the challenge of ensuring the representativeness and relevance of selected examples for accurate prompting.

Numerous research efforts have sought to tackle these difficulties. \citet{liu2023largelanguagemodelsfewshot} integrated LLMs with physiological and behavioral time-series data but reported information loss when converting numerical signals into text representations. Similarly, \citet{li2024sensorllmaligninglargelanguage} explored LLMs applications in human activity recognition but faced difficulties capturing subtle variations in sensor readings, which were poorly represented in text-based models. Advanced preprocessing pipelines, such as those developed by \citet{vibration6040059}, have been introduced to handle noisy sensor inputs. However, their high computational requirements have made real-time implementation impractical.

\subsubsection{Fatigue Detection as a Representative Case Study}
\label{subsubsec:rel-2}
Fatigue detection was chosen as a representative case study due to its practical importance and inherent complexity. Accurately detecting fatigue has substantial implications for health, safety, and productivity across numerous healthcare and occupational safety domains. Moreover, accelerometer-based fatigue detection is characterized by subtle and overlapping sensor patterns, which make example selection critically important \citep{weko_239518_1, li2024sensorllmaligninglargelanguage}. The subtle differences between fatigue and non-fatigue states, coupled with temporal dynamics and sensor noise, provide an ideal scenario to rigorously test and validate the effectiveness of example selection optimization methods like HED-LM.

A follow-up study \citep{weko_239518_1} evaluated fatigue detection using accelerometer data as a case study for few-shot prompting, comparing list-based and graph-based representations. Their results indicated that zero-shot learning often outperformed few-shot prompting when random example selection was used, highlighting a significant limitation of current selection strategies. This underscores the need for a more structured and informed approach to example selection, particularly in sensor-based few-shot prompting applications.

\subsection{Challenges and Motivations}
\label{subsec:rel-3}
Although the approaches discussed in the previous literature have significantly contributed to developing few-shot prompting, critical challenges still have not been fully resolved, especially in complex and highly variable sensor data. Firstly, example selection remains a weak point that significantly affects model performance under few-shot conditions. Random selection methods often produce irrelevant or even misleading examples, especially when the variability in the data is high. In contrast, numerical distance-based approaches (such as Euclidean distance) can provide clear similarity metrics but often neglect crucial contextual aspects in the learning process.

Secondly, the main challenge in using a large language models (LLMs) for few-shot prompting on sensor data is to bridge the gap between numerical data representation and LLMs' ability to perform context-based semantic reasoning. LLM has been proven effective in deep context understanding of textual data, but they tend to have difficulties handling numerical representations of sensor signals, leading to suboptimal example selection.

Thirdly, the fatigue detection case study used in this research was chosen because it presents a unique challenge suitable as a benchmark for evaluating sample selection methods. Sensor data patterns for fatigue conditions are often highly similar to those of non-fatigue conditions, which makes the challenge of sample selection even more complicated and important to solve appropriately. Previous studies have shown that zero-shot approaches can surpass poorly optimized few-shot approaches, suggesting that structured and contextualized sample selection is critical to ensure optimal model performance.

Given these challenges, this study is motivated to develop a new approach that is more effective in example selection for few-shot prompting on sensor data. Therefore, we propose a Hybrid Euclidean Distance with Large Language Models (HED-LM) approach specifically designed to integrate numerical similarity evaluation with domain-based contextual reasoning performed by LLM. With this approach, we aim to significantly improve the effectiveness of example selection, particularly in complex sensor-based application scenarios, such as fatigue detection. More broadly, this research also lays the foundation for an approach that can be applied to various scenarios in sensor data-driven learning.
\begin{figure}[ht]
\centering
\includegraphics[width=\textwidth]{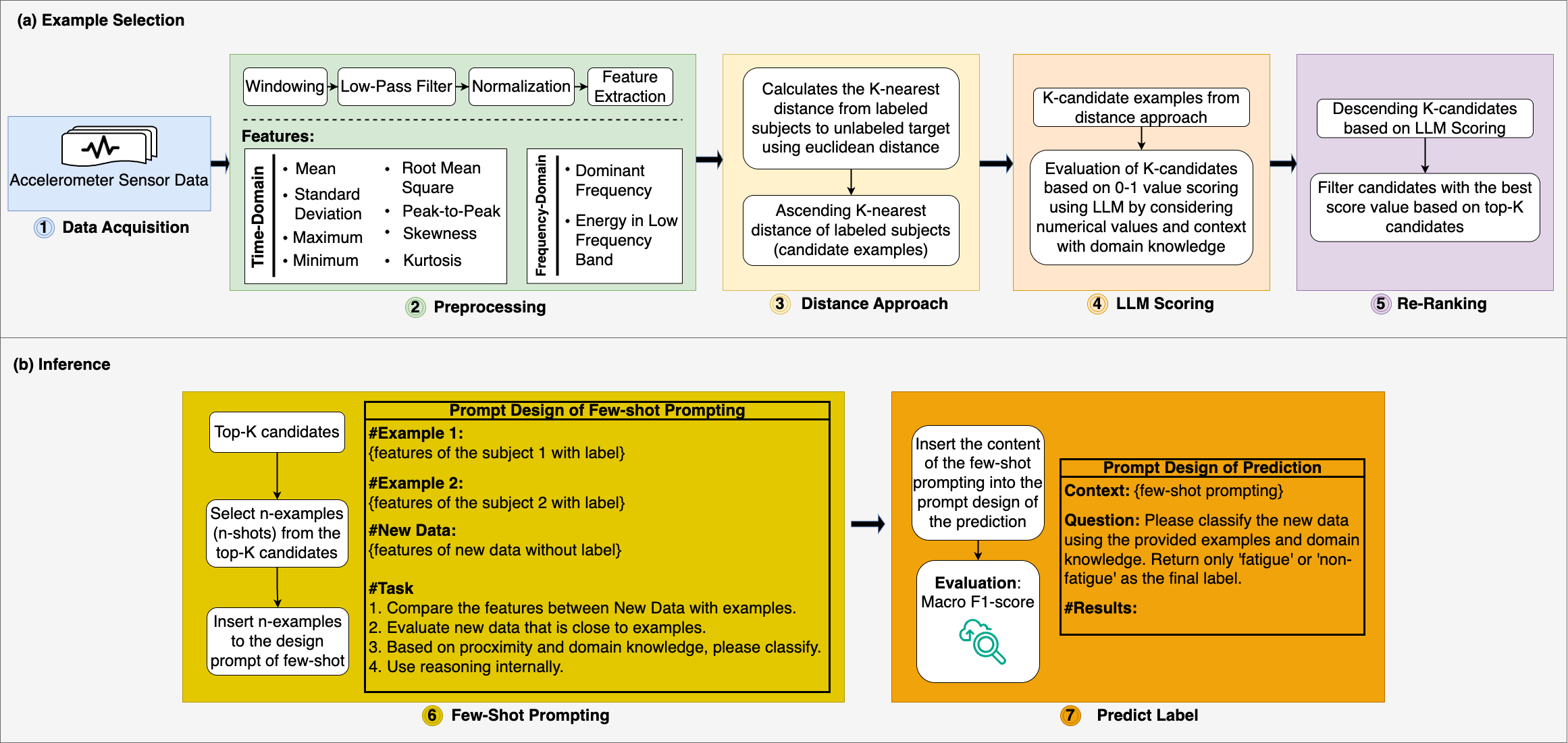}
\vspace{-1em}
\caption{\textbf{Proposed HED-LM framework for physical fatigue detection}. (a) \textit{Example Selection Stage}: The optimal examples are identified based on numerical proximity and contextual relevance using Euclidean Distance with LLM. (b) \textit{Inference Stage}: Few-shot prompting is conducted with LLM utilizing the selected examples to classify physical fatigue.}
\label{fig:HED-LM}
\vspace{-1.5mm}
\end{figure}

%%%%%%%%%%%%%%%%%%%%%%%%%%%%%%%%%%%%%%%%%%
\section{Proposed Method}
\label{methods}
We propose a hybrid framework called HED-LM (Hybrid Euclidean Distance with Large Language Models) to address the challenge of selecting high-quality examples in few-shot prompting for sensor-based classification tasks. As shown in Fig.~\ref{fig:HED-LM}, the structure is made up of two primary phases: the example selection stage, which identifies the most relevant labeled instances based on both numerical similarity and contextual relevance, and the inference stage, where a large language model uses a prompt constructed from the selected examples to predict the label of a new input. The overall design combines the strengths of numerical distance-based filtering, such as Euclidean distance, with LLM-driven semantic evaluation to ensure that the chosen examples are quantitatively similar and semantically aligned with the target input. This dual-filtering approach improves the accuracy, generalizability, and robustness of few-shot prompting, particularly in high-dimensional and noisy sensor data scenarios. The technical details of each component, including data preprocessing, feature extraction, distance-based filtering, LLM-based scoring, re-ranking, prompt construction, and label inference, are elaborated in the following subsections.

\subsection{Data Acquisition}
\label{data-acq}
In this study, we utilize a publicly available dataset comprising accelerometer magnitude signals collected from 19 participants during running activities, with a sampling frequency of 256 Hz \citep{kathirgamanathan2023generating}. Each subject's session contains 180 consecutive samples representing raw acceleration magnitudes calculated from the x, y, and z axes, as defined in Eq.~\ref{eq:1}. 
\begin{equation}
\vert a\vert = \sqrt{a_{x}^{2}+a_{y}^{2}+a_{z}^{2}}
\label{eq:1}
\end{equation}
where $a_{x},a_{y},$ and $a_{z}$ is the accelerometer value of the $x$, $y$, and $z$ axes, respectively. Ground-truth labels were assigned as "fatigue" or "non-fatigue" based on physiological indicators and expert annotation during the activity session. The dataset consists of 6006 labeled subjects, with each instance represented as a single 1D vector of 180 values. This dataset was selected due to its temporal complexity, class imbalance, and high intra-subject variability, making it an appropriate benchmark for evaluating few-shot prompting strategies in sensor-based fatigue detection.

However, this raw data cannot be directly utilized in the framework due to LLM's limited ability to interpret lengthy signal sequences effectively, often leading to misinterpretations \citep{weko_239518_1}. To address this, the data undergoes a preprocessing pipeline to enhance interpretability and align it with the LLM's requirements.

\subsection{Preprocessing}
\label{prep}
Preprocessing transforms the raw accelerometer signals into structured data that captures key characteristics while preserving temporal and spectral information.

\noindent\textbf{Windowing}. 
The accelerometer signal of each subject, comprising 180 samples, is segmented into three equal windows (0–60, 60–120, and 120–180 samples), representing the start, activity, and finishing phases, respectively. With a sampling frequency of 256 Hz, segment 0-60 represents a time duration of 0-234 milliseconds (ms), segment 60-120 represents 234-468 milliseconds (ms), and segment 120-180 represents 468-702 milliseconds (ms). Each phase encapsulates distinct characteristics, such as stabilization, consistent motion, and fatigue-related changes. This segmentation retains critical temporal variations, enabling the system to extract meaningful phase-specific features and detect subtle patterns indicative of fatigue.

By preserving this granularity, windowing supports detailed segment-level analysis, facilitating the extraction of robust time- and frequency-domain features. For each segment, metrics such as mean, standard deviation, min \& max value, peak-to-peak, root mean square (RMS), skewness, kurtosis, dominant frequency, and low-frequency energy are computed, resulting in a 30-dimensional feature vector per subject. This detailed representation ensures that downstream processes can capture signal behaviors crucial for differentiating fatigue from non-fatigue states.

\noindent\textbf{Low-Pass Filtering}.
The proposed method's low-pass filtering stage is crucial for removing high-frequency noise from accelerometer signals while preserving the low-frequency components that carry meaningful information about physical movement. This stage guarantees that the signal quality is adequate for identifying features essential for fatigue detection.

The filtering process is implemented using a Butterworth filter \citep{9408933}, chosen for its smooth frequency response and minimal distortion in the passband. The core idea of the filter is to attenuate frequency components above a defined cutoff frequency ($f_{\text{cutoff}} = 30 Hz$), as physical activities like running or walking typically occur at frequencies below this threshold. The relationship between the cutoff frequency and the sampling rate ($f_s = 256 Hz$) in the dataset is defined in terms of the Nyquist frequency in Eq.~\ref{eq:2}:
\begin{equation}
    f_{\text{Nyquist}} = \frac{f_s}{2}
    \label{eq:2}
\end{equation}
The normalized cutoff frequency is then calculated in Eq.~\ref{eq:3} as:
\begin{equation}
    f_{\text{normalized}} = \frac{f_{\text{cutoff}}}{f_{\text{Nyquist}}}
    \label{eq:3}
\end{equation}

The Butterworth filter is designed with an order of 4, determining the frequency roll-off's steepness. Using the normalized cutoff frequency, the filter coefficients $B$ and $A$ are computed to represent the transfer function in Eq.~\ref{eq:4}:
\begin{equation}
    H(s) = \frac{B(s)}{A(s)}
    \label{eq:4}
\end{equation}
Here, $B(s)$ and $A(s)$ are polynomials of the filter coefficients. To preserve the temporal integrity of the signal, a zero-phase filtering technique is applied using forward-backward filtering, mathematically expressed in Eq.~\ref{eq:5} as:
\begin{equation}
    y[n] = \text{Reverse}(\text{Forward}(x[n], H(s)))
    \label{eq:5}
\end{equation}
where $x[n]$ is the input signal and $y[n]$ is the filtered output. This approach ensures no phase distortion, maintaining the alignment of signal features across time.

The low-pass filtering process refines the accelerometer signal by removing irrelevant high-frequency components, making it cleaner and more representative of the underlying physical activity. This filtered signal becomes the foundation for accurate feature extraction and subsequent classification, contributing to the robustness of the proposed fatigue detection framework.

\noindent\textbf{Normalization}. After filtering, the signal is normalized using min-max scaling. This maps each segment's values into the range $[0, 1]$, preserving relative differences while ensuring uniform scaling across segments. Normalization mitigates disparities in amplitude caused by varying movement intensities, facilitating fair comparisons between features in subsequent analysis.

\noindent\textbf{Feature Extraction}. The filtered and normalized signals are transformed into numerical feature vectors. These features encapsulate both time- and frequency-domain characteristics \citep{s23125756}:
\begin{itemize}
    \item Time-Domain Features: Include mean, standard deviation, max \& min values, root mean square (RMS), skewness, kurtosis, and peak-to-peak values, highlighting statistical and energetic properties.
    \item Frequency-Domain Features: Computed using Fast Fourier Transform (FFT), these include dominant frequency and low-frequency energy, reflecting motion's spectral characteristics.
\end{itemize}

Combining these features, the framework captures a comprehensive representation of the accelerometer signal, enabling robust analysis of fatigue-related patterns.

\subsection{Distance-Based Filtering in Candidate Selection}
\label{distance}
The proposed method transforms each subject's accelerometer data into a feature vector by extracting and concatenating features from three distinct signal segments. For example, a subject's feature vector comprises 30 dimensions with 10 features per segment, drawn from statistical and frequency-based characteristics such as mean, standard deviation, max-min values, root mean square (RMS), skewness, kurtosis, peak-to-peak, dominant frequency and low-frequency energy. These features encapsulate key patterns and dynamics of the accelerometer signal, providing a compact representation for comparing labeled and unlabeled subjects.

When a new subject (unlabeled) is introduced, its accelerometer data undergoes the same preprocessing pipeline to generate a corresponding feature vector, denoted as $\mathbf{x}_{\text{new}}$. The Euclidean distance is then calculated to measure the similarity between this new vector and the feature vectors of previously labeled subjects. The Euclidean distance $d(\mathbf{x}_{\text{new}}, \mathbf{y})$ between $\mathbf{x}_{\text{new}}$ and a labeled subject vector $\mathbf{y}$ is computed in Eq.~\ref{eq:6} as:
\begin{equation}
    d(\mathbf{x}_{\text{new}}, \mathbf{y}) = \sqrt{\sum_{i=1}^{m} (x_{\text{new},i} - y_i)^2}
    \label{eq:6}
\end{equation}
where $m=30$ represents the dimensionality of the feature vectors, smaller distances indicate more remarkable similarities in numerical patterns, such as RMS values and average accelerations. This suggests the new subject may share the same fatigue label as the closest labeled subject. A subsequent LLM scoring stage evaluates label compatibility within a contextual framework to refine the prediction further.

After computing the distances, the labeled subjects are ranked in ascending order based on their proximity to the new subject. The system selects the $K-$nearest candidates (e.g., K=5 or K=10) for further evaluation. This step reduces the computational burden by narrowing the candidate pool, ensuring that the subsequent LLM scoring focuses only on the most relevant labeled subjects. Since LLM-based assessments can be resource-intensive, incorporating this distance-based filtering improves efficiency without compromising accuracy.

The preliminary filtering process identifies the most relevant labeled subjects and creates a structured foundation for the next evaluation stage. The system reduces complexity during the LLM scoring stage by leveraging this streamlined selection process. This approach enhances the model's ability to classify fatigue versus non-fatigue states with improved precision and reliability, ultimately advancing the framework's overall efficacy.

\subsection{LLM Scoring for Contextual Candidate Evaluation}
\label{llm-score}
The proposed method incorporates an advanced LLM-based scoring stage following the initial distance-based screening. This stage evaluates the relevance of labeled candidates (fatigue or non-fatigue) to a new subject by integrating contextual reasoning beyond numerical similarity. By leveraging the LLM’s ability to interpret patterns and assess label appropriateness, the scoring process enriches candidate selection, aligning it more closely with the unique characteristics of the new subject’s signal data. The LLM scoring is executed automatically via an Application Programming Interface (API), ensuring that the relevance assessment process is fully automated and does not require manual intervention.
\begin{enumerate}
    \item \textbf{Structuring Prompts for Effective Contextual Reasoning}. Structured prompts are created for each candidate-new subject pair to facilitate accurate evaluations. These prompts are designed with three essential components:
    \begin{itemize}
        \item \textbf{Numeric Features}: Each prompt provides statistical metrics and frequency-based characteristics extracted from the three signal segments (e.g., segment 1, segment 2, and segment 3). These features represent the signal's critical characteristics.
        \item \textbf{Labeled Subject (Old Subject Label)}: The candidate subject's label (e.g., fatigue or non-fatigue) offers a basis for comparison with the new subject.
        \item \textbf{LLM Guidance}: Specific instructions to direct the reasoning process, such as: "\textit{Please compare the numeric data segment by segment. If the differences in Mean, Std, RMS etc. are very small $\Rightarrow$ high relevance. Also check whether the labeled subject labels (Fatigue/Non-Fatigue) are aligned with the numeric pattern of the new subject.}"
    \end{itemize}
     This structured approach ensures that the LLM provides all necessary information for a detailed and comprehensive assessment, minimizing ambiguity in its evaluation process.
    \item \textbf{Assigning Relevance Scores with Justifications}. The LLM assesses the relevance of each candidate-new subject pair by assigning a score ranging from 0 to 1, where higher scores reflect more substantial alignment between the candidate’s label and the new subject’s signal patterns. An explanation accompanies each score, adding transparency and interpretability to the process. The output format follows a consistent template:
    \begin{itemize}
        \item \textbf{Score}: A numerical value indicating relevance (e.g., SCORE: 0.90).
        \item \textbf{Reason}: A concise justification, e.g.: "\textit{Segments 1 and 2 exhibit high similarity in RMS values; the label ‘fatigue’ is appropriate due to elevated RMS levels in these segments}."
    \end{itemize}
    This dual output format ensures that each decision is well-supported, combining numeric analysis with contextual reasoning. By leveraging the LLM's chain-of-thought reasoning, the scoring mechanism provides a nuanced understanding of the relationships between signal features and labels.
    \item \textbf{Contextual Label Synergy for Accurate Evaluations}. A critical aspect of the scoring process is the assessment of label synergy, where the LLM evaluates how well the candidate’s label aligns with the new subject's signal characteristics. For instance:
    \begin{itemize}
        \item If a labeled subject’s signal (e.g., “fatigue”) align with the numeric patterns of the new subject (e.g., high RMS values in segments 2 and 3), a high relevance score is assigned.
        \item Conversely, if the labeled subject’s label is “non-fatigue” but the new subject exhibits fatigue-like patterns, the relevance score is reduced, even in high numeric similarity.
    \end{itemize}
    This contextual evaluation is formalized in Eq.~\ref{eq:7} as follows:
    \begin{equation}
        \text{Relevance}_{\text{LLM}}(\mathbf{x}_{\text{old}}, \mathbf{x}_{\text{new}}, \text{label}_{\text{old}}) \approx f\big(\Delta(\mathbf{x}_{\text{old}}, \mathbf{x}_{\text{new}}), \text{synergy}(\text{label}_{\text{old}}, \mathbf{x}_{\text{new}})\big)
        \label{eq:7}
    \end{equation}
    where $\mathbf{x}_{\text{old}}$ is the feature vector on the labeled subject; $\mathbf{x}_{\text{new}}$ is the feature vector on the new subject (unlabeled subject); $\text{label}_{\text{old}}$ is the labeled subject; $\Delta(\cdot)$ measures numerical differences in features such as mean, RMS, etc.; $\text{synergy}(\cdot)$ evaluates the alignment of the old subject’s label with the new subject’s signal pattern; and $f(\cdot)$ represents the LLM’s reasoning process. The LLM scoring mechanism enhances the precision of candidate selection for few-shot prompting by combining numerical comparisons with label alignment.
\end{enumerate}

The LLM scoring stage bridges numerical signal analysis and label semantics, refining the few-shot prompting pipeline. Candidates with high numeric similarity and label alignment are prioritized, ensuring the most contextually relevant examples are selected. To further enhance this contextual alignment, domain knowledge explicitly guides the scoring process by providing clearly defined numerical thresholds derived from domain expert analysis and prior empirical studies. For instance, specific rules such as ‘RMS values above 0.5 in segments 2 and 3 typically indicate fatigue’ or ‘mean acceleration values below 0.31 strongly suggest fatigue conditions’ significantly increase the precision of contextual evaluation, reducing ambiguity during candidate assessment. This dual evaluation framework mitigates the risk of misclassification by deprioritizing candidates with conflicting labels, even if they exhibit numerical similarity.

This intelligent scoring process strengthens the few-shot prompting methodology by ensuring that the selected examples effectively guide the classification task. Consequently, the proposed approach demonstrates increased reliability and precision in differentiating between fatigue and non-fatigue states, rendering it a strong solution for detecting physical fatigue. More comprehensive details regarding the domain-specific thresholds and their integration into LLM prompts are explained in Appendix~\ref{apd:llm_score}. Moreover, more details on how to generate domain knowledge, the form of domain knowledge examples, and their placement in the prompting design are explained in Appendix~\ref{apd:dom_know}.

\subsection{Re-Ranking}
\label{rank}
The Re-Ranking stage in the proposed method serves as a critical intermediary between the distance-based filtering and the LLM scoring processes. After identifying the closest $\text{distance}_K$ subjects using Euclidean distance and computing their relevance scores $\text{LLMScore} \in [0,1]$ through the LLM evaluation, this stage reorders the candidates based on their LLM scores. Re-ranking transcends simple numerical similarity by prioritizing contextually relevant examples, ensuring that label alignment and contextual reasoning are incorporated into the candidate selection process.

Euclidean distance provides an initial measure of proximity between the new subject’s feature vector $\mathbf{x}^{(\text{new})}$ and the labeled candidates $\mathbf{x}^{(\text{old})}$. However, distance alone cannot capture the nuanced relationships LLM scoring offers. While numerical similarity identifies potentially relevant subjects, the LLMScore considers numerical patterns and label synergy, providing a more robust criterion for evaluating relevance.
Mathematically, the Re-Ranking process is represented in Eq.~\ref{eq:8} as:
\begin{equation}
    \text{Top-K Candidates}(\mathbf{x}^{(\text{new})}) = \underset{K-\text{ candidates}}{\mathrm{argsort\_desc}} \Bigl( \{\text{LLMScore}(\mathbf{x}^{(\text{old})}, \mathbf{x}^{(\text{new})})\}_{j=1}^{\text{distance}_{K}} \Bigr)
    \label{eq:8}
\end{equation}
where:
\begin{itemize}
    \item $\mathbf{x}^{(\text{new})}$: Feature vector of the new subject.
    \item $\mathbf{x}^{(\text{old})}x(old)$: Feature vector of a labeled candidate within the $\text{distance}_K$ closest subjects.
    \item $\text{LLMScore}(\cdot)$: Relevance score assigned by the LLM, reflecting numeric similarity and label alignment.
    \item $\mathrm{argsort\_desc}$: A function that indexes candidates in descending order of their LLM scores and selects the top-$K$ candidates.
\end{itemize}
This approach ensures that candidates with high LLM scores indicative of numeric and contextual relevance are prioritized for inclusion in the few-shot prompt or the final label prediction.

Distance-K and top-K are pivotal in shaping the efficiency and accuracy of the proposed HED-LM framework. Specifically, distance-K defines how many of the closest labeled candidates (based on Euclidean distance) are initially retrieved for evaluation. At the same time, top-K controls how many of those are selected after LLM-based scoring for the final decision-making. This hierarchical filtering strategy helps reduce computational overhead by narrowing down the candidate set before invoking the more expensive LLM reasoning. From a system design perspective, this ensures a balance between numeric similarity and semantic alignment, allowing the framework to prioritize quantitative and contextually relevant candidates. The specific values for distance-K and top-K used in this work (e.g., 5 and 3 or 10 and 5) were tuned based on empirical observations, as discussed further in the experimental setup.

\subsection{Few-Shot Prompt}
\label{few-shot}
The few-shot prompt includes two examples (2-shot) from the re-ranked results, ensuring representation from both label categories: fatigue and non-fatigue. Each example is designed to contain the following key components:
\begin{itemize}
    \item \textbf{Numeric Data Summary}: A concise representation of extracted features from the three signal segments (Segment 1, Segment 2, and Segment 3).
    \item \textbf{Relevance Score and Reason}: The relevance score assigned during the LLM scoring stage and a brief explanation of why the example aligns contextually with the new subject.
    \item \textbf{Label Conclusion}: A definitive statement summarizing the example’s label, such as “Conclusion: The label is fatigue” or “Conclusion: The label is non-fatigue.”
\end{itemize}
After presenting these examples, the numeric data of the new subject (unlabeled) is appended to the prompt, along with an instruction guiding the LLM to make a final decision:
“\textit{Please compare the new data with [Example 1 and Example 2] and determine the final label: ‘fatigue’ or ‘non-fatigue.’}”
This structured design ensures that the LLM provides concrete examples illustrating how specific numerical features correlate with their respective labels, enhancing its ability to generalize and make a well-informed prediction.

Few-shot prompting can be represented in Eq.~\ref{eq:9} as:
\begin{equation}
    \resizebox{.9\hsize}{!}{$\mathbf{P} = \underbrace{\text{(Example 1: numeric + label)}}_{\text{Shot-1}} + \underbrace{\text{(Example 2: numeric + label)}}_{\text{Shot-2}} + \underbrace{\text{(New Data: numeric)}}_{\text{Unlabeled Input}} \xrightarrow[]{\text{LLM}} \text{(Final Label)}$}
    \label{eq:9}
\end{equation}
Here:
\begin{itemize}
    \item $\mathbf{P}$ is the prompt that combines two labeled examples and one unlabeled target.
    \item Shot-1 and Shot-2 provide the LLM with explicit references to numerical patterns and their associated labels.
    \item The new data serves as the input for the LLM’s final label prediction.
\end{itemize}
In our case, Shot-1 and Shot-2 taken should have a representative criterion of one label “fatigue” and “non-fatigue”. If in the top-K candidates, no such criterion is found, then Shot-1 and Shot-2 may be labeled “fatigue” or “non-fatigue” respectively.

By combining these elements, the prompt ensures that the LLM has sufficient context to evaluate the new subject against the labeled examples and predict the appropriate label. More details of Few-shot prompting are described in Appendix~\ref{apd:few_shot}.

The few-shot prompt's use of two examples (2-shot) was carefully considered to balance performance, interpretability, and computational efficiency. Including one representative example from each label category, fatigue, and non-fatigue, ensures that the prompt provides clear reference points for the LLM compared with the unlabeled data. This binary-class representation structure is particularly effective for classification tasks with subtle label boundaries. Additionally, using only two examples helps reduce the prompt length. It minimizes the computational cost associated with API-based LLM inference while preserving the context for accurate decision-making.

\subsection{Predict Label}
\label{predict}
Once the few-shot prompt is constructed, the LLM is invoked to classify the new subject, assigning it a label of “fatigue” or “non-fatigue.” This stage integrates all preceding processes, for example, selection, feature extraction, and prompt formulation, guiding the LLM to make a precise and contextually informed decision.

\begin{enumerate}
    \item \textbf{Calling the LLM for Label Prediction}. The finalized prompt, including the two most contextually relevant examples (2-shot) and the new subject’s unlabeled numeric data, is submitted to the LLM with explicit instructions. The LLM’s task is to generate a concise, single-word output, either “fatigue” or “non-fatigue,” to ensure focus and eliminate unnecessary verbosity. The process can be represented in Eq.~\ref{eq:10} as:
    \begin{equation}
        \text{Reply} = \text{LLM}(\mathbf{P})
        \label{eq:10}
    \end{equation}
    where $\mathbf{P}$ is the constructed prompt containing the selected examples and the new subject's numeric data. This approach ensures that the LLM remains focused on label prediction, using the provided examples to make an informed decision based on numerical and contextual relevance.
    \item \textbf{Parsing the LLM’s Response}. The output of the LLM is parsed to extract the final label:
    \begin{itemize}
        \item If the response is “fatigue,” the new subject is labeled “fatigue.”
        \item If the response is “non-fatigue,” the new subject is labeled “non-fatigue.”
    \end{itemize}
    In cases where the LLM produces an ambiguous response (e.g., “I believe it is non-fatigue, but maybe fatigue?”), a fallback mechanism is employed. This involves analyzing the frequency of “fatigue” and “non-fatigue” in the LLM’s output. The label with the higher frequency is selected, ensuring systematic resolution of ambiguity without compromising the consistency of the prediction process.
    \item \textbf{Mapping Numeric Data to Labels}. The overall label prediction process can be formalized mathematically in Eq.~\ref{eq:11} as follows:
    \begin{equation}
        \text{label}(\mathbf{x}^{(\text{new})}) = \mathrm{Argmax}\Bigl(\text{LLM}\bigl(\text{prompt shots}, \mathbf{x}^{(\text{new})}\bigr)\Bigr)
        \label{eq:11}
    \end{equation}
    where:
    \begin{itemize}
        \item $\mathbf{x}^{(\text{new})}$: Feature vector of the new subject.
        \item \text{prompt shots}: The labeled examples are provided in the few-shot prompt.
        \item $\mathrm{Argmax}$: Function selecting the label (“fatigue” or “non-fatigue”) with the highest confidence or frequency.
    \end{itemize}
    This representation captures the relationship between the new subject’s numeric data, the prompt examples, and the LLM’s output. It formalizes the label prediction process as a function of both prompt design and the LLM’s reasoning capabilities.
\end{enumerate}

Algorithm~\ref{algo:1} outlines the end-to-end workflow of the HED-LM approach. The framework systematically integrates preprocessing, feature extraction, candidate selection, LLM evaluation, and few-shot prompting to deliver accurate fatigue detection. Performance is assessed using the macro F1-Score, ensuring balanced evaluation across fatigue and non-fatigue classifications.
\begin{algorithm}[H]
\caption{HED-LM Algorithm with k-shot Examples}
\label{algo:1}
\begin{algorithmic}[1]
\Require File Path ($F_{\text{path}}$), Distance ($k_d$), Top-$K$ ($k_t)$
\Require Domain Knowledge ($\text{DK}$), Few-Shot Examples ($n_{\text{shots}} = k$)

\State \textbf{Load Data:}
\State $D \gets \text{Load dataset from } F_{\text{path}}$
\State $X \gets \text{Extract features: } X = \{x_1, x_2, \dots, x_n\}$
\State $L \gets \text{Extract labels: } L = \{\ell_1, \ell_2, \dots, \ell_n\}$

\State \textbf{Process Features from preprocessing stages:}
\State $F \gets \text{Feature set: } F = \{f_1, f_2, \dots, f_n\} \text{ derived from } X \text{ and } L$

\State \textbf{Initialize LLM:}
\State $\text{LLM} \gets \text{Language model initialized with specified parameters}$

\State \textbf{Define Prompt and Create Chain:}
\State $P \gets \text{Prompt generated using } \text{DK and task-specific questions}$
\State $\text{Chain} \gets \text{Combine } P \text{ and LLM}$

\State \textbf{Hybrid Euclidean Distance with LLM Scoring (HED-LM):}
\State $R \gets \text{Score features: } R = \{r_1, r_2, \dots, r_n\} \text{ using } F, \text{Chain}, k_d, k_t$

\State \textbf{Final Prediction:}
\For{$i = 1$ to $n$}
    \State $T_i \gets k\text{-shot} \text{ examples from } R[i]$
    \State $P_i \gets \text{Few-shot prompt generated for feature } f_i \text{ using } T_i$
    \State $\ell_{\text{pred}, i} \gets \text{Predicted label from } \text{Chain and } P_i$
    \State $\text{Results} \gets \text{Results} \cup \{\ell_i, \ell_{\text{pred}, i}\}$
\EndFor

\State \textbf{Evaluation Results:}
\State $\text{MacroF1} \gets \frac{1}{C} \sum_{c=1}^C \frac{2 \cdot \text{Precision}_c \cdot \text{Recall}_c}{\text{Precision}_c + \text{Recall}_c}$
\State Output: $\text{MacroF1}$

\end{algorithmic}
\end{algorithm}

%%%%%%%%%%%%%%%%%%%%%%%%%%%%%%%%%%%%%%%%%%
\section{Experiments}
\label{experiments}
In this section, we detail the experiments conducted to assess the performance of our approach (HED-LM), which combines a distance approach and LLM scoring for few-shot optimization in detecting physical activity fatigue based on accelerometer signal data. We divide our experimental description into several points: the experimental setup and the experimental results. 

\subsection{Experimental Setup}
\label{ex_setup}
\noindent\textbf{Experiment Objectives}.
There are two main points of objectives carried out in our experiments, including:
\begin{enumerate}
    \item \textbf{Testing the effectiveness of our approach}. We intend to examine whether the end-to-end framework of our approach can improve the classification of physical fatigue detection compared to three baselines: (a) traditional machine learning, (b) randomized approach, and (c) distance approach.
    \item \textbf{Assessing the role of domain knowledge}. We also investigate how the influence of domain knowledge inserted in the LLM scoring and prediction process can affect the relevance assessment of the original subject to the new subject so that the final label becomes more precise in performance.
\end{enumerate}

\noindent\textbf{Datasets and Sensor Contexts}. 
We utilized the publicly available dataset from Kathirgamanathan et al.~\citep{kathirgamanathan2023generating}, comprising accelerometer-based time series data collected from 19 recreational runners using lumbar-mounted Shimmer IMUs at a sampling rate of 256 Hz. Each subject completed three tasks on an outdoor running track: (1) a 400-meter run under non-fatigued conditions, (2) a multistage beep test used to induce fatigue, and (3) a follow-up 400-meter run in a fatigued state. The primary signal was the acceleration magnitude derived from the tri-axial accelerometer data. From the two 400-meter runs per subject, individual strides were segmented and resampled using a Soft-DTW barycenter smoothing technique, yielding approximately 6,006 instances for binary classification (fatigue vs. non-fatigue). This protocol ensures well-controlled fatigue induction and allows for subject-specific, stride-level analysis.

\noindent\textbf{Preprocessing and Feature Extraction}. Each 180-sample signal was windowed into three segments x 60 samples. We then applied a low-pass filter (30 Hz cutoff) to reduce noise, followed by min-max normalization per segment. The retrieved features include men, standard deviation, max, min, peak-to-peak, RMS, skewness, kurtosis, dominant frequency, and energy lowband for each segment. Thus, each subject has 30 features (3 segments x 10 features).

\noindent\textbf{Distance and LLM parameters}. In our approach, we set the parameters of the distance-K ($distance_{K}$) and top-K ($top_{K}$) effects of LLM Scoring as follows:
\begin{itemize}
    \item \#Param{A} has a distance parameter setting and LLM scoring that uses Euclidean distance with $distance_{K} = 5$, meaning we only take the five closest subjects as candidates and perform LLM scoring with $top_{K} = 3$, which only takes the three best candidates from the five candidate distance approach based on integrated relevance assessment with domain knowledge.
    \item \#Param{B} has distance and LLM scoring parameter settings, namely using Euclidean distance with $distance_{K} = 10$, meaning that we only take the 10 closest subjects as candidates and perform LLM scoring with $top_{K} = 5$, which only takes the five best candidates from 10 distance approach candidates based on integrated relevance assessment with domain knowledge.
\end{itemize}
The comparison of the two parameters-(\#ParamA) with $distance_K=5$, $top_K=3$ and (\#ParamB) with $distance_K=10$, $top_K=5$-was done to see the effect of the size of the candidates called and selected by the LLM on the performance of the HED-LM method. When $distance_K$ and $top_K$ are more minor (as in \#ParamA), the process becomes more efficient due to fewer LLM calls and the number of final examples selected. However, the candidate coverage is also narrower. In contrast, in \#ParamB, more candidates are assessed and retrieved, increasing the chance of finding the most similar subjects. By comparing the two, we can assess the coverage-related trade-off (potentially higher performance) in parameterizing the HED-LM method. 

We further conducted sensitivity experiments to evaluate the impact of varying $distance_K$ and $top_K$ values on performance. Lower values of $distance_K$ (e.g., 3 or 5) reduced LLM call costs. However, they often excluded contextually relevant candidates, while higher values (e.g., 15) increased processing time without consistent performance gains due to the inclusion of less relevant examples. Similarly, increasing $top_K$ beyond 5 diluted the label relevance scoring, as weaker candidates were retained. Empirically, we found that $distance_K=10$ and $top_K=5$ yielded the best balance between coverage and LLM scoring accuracy, while $distance_K=5$ and $top_K=3$ offered faster computation with slightly reduced performance. These configurations reflect trade-offs between precision and efficiency, which are further analyzed in subsection~\ref{dist-top_k}.

For both parameter settings, we use the same model, GPT-4o-mini, with parameter setting temperature = 0.3, which can provide more controllable and consistent model control. We chose the GPT-4o-mini because it performs well in understanding accelerometer sensor magnitude data \citep{weko_239518_1}.

\noindent\textbf{Performance evaluation}. In this study, we evaluate four approaches for fatigue detection from wearable sensor data:
\begin{enumerate}
    \item \textbf{Traditional Machine Learning (ML)}: We used a Random Forest classifier trained on 30 features, namely eight time-domain statistical features and two frequency-domain features, which were calculated across three equally sized segments per input trace from raw accelerometer data. It does not involve any prompting or interaction with large language models. The inference is conducted locally, making it an LLM-independent, fully offline baseline.
    \item \textbf{Random Approach}: A random selection of examples is incorporated into few-shot prompting with an API GPT-4o-mini model, temperature = 0.3, and added domain knowledge information generated with the GPT-4o model.
    \item \textbf{Distance Approach}: Using Euclidean distance only (KNN-based), subjects with numerically close distances are input examples into few-shot prompting with GPT-4o-mini model and temperature = 0.3, adding domain knowledge information generated with the GPT-4o model.
    \item \textbf{Our proposed method, HED-LM (Hybrid Euclidean Distance with Large Language Models)}, filters distance-based similarity and then ranks the selected examples based on contextual relevance scored by an LLM before constructing the prompt.
\end{enumerate}
To ensure clarity in the computational and architectural differences between these methods, Table~\ref{tab:method_comparison} summarizes the core characteristics of each approach.
\begin{table}[ht]
\centering
\caption{Summary of characteristics across evaluated approaches. Only the ML model is LLM-independent and operates without prompt-based inference.}
\label{tab:method_comparison}
\resizebox{\textwidth}{!}{%
\begin{tabular}{|l|c|c|c|c|}
\hline
\textbf{Approach} & \textbf{Uses LLM} & \textbf{Example Selection} & \textbf{API Call} & \textbf{Prompt-Based} \\
\hline
ML (Random Forest)    & No  & No (Offline learning)           & No  & No \\
Random Approach       & Yes & Random selection                & Yes & Yes \\
Distance Approach     & Yes & Euclidean distance              & Yes & Yes \\
HED-LM (Ours)         & Yes & Distance + LLM semantic scoring & Yes & Yes \\
\hline
\end{tabular}
}
\end{table}
\begin{equation}
    \text{Macro F1-Score} = \frac{1}{k} \sum_{i=1}^{k} \frac{2 \cdot \text{Precision}_i \cdot \text{Recall}_i}{\text{Precision}_i + \text{Recall}_i}
    \label{eq:12}
\end{equation}
where $k$ is the total number of classes; $\text{Precision}_i$  and $\text{Recall}_i$  is calculated for each class $i$.

All models are evaluated using macro F1-score, as shown in Eq.~\ref{eq:12}, as the primary metric under a 2-shot configuration. A 2-shot configuration was used based on exploratory experiments comparing 2-shot prompting and full-shot. Our findings showed that performance gains beyond the two examples were marginal, while longer prompts increased LLM latency and cost. Furthermore, the 2-shot format provides a balanced and interpretable structure for binary classification by selecting one example per class, fatigue, and non-fatigue. This setup offers a practical compromise between model effectiveness and computational efficiency, making it suitable for real-time or resource-constrained scenarios. The 2-shot usage evaluation details are explained in subsection~\ref{number-shot}.

To ensure robust and fair assessment across all methods, especially given the inter-subject variability inherent in physiological data, each experimental run was conducted on an isolated per-user subset. This user-specific evaluation setup prevents cross-user information leakage, supports subject-wise generalization analysis, and ensures that all classification results reflect personalized model behavior rather than pooled training effects.

\textbf{Evaluation Protocol and Prompt Construction}. To ensure methodological consistency and fair evaluation across all approaches, we adopted a unified experimental protocol that applies equally to the traditional machine learning (ML) baseline and the three LLM-based prompting strategies. Although the full dataset comprises 6,006 labeled sensor instances, all experiments were conducted independently on a \textit{per-user} basis, using a fixed slice of the dataset corresponding to each subject (e.g., user10: samples 1736 to 2063, totaling 327 instances). For the ML baseline, we trained a Random Forest classifier implemented using the \texttt{RandomForestClassifier} from \texttt{scikit-learn} (version 1.6.1) with 100 estimators and \texttt{random\_state=42}. A minimal few-shot configuration was used, where $n = 2$ examples, one fatigue and one non-fatigue sample, were selected for training, and the remaining instances served as the test set. The same test set was reused without modification across all LLM-based methods, including random selection, distance-only, and our proposed Hybrid Euclidean Distance with Large Language Models (HED-LM).

For each test instance, a prompt was constructed using exactly two support examples drawn from the same user's data, strictly excluding the test instance itself to ensure \textit{no data leakage or label contamination}. In the random approach, examples were sampled uniformly at random; in the distance-only method, we selected the two most similar examples based on Euclidean distance in the 30-dimensional feature space; and in the HED-LM method, we first filtered candidates based on distance and then ranked them using an LLM relevance scoring mechanism that evaluated numerical similarity and semantic label alignment. All test samples were predicted independently using these prompt structures. Using a consistent $n = 2$ across both ML and LLM paradigms ensures a balanced and realistic few-shot comparison, particularly under low-resource settings. The fixed-size prompting strategy was also chosen to remain within the token limits of the LLM model (GPT-4o-mini) and to reflect practical constraints in real-world deployment scenarios. To further clarify the consistency of the evaluation setup across all approaches, Table~\ref{tab:evaluation-setup} summarizes the key components of the experimental design. The table outlines the data source, training configuration, test isolation strategy, selection methods, and prompt construction across the machine learning baseline and all LLM-based methods. All models were evaluated on the same per-user data slice using exactly two support examples per test instance, ensuring methodological alignment and eliminating the risk of data leakage.
\begin{table}[ht]
\centering
\caption{Evaluation setup and prompting strategies across ML and LLM-based methods.}
\renewcommand{\arraystretch}{1.2}
\resizebox{\textwidth}{!}{
\begin{tabular}{|>{\color{black}}l|>{\color{black}}c|>{\color{black}}c|>{\color{black}}c|>{\color{black}}c|}
\hline
\rowcolor{black!15}
\textbf{Component} & \textbf{ML (Random Forest)} & \textbf{Random Approach} & \textbf{Distance Approach} & \textbf{HED-LM (Ours)} \\
\hline
Data Source & Per-user slice & Same & Same & Same \\
Data Range & Per-user slice & Same & Same & Same \\
Training Samples ($n$) & 2 & 2 & 2 & 2 \\
Test Set & Held-out & Same & Same & Same \\
Selection Strategy & Random from class & Random & Euclidean distance & Distance + LLM scoring \\
Prompt Structure & N/A & 2-shot & 2-shot & 2-shot, label-balanced \\
Leakage Prevention & Yes & Yes & Yes & Yes \\
\hline
\end{tabular}
}
\label{tab:evaluation-setup}
\end{table}

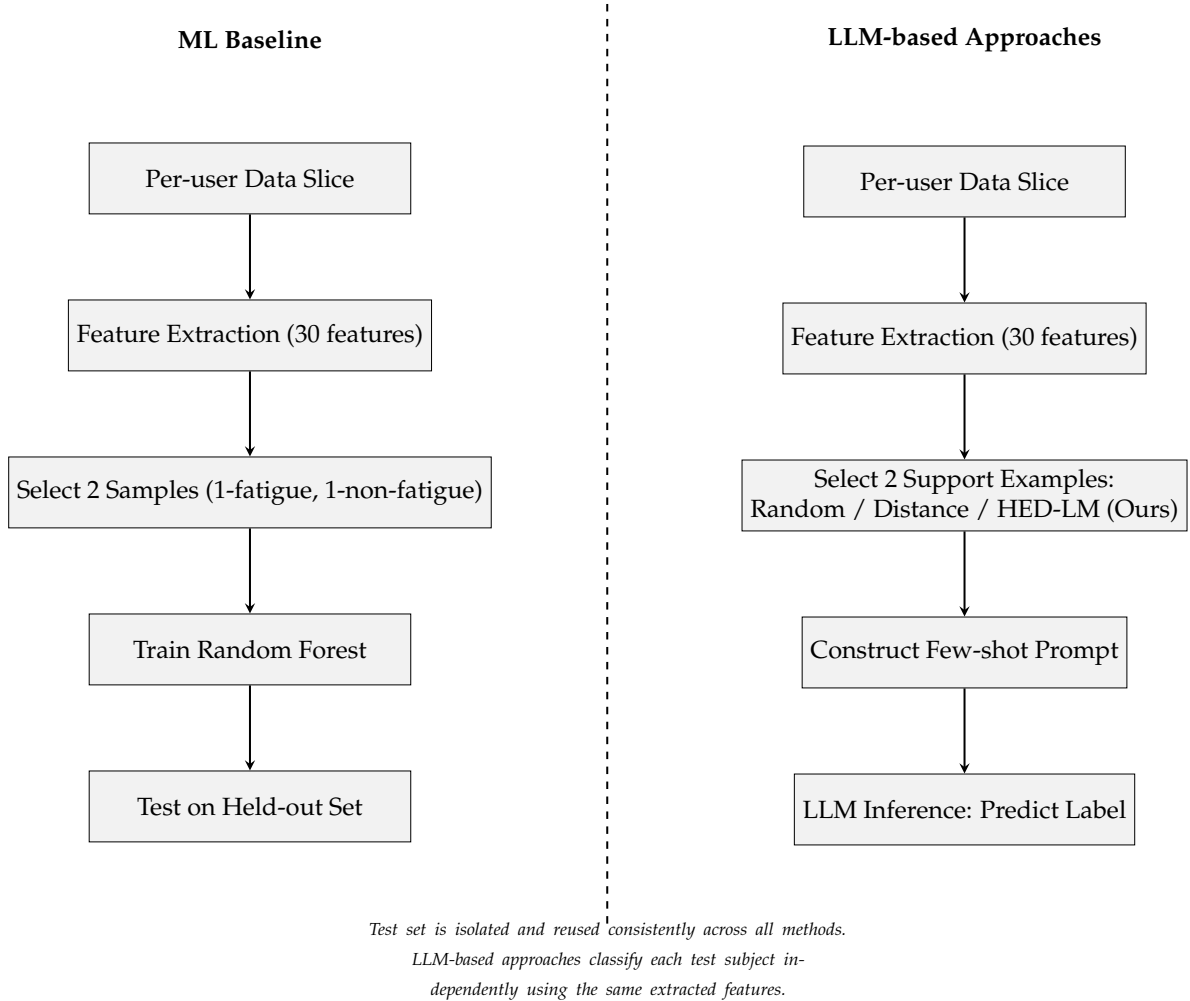
\begin{figure}[ht]
\centering
\resizebox{\linewidth}{!}{%
\begin{tikzpicture}[
    node distance=1.2cm,
    process/.style={
        rectangle, 
        minimum width=4.5cm, 
        minimum height=1cm, 
        text centered, 
        draw=black, 
        fill=gray!10, 
        text=black
    },
    arrow/.style={thick, ->, >=stealth}
]
% ML Column
\node (ml_title) at (-5,1) {\textbf{ML Baseline}};
\node (ml_data) [process, below=of ml_title] {Per-user Data Slice};
\node (ml_feat) [process, below=of ml_data] {Feature Extraction (30 features)};
\node (ml_sel)  [process, below=of ml_feat] {Select 2 Samples (1-fatigue, 1-non-fatigue)};
\node (ml_train) [process, below=of ml_sel] {Train Random Forest};
\node (ml_test) [process, below=of ml_train] {Test on Held-out Set};
\draw [arrow] (ml_data) -- (ml_feat);
\draw [arrow] (ml_feat) -- (ml_sel);
\draw [arrow] (ml_sel) -- (ml_train);
\draw [arrow] (ml_train) -- (ml_test);
% LLM Column
\node (llm_title) at (5,1) {\textbf{LLM-based Approaches}};
\node (llm_data) [process, below=of llm_title] {Per-user Data Slice};
\node (llm_feat) [process, below=of llm_data] {Feature Extraction (30 features)};
\node (llm_sel)  [process, below=of llm_feat, align=center] 
    {Select 2 Support Examples:\\ Random / Distance / HED-LM (Ours)};
\node (llm_prompt) [process, below=of llm_sel, align=center] {Construct Few-shot Prompt};
\node (llm_pred) [process, below=of llm_prompt, align=center] {LLM Inference: Predict Label};
\draw [arrow] (llm_data) -- (llm_feat);
\draw [arrow] (llm_feat) -- (llm_sel);
\draw [arrow] (llm_sel) -- (llm_prompt);
\draw [arrow] (llm_prompt) -- (llm_pred);
% Vertical separator with red annotation
\draw[dashed, thick] (0,1.5) -- (0,-11.5);
\node at (0,-12) [
    align=center, 
    text width=9cm, 
    draw=none, 
    inner sep=1pt,
    text=black
] {
    \scriptsize\textit{Test set is isolated and reused consistently across all methods.} \\
    \scriptsize\textit{LLM-based approaches classify each test subject independently using the same extracted features.}
};
\end{tikzpicture}
}
\caption{Comparison of evaluation pipelines between the ML baseline and LLM-based approaches.}
\label{fig:eval-pipeline}
\end{figure}

Figure~\ref{fig:eval-pipeline} illustrates the complete data processing and inference pipelines across all compared approaches to support the detailed evaluation setup described earlier visually. All methods begin with a per-user data slice, followed by preprocessing and feature extraction steps. For the machine learning baseline (Random Forest), two samples (1 fatigue, 1 non-fatigue) are used to train the classifier, and the remaining instances are held out for testing. In contrast, the LLM-based approaches use the remaining pool (excluding the test sample) to construct prompts based on different support example selection strategies. The random approach selects support examples arbitrarily, while the distance-only approach selects based on Euclidean similarity. At the same time, our proposed HED-LM method combines distance filtering with LLM-based scoring to select the most semantically aligned examples. All test predictions are made independently, and the same test set is used consistently across methods. This diagram emphasizes the strict isolation of test instances and ensures that all predictions are made without leakage from the evaluation set.

\subsection{Experimental Results}
\label{ex_result}
\subsubsection{Impact of Distance-K and Top-K Parameters}
\label{dist-top_k}
\begin{figure}[ht]
\begin{center}
 \includegraphics[width=0.8\textwidth]{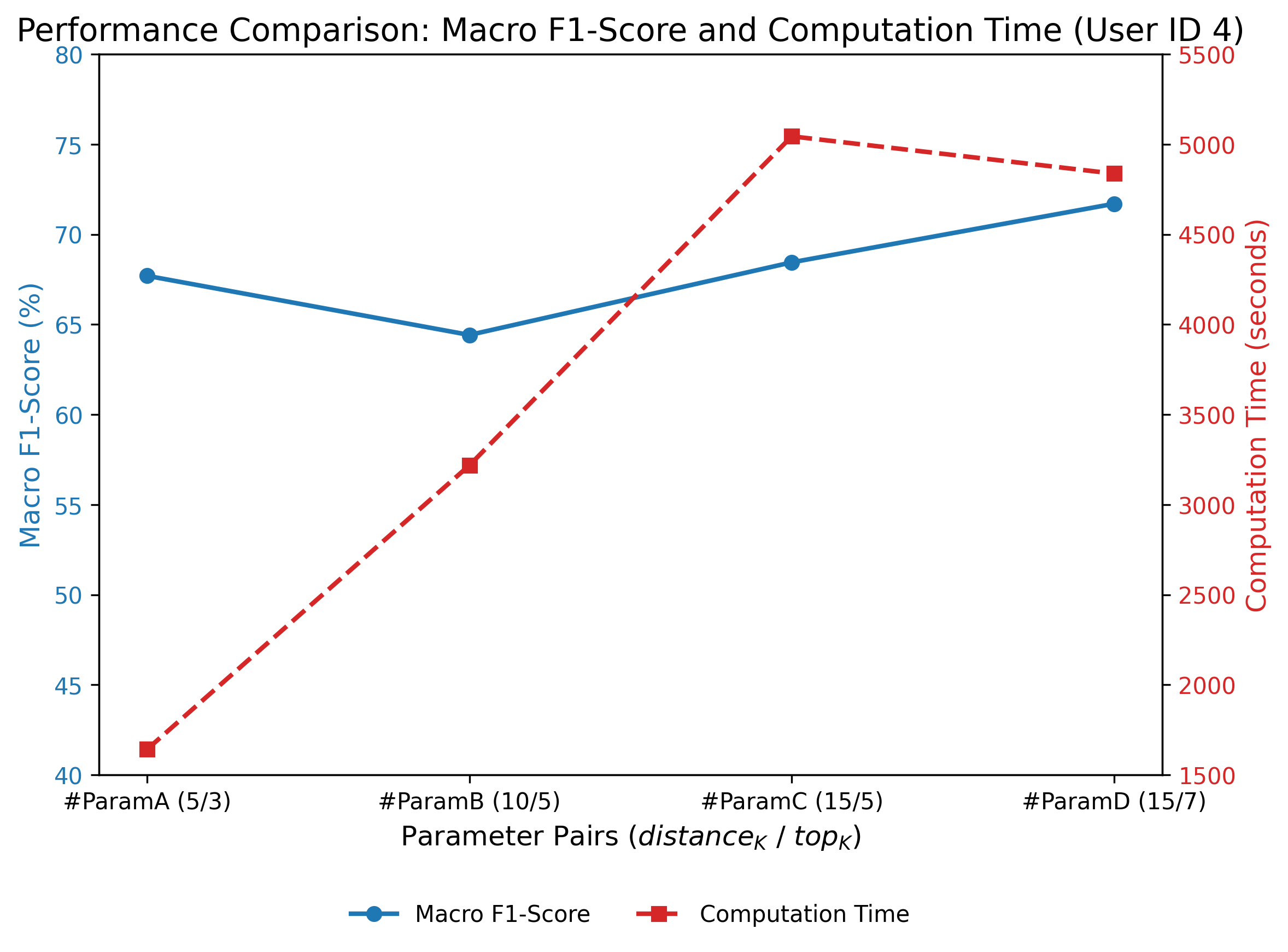}
\end{center}
\vspace{-1em}
\caption{\textbf{Comparison of HED-LM performance and computation time across different parameter configurations of distance-K and top-K on User ID 4}. The blue line (left axis) shows the macro F1-score (\%) for each configuration, while the red dashed line (right axis) represents the total computation time in seconds. While higher parameter values such as (15/7) yield better accuracy, they come with substantially higher computational costs. Configurations (10/5) and (5/3) demonstrate a more favorable trade-off between performance and efficiency.}
\label{fig:param_effect}
\vspace{-2mm}
\end{figure}

Before analyzing the final performance comparison with baseline methods, we first evaluated the effect of key parameters on the performance of HED-LM. In particular, we compared four configurations: \#ParamA with $ distance_K=5$, $ top_K=3$, \#ParamB with $ distance_K=10$, $ top_K=5$, \#ParamC with $ distance_K=15$, $ top_K=5$, and \#ParamD with $ distance_K=15$, $ top_K=7$. The results, illustrated in Figure~\ref{fig:param_effect}, demonstrate a clear trade-off between classification performance and computational cost.

From a performance perspective, increasing distance-K and top-K generally improved macro F1-scores. Specifically, the configuration (15/7) achieved the highest performance at 71.70\%, followed closely by (15/5) at 68.45\% and (10/5) at 64.42\%. However, this increase in performance comes with a steep rise in computational time. Computation time more than doubled from (5/3) to (15/5), reaching over 5000 seconds for the latter. Interestingly, although (15/7) slightly outperformed (15/5) in macro F1-score, it required comparable computation time, yielding only a marginal improvement of approximately 3\%.

These findings suggest that while larger values of distance-K and top-K enhance the contextual richness of few-shot prompts by broadening the candidate pool and improving relevance selection, performance gains plateau while computational costs continue to rise. Therefore, configurations (10/5) and (5/3) offer more favorable trade-offs.

The configuration (10/5) balances accuracy and efficiency, achieving a solid macro F1 Score of approximately 64–66\% with a moderate computation time. Meanwhile, (5/3) is particularly attractive for resource-constrained environments, as it offers competitive accuracy (around 67–68\%) with substantially lower computation time, less than half of that required by the (15/5) and (15/7) configurations.

In summary, while the (15/7) configuration yields the highest macro F1-score, its computational demands make it less practical for real-world deployment. We, therefore, recommend using (10/5) as the default configuration for general use cases requiring a balance between performance and latency and (5/3) for deployment scenarios with tighter computational or time constraints. For this reason, our research focuses on performance comparison using \#ParamA and \#ParamB.

\subsubsection{Effect of the Number of Shots in Few-Shot Prompting}
\label{number-shot}
To assess the effect of the number of shots in few-shot prompting, we compared the performance and computation time of the HED-LM method using 2-shot and full-shot configurations, where all available support instances, excluding the test sample, are used in the prompt without filtering or semantic selection. The 2-shot setting includes two representative labeled examples (one per class) as input prompts, whereas the full-shot setting uses all available top-K examples selected via LLM scoring.
\begin{figure}[ht]
\begin{center}
 \includegraphics[width=0.7\textwidth]{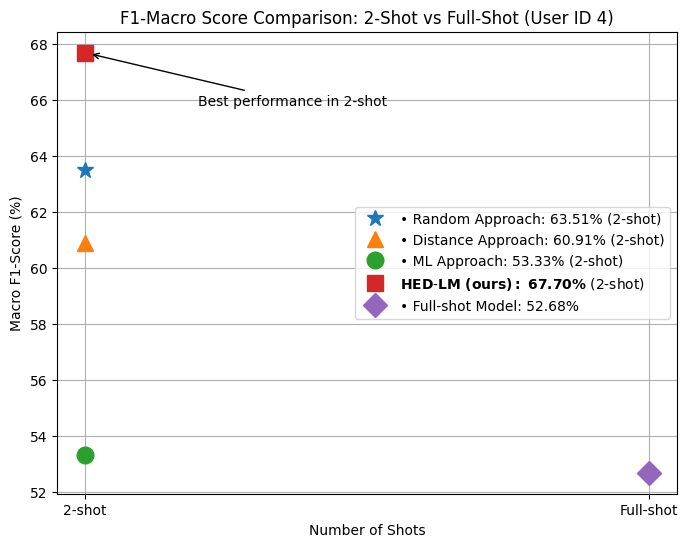}
\end{center}
\vspace{-1em}
\caption{\textbf{Macro F1-score comparison for different methods using 2-shot and full-shot setups on User ID 4}. HED-LM (2-shot) delivers the best performance, outperforming even the full-shot variant, which suffered from reduced accuracy.}
\label{fig:comp_with_full_shot}
\vspace{-2mm}
\end{figure}

As shown in Fig.~\ref{fig:comp_with_full_shot}, HED-LM in the 2-shot setting achieved the highest macro F1-score of 67.70\% for User ID 4, outperforming not only other 2-shot baselines (Random: 63.51\%, Distance: 60.91\%, ML: 53.33\%) but also the full-shot variant of HED-LM itself, which yielded only 52.68\%. This finding highlights that using more examples does not always translate into better performance; the added complexity in full-shot prompting may introduce noise or redundancy that hinders LLM inference.

\begin{figure}[H]
\begin{center}
 \includegraphics[width=0.7\textwidth]{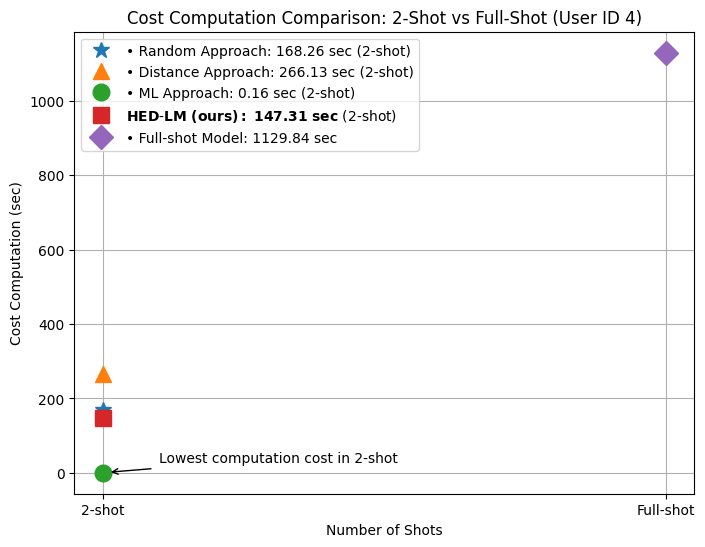}
\end{center}
\vspace{-1em}
\caption{\textbf{Computation time comparison between 2-shot and full-shot configurations for different methods on User ID 4}. HED-LM (2-shot) achieves the lowest computation cost among LLM-based methods, with significantly lower runtime than the full-shot variant.}
\label{fig:comp_fullshot}
\vspace{-2mm}
\end{figure}

It is also important to clarify that this comparison isolates the inference stage only. The computation times reported in Fig.~\ref{fig:comp_fullshot} do not include the time spent on example selection for HED-LM. This design decision ensured a fair and consistent comparison across all methods. Since the baseline approaches, Random, Distance, and Full-shot, do not involve any selection or ranking process; their execution begins directly at the inference phase. Accordingly, the inference time measured for HED-LM excludes preprocessing overhead, making the comparison focused solely on prompt execution efficiency.

Regarding computation cost in Fig.~\ref{fig:comp_fullshot}, 2-shot HED-LM was significantly more efficient, requiring only 147.31 seconds, far below the full-shot variant, which required 1129.84 seconds. This makes the 2-shot configuration almost 7.7 times faster than the full-shot, demonstrating a substantial computational advantage. The traditional machine learning (ML) model demonstrated the fastest computation time across all approaches. ML models operate entirely locally and do not involve external calls to large language models APIs. In contrast, all prompt-based methods (HED-LM, Random, Distance) depend on interaction with LLMs APIs during inference, which introduces additional latency. Although this overhead is expected, it reflects a realistic deployment scenario for LLM-integrated systems.

These results show that 2-shot prompting performs better classification while dramatically reducing inference time. The 2-shot configuration balances efficiency and effectiveness and proves more scalable for real-time or resource-constrained deployment. For this reason, we adopt 2-shot prompting as the standard setup in all subsequent experiments.

\subsubsection{Per-User Performance Evaluation}
\label{user-eval}
\begin{table}[t]
    \centering
    \footnotesize
    \caption{\textbf{Macro F1-score evaluation: Comparison between the proposed HED-LM approach and baseline methods}. The results illustrate the superior performance of HED-LM, demonstrating its effectiveness in achieving higher macro F1-scores compared to the baseline approaches. Bold values indicate the best performance for each comparison.}
     \label{tab:ex_results}
    \begin{tabular}{l|ccccc}
        \toprule
        \multirow{4}{*}{User-ID} & {\textbf{ML Approach}} & {\textbf{Random Approach}}  & {\textbf{Distance Approach}}  & \multicolumn{2}{c}{\textbf{HED-LM (Ours)}}\\
        \cmidrule(lr){2-4} \cmidrule(lr){5-6}
        & (\%) &  (\%) &  (\%) &  \multicolumn{2}{c}{(\%)}  \\
         & & & & \textbf{\begin{tabular}[c]{@{}c@{}}\#\text{ParamA}\end{tabular}} & \textbf{\begin{tabular}[c]{@{}c@{}}\#\text{ParamB}\end{tabular}}  \\
        \midrule
        4  &53.92  &59.82    &62.89       & \textbf{67.70}   & 64.42    \\
        5 &70.51   & 85.15   & 88.30  & 87.23  & \textbf{88.83}   \\
        6 &69.22  & 67.61  & 71.04  & \textbf{72.45}  & 71.57     \\
        7  &59.97  & 49.93 & 62.26  & \textbf{65.14}     & 61.93    \\
        8  &52.57   &  68.90  & 79.40  & \textbf{83.19}  & 81.87     \\
        9 &43.52 &61.64 &65.81 & 65.25 & \textbf{66.19}    \\  
        10 &19.10 &62.69 &\textbf{90.51} & 89.88 & 86.83    \\  
        11 &52.85 &48.30 &51.55 & 57.24 & \textbf{59.49}    \\  
        12 &71.45 & 73.17 &81.67 & 83.81 & \textbf{84.25}    \\  
        13 &49.11 &49.88 &50.81 &52.76 & \textbf{58.89}   \\  
        14 &34.79 &50.95 &56.66 &\textbf{59.31} & 58.64    \\  
        15 &36.39 &52.47 &57.40 &\textbf{60.85} & 59.37    \\  
        17 &45.46 & 47.95 &59.73 &\textbf{59.97} & 59.18    \\  
        18 &39.99 &62.91 &63.79 &63.00 & \textbf{64.08}   \\  
        19 &74.83 & 44.11 &73.20 &\textbf{74.90} & 70.93    \\  
        20 &70.46 &61.36 &75.43 &76.68 & \textbf{77.57}    \\  
        21 &60.01 &63.02 &64.93 &64.93 & \textbf{65.28}   \\ 
        22 &37.07 &58.30 &61.46 &\textbf{63.11} & 59.63    \\  
        23 &17.32 &58.49 &67.70 &66.11 & \textbf{68.09}    \\  
        \bottomrule
        (Mean $\pm$ SD)  &(50.4 $\pm$ 17.13) &(59.30 $\pm$ 10.13) &(67.61 $\pm$ 11.39) &(\textbf{69.13 $\pm$ 10.71}) & (68.79 $\pm$ 10.24)    \\  
        \bottomrule
    \end{tabular}
\end{table}
To better understand how our method performs across different individuals, we detailedly evaluated the macro F1-score per User ID. This per-user analysis is critical in fatigue detection using accelerometer data, where physiological and behavioral variability across users can significantly affect signal patterns. Each subject may exhibit unique movement dynamics and fatigue responses, and thus, evaluating the macro F1-score individually allows us to assess the adaptability and generalization capability of the model in real-world settings.

By presenting macro F1 scores per user, we aim to evaluate both the robustness and fairness of the HED-LM framework. This approach also helps identify whether the few-shot prompting mechanism effectively adapts to unseen users using only a small number of examples. Table~\ref{tab:ex_results} presents the macro F1-score comparison of four different approaches, such as Traditional Machine Learning (ML), Random, Distance, and our proposed HED-LM with two parameter configurations: \#ParamA ($distance_K=5$, $top_K=3$) and \#ParamB ($distance_K=10$, $top_K=5$). The results are reported as percentages for each User ID, highlighting that HED-LM consistently outperforms the baseline methods. 

One of the most prominent findings is observed for User ID 4, where HED-LM with 2-shot prompting (\#ParamA) achieved a macro F1-score of 67.70\%, outperforming all baseline approaches (Random: 59.82\%, Distance: 62.89\%, ML: 53.92\%). This suggests that HED-LM can effectively align semantic relevance and numerical similarity even with a minimal prompt to enhance classification for users with moderately distinguishable fatigue patterns.

In contrast, User ID 10 presents an unusual scenario where the traditional ML approach drastically underperforms (only 19.10\%), whereas all prompt-based methods achieve much higher scores (HED-LM: 89.88\%, Distance: 90.51\%). This highlights the LLM's ability to generalize from a few examples, even in cases where classical models struggle due to data imbalance or noise sensitivity.

A significant gain is also seen in User ID 8, where HED-LM (\#ParamA: 83.19\%) substantially outperforms the ML approach (52.57\%) and other baselines. This illustrates the benefit of semantic filtering for users with high inter-class confusion, where fatigue and non-fatigue signals may have overlapping statistical features but clearer contextual signals.

Meanwhile, User ID 17 shows a relatively stable yet modest improvement, with all methods clustered around mid-range scores (HED-LM: $\sim$59\%). This suggests that model performance may plateau regardless of selection strategy in certain users with ambiguous or low-quality sensor signals, indicating a potential limitation of LLM-based prompting without contextual enrichment.

Interestingly, for User ID 13, all methods perform similarly poorly ($\sim$50\%), indicating that prompt optimization has a limited effect when the signal characteristics are inherently weak or indistinct. This case supports the idea that data quality remains a bottleneck that even LLM-based reasoning cannot fully overcome.

Finally, on average, HED-LM (\#ParamA) achieved the highest macro F1-score across all users (69.13 $\pm$ 10.71\%), outperforming Random (59.30 $\pm$ 10.13\%), Distance (67.61 $\pm$ 11.39\%), and ML (50.4 $\pm$ 17.13\%). These findings further reinforce that the hybrid strategy in HED-LM, combining numerical filtering and semantic scoring, is consistent across users and resilient to inter-subject variability. Additionally, HED-LM exhibits a lower standard deviation than ML across users, suggesting improved stability and fairness in subject-level prediction. More details about the confusion matrix of this performance comparison are shown in the Appendix~\ref{apd:conf_mat}.

\subsubsection{Contribution of Domain Knowledge in Prompting}
\label{contr-dom}
\begin{figure}[ht]
\begin{center}
 \includegraphics[width=0.8\textwidth]{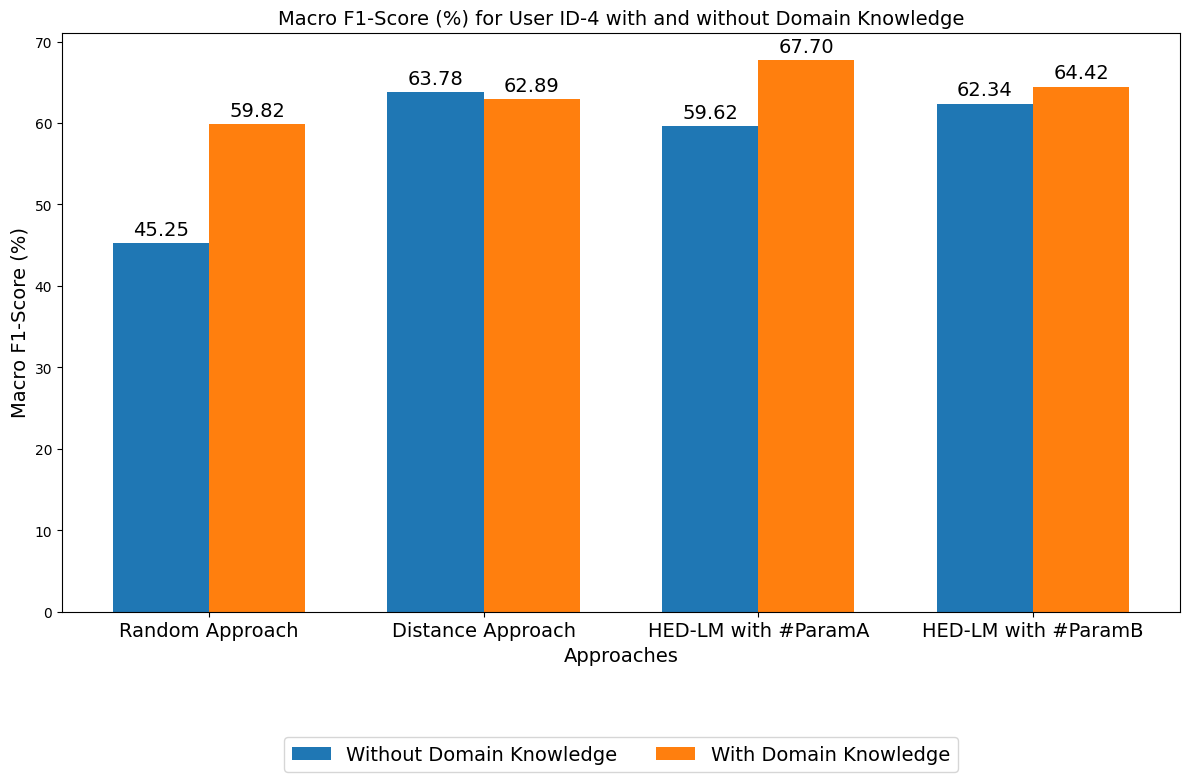}
\end{center}
\vspace{-1em}
\caption{\textbf{Macro F1-score performance comparison: Influence of domain knowledge on User-ID 4}. The results highlight the impact of incorporating domain knowledge, showcasing improved performance for User-ID 4 compared to scenarios without domain knowledge integration.}
\label{fig:comp-dom}
\vspace{-2mm}
\end{figure}

In addition to comparing the performance of the HED-LM approach against the baseline methods (Random, Distance, and ML Approach), we also analyze the effect of domain knowledge on our approach. In this experiment, we measured performance with and without domain knowledge. Without domain knowledge, LLM only assesses numerical data without applying relevant threshold rules. For instance, Fig.`\ref{fig:comp-dom} compares these results for User-ID 4, where the highest increase value is (+14.6\%) in the random approach, and the lowest is (+2.1\%) in HED-LM with \#ParamB.

The results show that using domain knowledge significantly improves the accuracy of the HED-LM approach. With domain knowledge, the HED-LM approach achieves a performance of 67.70\% for \#ParamA and 64.42\% for \#ParamB, while without domain knowledge, the performance decreases to lower (-8.08\%) and (-2.08\%), respectively. This shows that domain knowledge helps LLM understand the numerical context more deeply, resulting in more accurate predictions. In contrast, without domain knowledge, the performance of the baseline approaches, such as the Random Approach (45.25\%) and Distance Approach (63.78\%), where the random approach with domain knowledge is better, but the distance approach with domain knowledge is less good. Although the distance approach with domain knowledge has decreased insignificantly with a difference of (-0.89\%) when compared to the performance of the distance approach without domain knowledge, it can be seen that our approach (HED-LM) is still better than the distance approach without domain knowledge. In addition, compared to other baselines, such as the ML Approach, the performance of HED-LM with domain knowledge is far superior (67.70\% vs. 53.92\%). This indicates that relying solely on machine learning algorithms without the support of domain knowledge is not enough to achieve optimal results, especially on data with complex structures.

Using domain knowledge improves performance and provides higher reliability in various usage scenarios, as seen in User-ID 4. This shows that an approach based on combining LLM and domain knowledge can be a convenient solution to improving performance on complex tasks. Appendix~\ref{apd:conf_mat} explains more details about the differences in the confusion matrix when comparing the influence of domain knowledge.

%%%%%%%%%%%%%%%%%%%%%%%%%%%%%%%%%%%%%%%%%%
\section{Discussion}
\label{discussion}
\subsection{Comparative Performance Overview}
\begin{figure}[ht]
\begin{center}
 \includegraphics[width=1\textwidth]{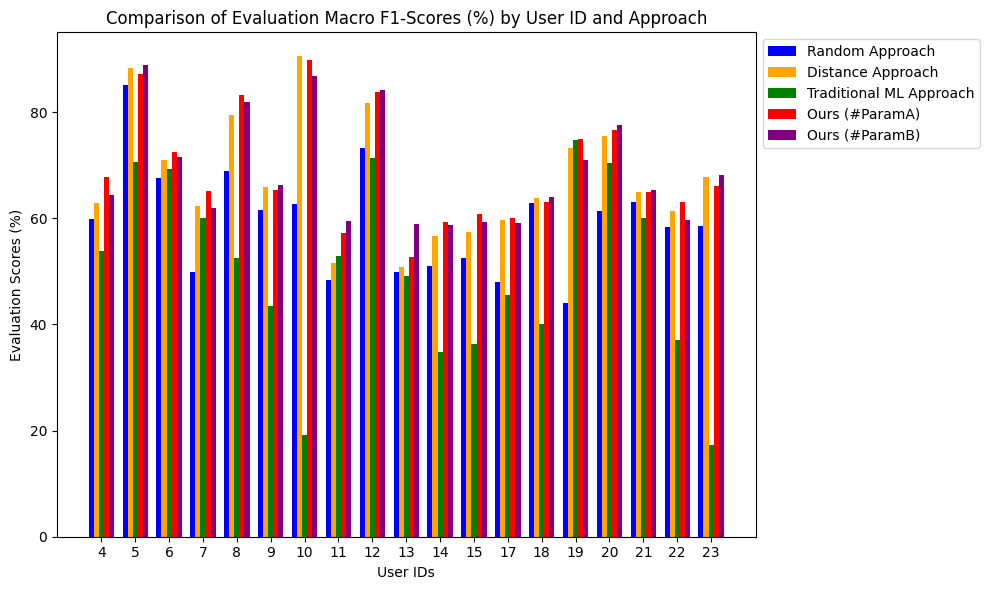}
\end{center}
\vspace{-1em}
\caption{\textbf{Overall user comparison: HED-LM and its parameters versus baselines}. The comparison evaluates the performance of the proposed HED-LM framework and its key parameters across all users, demonstrating its superiority over baseline methods in terms of effectiveness and consistency.}
\label{fig:all-dom}
\vspace{-2mm}
\end{figure}

In this section, we discuss the experimental results in more detail. We show that HED-LM (Hybrid Euclidean Distance-Language Model) generally provides higher macro F1-score performance than the three primary baselines, namely the random approach, the distance approach, and traditional machine learning (ML), as shown in Fig.~\ref{fig:all-dom}. This finding confirms that the distance-based filtering process followed by LLM scoring and few-shot prompting can utilize numeric synergy (distance) and label synergy (domain knowledge-based LLM reasoning) to improve the accuracy of melt detection.

First, we compare HED-LM with the random approach that randomly selects few-shot examples without considering numeric data similarity. The results show that the random approach is prone to giving examples of irrelevant labels, especially when the variation of sensor data is large enough, resulting in a lower macro F1-score. Meanwhile, HED-LM uses Euclidean distance to select the closest subject and then refines it through LLM scoring, which prioritizes subjects with aligned labels (fatigue/non-fatigue) according to the numeric pattern of the new subject. Based on the observations from our experiments, HED-LM has a performance improvement of about (+1.17\%)$\sim$(+30.79\%) compared to the random approach, with the lowest improvement for User-ID 18 and the highest for User-19.

Second, compared to the distance approach, the HED-LM method stands out because it adds a layer of “label synergy” in the LLM scoring. The distance approach only relies on numeric similarity, so “similar” but mislabeled subjects can be included in the list of few-shot examples. The HED-LM downgrades such subjects through the LLM relevance score, resulting in a higher macro F1-score. However, we found one case (e.g., User-ID 10) where the distance approach was better than HED-LM. Our analysis shows that in that subject, the numeric data of the new subject is ambiguous. The domain knowledge is not fully applicable (e.g., RMS and mean are in the ambiguous boundary range). Hence, LLM tends to give the old subject a “middle” relevancy score, which is numerically close but has an incorrect label.

On the other hand, the distance approach successfully places the closest distance subject with the correct label, so the classification on User-ID 10 performs better with the distance approach. Cases like this emphasize the importance of calibrating domain knowledge to translate numeric data “in the gray range” more decisively. Based on the experimental results, we see that HED-LM can improve the performance by about (+0.24\%)$\sim$(+8.08\%), where the lowest improvement is for User-ID 17 and the highest is for User-ID 13. 

Third, HED-LM provides more adaptive outcomes than traditional machine learning (ML). ML models (e.g., Random Forest) effectively extract global regularities in data features. However, the few-shot prompting in HED-LM allows the LLM to more personally assess new subjects with the most relevant examples, where domain knowledge emphasizes specific thresholds (e.g., “mean lower than 0.31 then fatigue-like”). This LLM reasoning adds a layer of interpretation that traditional training-based ML, significantly if movement intensity varies (beginning vs. end of the segment). As a result, HED-LM has a significant performance improvement over the ML Approach of about (+0.07\%)$\sim$(+70.78\%), which is lowest from User-ID 19 and highest from User-ID 10.

We also observed two parameter configurations, namely \#ParamA ($distance_k=5$, $top_k=3$) versus \#ParamB ($distance_k=10$, $top_k=5$). In \#ParamA, the system only calls LLM for the five closest candidates, then selects the best three after re-ranking. This approach saves overhead and is suitable if the dataset has relatively straightforward distances between subjects. However, if the dataset tends to be more varied and many subjects are “on the border,” \#ParamB can increase the coverage by scoring 10 candidates and finally selecting the top 5, potentially resulting in a better macro F1-score as subjects that are “not too close in proximity but have strong label synergy” can be accommodated. Of course, \#ParamB also increases the overhead of LLM calls by almost double. In our evaluation, there are cases where \#ParamA efficiently yields high results when the numeric features of the dataset are more stable. At the same time, \#ParamB is useful when the data displays various borderline patterns, so the wider coverage helps LLM re-ranking filter out truly suitable label candidates. Based on our experimental analysis, when comparing parameters, the highest increase (+9.78\%) occurs with \#ParamB at User-ID 13 and the lowest (+0.35\%) with \#ParamB at User-ID 21.

\begin{table}[ht]
\centering
\footnotesize
\caption{Nemenyi post-hoc test results between Random, Distance, ML, HED-LM \#ParamA, and HED-LM \#ParamB methods. The values in the table are the exact comparison p-values without using scientific notation.}
\label{tab:posthoc_nemenyi}
\begin{tabular}{lccccc}
\hline
              & Random   & Distance   & ML           & HED-LM \#ParamA       & HED-LM \#ParamB       \\ \hline
Random        & 1.000000 & 0.007625   & 0.972672     & 0.00001085355  & 0.00002331567  \\
Distance      & 0.007625 & 1.000000   & 0.0007448186 & 0.5369701     & 0.6372963     \\
ML            & 0.972672 & 0.000745   & 1.000000     & 0.0000004028977 & 0.0000009506715 \\
HED-LM \#ParamA      & 0.000011 & 0.536970   & 0.0000004028977 & 1.000000   & 0.9998743     \\
HED-LM \#ParamB      & 0.000023 & 0.637296   & 0.0000009506715 & 0.9998743  & 1.000000      \\ \hline
\end{tabular}
\end{table}
To test the significance of performance differences among the five methods (Random, Distance, ML, HED-LM \#ParamA, HED-LM \#ParamB), we performed Friedman test on the macro F1-score results for 19 subjects. The test results show that the difference is significant (F-statistic = 54.55, p-value < 0.0001), indicating that at least one method performs significantly differently from the other methods. In order to find out which method pairs were significantly different, we proceeded with the Nemenyi post-hoc. The p-value table of post-hoc results shown in Table~\ref{tab:posthoc_nemenyi} indicates some important findings:
\begin{itemize}
    \item Random vs. Distance (p = 0.0076) and Random vs. HED-LM (p < 0.0001) show that the Random method has a significant difference compared to Distance, HED-LM \#ParamA, and HED-LM \#ParamB.
    \item Random vs. ML (p = 0.9727) showed no significant difference, implying that both methods are relatively less accurate than the other groups.
    \item Distance vs. ML (p = 0.0007) was significantly different, highlighting that Distance was clearly better than ML.
    \item Distance vs. HED-LM \#ParamA /HED-LM \#ParamB (p > 0.05) was not significantly different; neither was HED-LM \#ParamA vs. HED-LM \#ParamB (p $\approx$ 1.0). This indicates that the three methods (Distance, HED-LM \#ParamA, and HED-LM \#ParamB) are potentially in roughly equal performance clusters.
    \item ML vs. HED-LM \#ParamA/HED-LM \#ParamB (p < 0.0001) is significantly different, indicating HED-LM \#ParamA and HED-LM \#ParamB are statistically superior to ML.
\end{itemize}
From these post-hoc results, it can be concluded that:
\begin{itemize}
    \item Random and ML belong to lower-performing or unstable clusters; they are not significantly different from each other, but clearly differentiate against other better methods.
    \item Distance, HED-LM \#ParamA, and HED-LM \#ParamB all form high-performance clusters that are not significantly different from each other. A p-value > 0.05 for each pair indicates there is no statistical evidence that one is truly superior.
    \item Overall, these results validate our initial findings-that both hybrid methods (HED-LM \#ParamA and HED-LM \#ParamB) and Distance Approach perform better than Random or ML.
\end{itemize}
Thus, the Friedman and post-hoc Nemenyi tests confirm that the improved macro F1-score results in the Distance and HED-LM variants are not a statistical fluke, but rather represent real performance differences.

After confirming that there were significant differences between methods (Friedman test) and knowing which pairs of methods differed (post-hoc Nemenyi), we calculated Cliff's Delta ($\delta$) to assess how large (practical effect) the differences between two specific methods were. Cliff's Delta is a non-parametric metric that does not require the assumption of normality, as per our data condition (macro F1-score in 19 subjects). Basically, $\delta$ is in the range of -1 to +1:
\begin{itemize}
    \item $\delta$ > 0 means the method to the left of “vs.” (e.g. “X vs. Y”) tends to be superior.
    \item $\delta$ < 0 means the method to the right of “vs.” is superior.
    \item $\vert\delta\vert\geq$ 0.47-0.50 is often interpreted as a large effect, implying substantial differences in practice.
\end{itemize}
Our pairwise results confirm the existence of two practically distinct performance clusters:
\begin{enumerate}
    \item Superior Cluster: Distance, HED-LM \#ParamA, and HED-LM \#ParamB
    \begin{itemize}
        \item Cliff's Delta shows values close to zero when these three methods are compared against each other. For example, Distance vs. HED-LM \#ParamA ($\delta$ = -0.097) and Distance vs. HED-LM \#ParamB ($\delta$ = -0.053), which are each classified as negligible effects ($\vert\delta\vert$ < 0.1). Similarly, HED-LM \#ParamA vs. HED-LM \#ParamB ($\delta$ = 0.042), so there is no strong indication that one of the three stands out significantly from the others.
        \item In other words, within this superior cluster, the performance of Distance, HED-LM \#ParamA, and HED-LM \#ParamB were relatively balanced according to effect size, in line with the post-hoc results which also stated that they were not significantly different.
    \end{itemize}
    \item Lower Cluster: Random and ML
    \begin{itemize}
        \item The inter-method comparison in this cluster also shows relatively close results (Random vs. ML: $\delta$ = 0.274, small-medium effect), indicating Random is slightly better but not too far behind ML.
        \item It is when these two methods are compared with the methods in the superior cluster that large effects are seen. For example, Random vs. HED-LM \#ParamA ($\delta$ = -0.524) and Random vs. HED-LM \#ParamB ($\delta$ = -0.485) confirm that HED-LM \#ParamA and HED-LM \#ParamB are far superior to Random. In addition, Distance vs. ML ($\delta$ = 0.562) illustrates a significant difference with a large effect to support the superiority of Distance. A value of $\vert\delta\vert$ above 0.47 indicates a truly different performance in practical terms, not just statistically significant.
    \end{itemize}
\end{enumerate}

Based on the overall effect size, we can conclude that the Distance, HED-LM \#ParamA, and HED-LM \#ParamB methods form one high-performing and mutually equivalent group, while Random and ML are in the lower-performing group-with often “large” effect differences when compared against the superior group. This result is consistent with the previous statistical analysis (Friedman test and Nemenyi post-hoc) which suggested that although globally there are significant differences, the “top cluster” methods (Distance, HED-LM \#ParamA, HED-LM \#ParamB) are difficult to distinguish from each other convincingly. Thus, Cliff's Delta provides strong evidence that the performance gap between the top cluster and the lower clusters is substantial, while affirming the stability of performance within the top cluster itself.

\begin{table}[ht]
\centering
\caption{Statistical test results (p-value) for performance comparison between HED-LM \#ParamA and HED-LM \#ParamB methods with Random, Distance, and ML methods.}
\label{tab:results_test}
\begin{tabular}{lcc}
\hline
Comparison              & t-test p-value & Wilcoxon p-value \\ \hline
HED-LM \#ParamA vs Random        & 0.00004        & 0.00000          \\
HED-LM \#ParamA vs Distance      & 0.02716        & 0.00240          \\
HED-LM \#ParamA vs ML            & 0.00006        & 0.00002          \\
HHED-LM \#ParamB vs Random        & 0.00005        & 0.00000          \\
HED-LM \#ParamB vs Distance      & 0.09976        & 0.06021          \\
HED-LM \#ParamB vs ML            & 0.00009        & 0.00004          \\ \hline
\end{tabular}
\end{table}

Meanwhile, based on the p-value table in Table~\ref{tab:results_test} for the six pairwise comparisons between HED-LM \#ParamA/B and the other three methods (Random, Distance, ML), it appears that HED-LM \#ParamA is consistently significantly different from Random (p < 0.001 in both tests) and ML (p < 0.001), and also significant when compared to Distance (p < 0.05 for both the t-test and Wilcoxon). This shows that HED-LM \#ParamA is statistically superior in all these comparisons. In contrast, HED-LM \#ParamB was also clearly better than Random and ML (p < 0.001), but showed no significant difference when compared to Distance (t-test p=0.09976; Wilcoxon p=0.06021), both of which exceeded the $\alpha$ = 0.05 threshold. Thus, these test results indicate that HED-LM \#ParamB is equivalent to Distance, there is no statistical evidence that they are different, whereas HED-LM \#ParamA is significantly different (and likely superior) to Distance. Overall, both HED-LM approaches were shown to be superior to both Random and ML, but only HED-LM \#ParamA displayed a significant difference to Distance according to this test data.

Embedding thresholds (mean, std, RMS) in the LLM messaging system increases the sensitivity of the relevancy score. We observed that when domain knowledge is disabled, LLM scoring tends to “blend” (score around 0.4-0.6), decreasing re-ranking power. However, if domain knowledge is available and aligned with the data, LLM does not hesitate to give extreme scores (e.g., 0.1 or 0.9). As a result, old subjects whose labels and features perfectly match the new subjects jump to the top of the rankings. However, the case of User-ID 10 shows that inaccurate domain knowledge, especially if the subject's numeric data is in the vulnerable zone, can negatively impact LLM scores.

The overall results of this study support our initial goal of optimizing few-shot prompting for fatigue detection in complex sensor data by combining numerical analysis (distance) and label synergy reasoning (LLM). Most subjects experienced significant macro F1-score improvement when domain knowledge was included, indicating that the HED-LM method has broad potential to detect fatigue vs. non-fatigue with precision. However, we also noticed that there are “anomalous” cases, such as User-ID 10, that should be further investigated, both in terms of the validity of the knowledge domain and the adjustment of the distance parameter. Thus, although HED-LM is generally superior, there are still opportunities to update the domain knowledge and adapt the scoring mechanism to make the method more robust in various sensor data conditions.

\subsection{Limitations and Future Work}
\noindent\textbf{Alternative distance metrics}. Euclidean distance was chosen because it runs in only $\mathcal{O}(d)$ time, is easy to interpret geometrically, and is still the most common baseline in recent surveys of time‑series similarity measures \cite{Chen2020InfoSciSurvey}.  Its main limitation is that it treats the feature space as isotropic, ignoring directional information and any correlation between features.  Cosine similarity addresses the first issue by focusing on angular alignment, whereas Mahalanobis distance addresses the second by weighting each dimension with the inverse covariance matrix.  Computing the Mahalanobis form, $(x-\mu)^{\top}\Sigma^{-1}(x-\mu)$, requires a matrix–vector multiplication and thus scales quadratically with the feature dimension $d$, making it noticeably slower than Euclidean.  Even so, \cite{Chen2020InfoSciSurvey} reports several biomedical applications where Mahalanobis outperforms Euclidean and cosine because the covariance structure carries important class information.  We therefore plan a systematic ablation study of Euclidean, cosine, and Mahalanobis distances in future work.  A longer‑term goal is to develop an adaptive, data‑aware metric‑selection module that can trade off the potential accuracy gains of richer metrics against their higher computational cost.

\noindent\textbf{Window‑Segmentation Strategy}. HED‑LM currently divides each 180‑sample magnitude trace into three fixed windows.  A rigid scheme guarantees identical feature dimensions across subjects and keeps the numeric summary compact enough for the LLM token budget.  The downside is that physiological transitions seldom align with rigid boundaries; class-specific cues can be diluted if fatigue onset falls inside a window. Adaptive segmentation has therefore become an active research topic. Truong et al. \cite{Truong2020ReviewCPD} provide a comprehensive review of recent change‑point detection (CPD) algorithms that identify statistical shifts in multivariate biosignals.  CPD‑driven windowing can dynamically resize segments around behavioural changes and has been shown, across multiple studies cited in that review, to improve activity‑ and health‑state recognition without excessive computational overhead. Dynamic time‑warping alignment is another adaptive alternative, but its quadratic cost makes real‑time deployment more challenging. To balance fidelity and efficiency, the next version of HED‑LM will adopt a two‑stage design: a lightweight CPD routine will first mark coarse break‑points; very short segments will then be merged so that no more than two examples are passed to the LLM.  This pipeline preserves temporal nuance without inflating prompt length or latency, remaining compatible with edge‑deployment constraints.

\noindent\textbf{Hyper‑Parameter Sensitivity}.
Section \ref{dist-top_k} shows that macro‑F1 varies with the values of \textit{distance‑K} and \textit{top‑K}. Large swings caused by example selection are also reported for in‑context learning \cite{Min2022RethinkICL}. A small $K$ risk omits informative neighbours, whereas a large $K$ increases latency and may introduce noise. We therefore plan to add an \emph{adaptive‑$K$} controller that expands the candidate pool until the local distance entropy stabilises, similar in spirit to the adaptive distance‑weighted $k$‑NN proposed by W. Xue et al. \cite{XUE2025113171}. This mechanism should reduce manual tuning and lower the risk of over‑fitting to personal subjects.

\noindent\textbf{Feature‑Extraction Scope}. Our present features, mean, RMS, skewness, and dominant frequency, are token‑efficient and interpretable but may miss non‑linear or multi‑scale patterns linked to fatigue.  Wang et al. \cite{s21196369} showed that wavelet‑packet energy features boost wearable‑sensor fatigue detection by 7\% over raw statistics. Transformer‑based embeddings capture long‑range dependencies in multivariate time series \cite{Zhou2021Informer}.  Integrating such rich features, followed by dimensionality reduction to fit the LLM prompt budget, is a promising avenue for the next version of HED‑LM.

\noindent\textbf{Comparison with Other Few‑Shot Paradigms}. HED‑LM is training‑free and contrasts with meta‑learning approaches such as MAML.  Hospedales et al. \cite{Hospedales2021MetaSurvey} review these methods and note their computational footprint and need for labelled support sets, constraints that conflict with our edge deployment scenario.  Parameter‑efficient prompt tuning \cite{Lester2021PromptTuning} and retrieval‑augmented generation \cite{Lewis2020RAG} offer a middle ground, requiring only a small set of frozen LLM parameters or an external memory.  Exploring these hybrids with HED‑LM is an important direction for future work.

\noindent\textbf{Broader Applicability and Domain Extension}. While this study adopts fatigue detection from unimodal accelerometer data as a representative use case, the proposed HED-LM framework is designed to be modular and domain-agnostic. It can be extended to other temporal signal classification problems that rely on physiological sensors such as heart rate, skin temperature, and electromyography, enabling multimodal integration in healthcare and industrial settings. Beyond fatigue, this approach can support tasks such as stress detection, sleep stage recognition, or ergonomic risk classification, where numerical features and contextual reasoning are essential. Future work will involve applying HED-LM across diverse domains and signal modalities to evaluate its effectiveness under more complex, real-world conditions. These cross-domain explorations will help identify scenarios where semantic reasoning from LLMs provides significant gains over purely numerical similarity measures.

%%%%%%%%%%%%%%%%%%%%%%%%%%%%%%%%%%%%%%%%%%
\section{Conclusion}
\label{conclusion}
In this study, we have proposed HED-LM (Hybrid Euclidean Distance-Language Model) for fatigue detection in accelerometer data. Through comprehensive experiments, HED-LM consistently outperforms baselines such as random approach, distance-only, and traditional machine learning (ML), especially in the F1-score macro metric. The success of HED-LM is supported by two important mechanisms: (1) distance-based filtering that selects numerically closest subjects, and (2) LLM scoring that combines domain knowledge (threshold mean, RMS, etc.) to assess the suitability of fatigue/non-fatigue labels more semantically. Finally, candidate re-ranking and few-shot prompting ensure new subjects acquire the most relevant examples, resulting in more accurate mapping of sensor data to final labels.

On the other hand, the results also highlighted some aspects that require further attention. For example, we observed one case (User-ID 10) where the distance approach outperformed HED-LM-this emphasizes the need for more precise calibration of domain knowledge, especially when the subject's numeric data is in the gray zone (the range between “fatigue-like” vs. “non-fatigue-like”). In addition, the selection of the parameters $distance_k$ and $top_k$ also significantly impacts the efficiency and coverage of candidates assessed by the LLM, requiring a balance between the overhead of LLM calls and the potential performance improvement.

In addition to finding a consistent improvement in the macro F1-score, we also conducted statistical analysis to validate the performance differences between the methods. The Friedman test results (p-value < 0.01) confirmed that at least one method was significantly superior to the others. Post-hoc Nemenyi showed that the HED-LM approach (both \#ParamA and \#ParamB) formed a superior group significantly different from Random and traditional ML methods. Effect size measurements (Cliff's Delta or Kendall's W) support these findings with large effect category values, indicating that the superiority of HED-LM is substantial in practical terms, not just a statistical fluke. Therefore, the performance improvement we obtained is superior in metrics (macro F1-score) and proves to have significant differences and noticeable effects in the context of applying few-shot prompting on accelerometer data. Going forward, The outcomes of this statistical analysis can provide a strong basis for additional development, such as integrating more adaptive domain knowledge and evaluation in other domains that require robust few-shot classification.

For future work, We see several development opportunities to make the HED-LM approach more effective and widely applicable. First, the knowledge domain can be enriched by adding more holistic rules, such as including scenarios of varying movement intensity or emphasizing the analysis of intermediate frequencies that may affect fatigue patterns. Secondly, evaluating more diverse datasets ranging from light intensity to high load can gauge the method's adaptability under more extreme real-world conditions. Thirdly, we consider fine-tuning LLM to handle numeric data more optimally, reducing reliance on manual prompts and deepening the analysis of quantitative feature differences (e.g., mean, RMS, kurtosis). Finally, future research could include an adaptive mechanism for the parameters $distance_k$ and $top_k$ so that the system automatically balances the candidate coverage and computational overhead of LLM according to the data characteristics. These efforts will improve the capabilities of HED-LM and expand its applicability in various fatigue detection scenarios on complex sensor data.

Overall, this study highlights the potential of HED-LM as a hybrid approach in the domain of few-shot prompting on feature-rich sensor data. By combining numerical analysis (distance) and semantic reasoning (LLM scoring), this method can be further improved in terms of efficiency, interpretability, and generalizability to support various fatigue vs. non-fatigue detection applications in the context of healthcare and other fields.

%%%%%%%%%%%%%%%%%%%%%%%%%%%%%%%%%%%%%%%%%%
\vspace{6pt} 

%%%%%%%%%%%%%%%%%%%%%%%%%%%%%%%%%%%%%%%%%%
%% optional
%\supplementary{The following supporting information can be downloaded at:  \linksupplementary{s1}, Figure S1: title; Table S1: title; Video S1: title.}

% Only for journal Methods and Protocols:
% If you wish to submit a video article, please do so with any other supplementary material.
% \supplementary{The following supporting information can be downloaded at: \linksupplementary{s1}, Figure S1: title; Table S1: title; Video S1: title. A supporting video article is available at doi: link.}

% Only used for preprtints:
% \supplementary{The following supporting information can be downloaded at the website of this paper posted on \href{https://www.preprints.org/}{Preprints.org}.}

% Only for journal Hardware:
% If you wish to submit a video article, please do so with any other supplementary material.
% \supplementary{The following supporting information can be downloaded at: \linksupplementary{s1}, Figure S1: title; Table S1: title; Video S1: title.\vspace{6pt}\\
%\begin{tabularx}{\textwidth}{lll}
%\toprule
%\textbf{Name} & \textbf{Type} & \textbf{Description} \\
%\midrule
%S1 & Python script (.py) & Script of python source code used in XX \\
%S2 & Text (.txt) & Script of modelling code used to make Figure X \\
%S3 & Text (.txt) & Raw data from experiment X \\
%S4 & Video (.mp4) & Video demonstrating the hardware in use \\
%... & ... & ... \\
%\bottomrule
%\end{tabularx}
%}

%%%%%%%%%%%%%%%%%%%%%%%%%%%%%%%%%%%%%%%%%%
\authorcontributions{Conceptualization, E.R. and S.I.; methodology, E.R.; software, E.R. and S.I.; validation, E.R. and S.I.; formal analysis, E.R.; investigation, E.R.; resources, E.R.; data curation, E.R.; writing—original draft preparation, E.R.; writing—review and editing, E.R. and S.I.; visualization, E.R.; supervision, S.I.; project administration, E.R.; funding acquisition, S.I. All authors have read and agreed to the published version of the manuscript.}

\funding{This research was funded by the JST-Mirai Program, Creation of Care Weather Forecasting Services in the Nursing and Medical Field, Grant Number JPMJMI21H3 in Japan.}

\institutionalreview{The study was conducted according to the guidelines of the Declaration of Helsinki.}

\informedconsent{Informed consent was obtained from all subjects involved in the study.}

\dataavailability{The data presented in this study is available using public datasets that can be accessed at \url{https://zenodo.org/records/7997851}.}

\acknowledgments{The authors gratefully acknowledge the support provided by the Ministry of Education, Culture, Sports, Science, and Technology (MEXT), Japan, during the course of this study. We also sincerely thank the members of the Sozo Laboratory for their invaluable assistance and collaboration throughout this research. Further appreciation is extended to Universitas 17 Agustus 1945 (Untag) Surabaya and the Ministry of Higher Education, Science, and Technology of the Republic of Indonesia for their continuous support and encouragement. Special thanks go to Enny Indasyah and Ryuga Alfatih Ernando for their unwavering support and motivation in completing this research.}

\conflictsofinterest{The authors declare no conflicts of interest.} 

%%%%%%%%%%%%%%%%%%%%%%%%%%%%%%%%%%%%%%%%%%
%% Optional

%% Only for journal Encyclopedia
%\entrylink{The Link to this entry published on the encyclopedia platform.}

\abbreviations{Abbreviations}{
The following abbreviations are used in this manuscript:\\

\noindent 
\begin{tabular}{@{}ll}
API & Application Programming Interface\\
SD & Standard Deviation\\
GPT & Generative Pre-trained Transformer\\
HED-LM & Hybrid Edit Distance - Language Model\\
KNN & K-Nearest Neighbors\\
LLM & Large Language Model\\
LLMs & Large Language Models\\
ML & Machine Learning\\
RMS & Root Mean Square
\end{tabular}
}

%%%%%%%%%%%%%%%%%%%%%%%%%%%%%%%%%%%%%%%%%%
%% Optional
\appendixtitles{no} % Leave argument "no" if all appendix headings stay EMPTY (then no dot is printed after "Appendix A"). If the appendix sections contain a heading then change the argument to "yes".
\appendixstart
\appendix
\section[\appendixname~\thesection]{}
\subsection[\appendixname~\thesubsection]{}
\label{apd:llm_score}
This section describes the process of example selection using LLM Scoring through carefully designed prompting. The first step involves creating a context incorporating feature information from each segment, paired with its corresponding label, based on examples provided by $distance_{k}$. Additionally, feature details from unlabeled subjects, predicted during the evaluation, are included in the context to enrich the scoring framework.

To ensure the scoring process is systematic and practical, we design prompts that guide the LLM in evaluating relevance at each stage. These prompts include detailed instructions and domain-specific knowledge to help the model follow a structured scoring methodology. The questions are presented progressively, enabling the LLM to generate thoughtful and well-aligned responses to the assessment criteria.

Throughout the LLM Scoring process, we use the GPT-4o-mini model with a temperature setting 0.3 to maintain consistency and reliability. The interplay of domain knowledge, instructions, context, and staged questioning enhances the overall effectiveness of the scoring mechanism. A detailed illustration of the prompting design for this process is provided in Fig.~\ref{fig:LLM-Scoring}.
\tcbset{colback=white, colframe=black, fonttitle=\bfseries}
\begin{tcolorbox}[title=LLM Scoring Prompting, breakable]
\textbf{\#Domain Knowledge}: \texttt{\{domain\_knowledge\}}\\
\textbf{\#Instruction}: Perform step-by-step numeric comparison internally, don't reveal it.\\
\textbf{\#Context}:\\
\textnormal{[Labeled Subject | ID=\{id\}, Label=\{label\}]} \\
Segment1 $\Rightarrow$ \{\texttt{features}\} \\
Segment2 $\Rightarrow$ \{\texttt{features}\} \\
Segment3 $\Rightarrow$ \{\texttt{features}\} \\
\\
\textnormal{[New Subject | ID=\{id\}, Label=?]} \\
Segment1 $\Rightarrow$ \{\texttt{features}\} \\
Segment2 $\Rightarrow$ \{\texttt{features}\} \\
Segment3 $\Rightarrow$ \{\texttt{features}\} \\

[Guidance for Relevancy Scoring]:
\begin{enumerate}
    \item Check the numeric difference (Mean, Std, RMS, etc.) between segments.\\
    The smaller the difference $\Rightarrow$ higher relevance (0..1).
    \item Check if the labeled subject label (Fatigue vs Non-Fatigue) matches the numeric pattern of the new subject.\\
    For example, if the labeled subject was Fatigue and the new data appears to have fatigue characteristics (high RMS, etc.), the relevance can increase.
    \item Use range \textbf{0.0..1.0}:
    \begin{itemize}
        \item 0.0 $\Rightarrow$ very different \& context labels do not match.
        \item 1.0 $\Rightarrow$ very similar \& context labels match.
    \end{itemize}
    \item Output format:\\
    SCORE: <float>\\
    REASON: <briefly alluding to (a) numeric differences, (b) label context>\\
    Only concise, without chain-of-thought details.
\end{enumerate}

\textbf{\#Question}:\\
Please compare the numeric data segment by segment. If the differences in Mean, Std, RMS etc. are very small $\Rightarrow$ high relevance.  
Also check whether the labeled subject labels (Fatigue/Non-Fatigue) are aligned with the numeric pattern of the new subject.\\
Output format:\\
SCORE: <float 0..1>\\
REASON: <1-2 lines about numeric differences \& label context match>\\
Do not display long reasoning.

Example:\\
SCORE: 0.90\\
REASON: 'Segment1-2 very similar in RMS, label=Fatigue matches high RMS pattern.'
\end{tcolorbox}
\captionof{figure}{\textbf{Prompting design for LLM scoring}. The design incorporates several key components: (1) Domain knowledge to establish threshold values for performance enhancement, (2) Instructions to guide the LLM's focus, (3) Context to provide relevant background, and (4) Questions to elicit targeted responses.}
\label{fig:LLM-Scoring}

\subsection[\appendixname~\thesubsection]{}
\label{apd:dom_know}
We use feature analysis to integrate domain knowledge produced by an LLM to improve HED-LM's performance. In particular, we employ the GPT-4o model in the feature analysis procedure to extract domain knowledge that closely corresponds to the context of the feature values in the sensor data. This guarantees that the insights gleaned are highly pertinent and customized to the dataset's features.

Fig.~\ref{fig:domain-know-prompt} thoroughly outlines the architecture of the prompting system utilized to acquire this domain knowledge. This design is set up to let the LLM analyze the features methodically and derive insightful information. Furthermore, Fig.~\ref{fig:dom-know-res} shows an example of the LLM's answer that demonstrates how domain knowledge is created through this approach. These figures show how the approach is applied and how LLM-generated domain knowledge is incorporated into the HED-LM framework.
\tcbset{colback=white, colframe=black, fonttitle=\bfseries}
\begin{tcolorbox}[title=Design Prompting to Acquire Domain Knowledge, breakable]
\textbf{\#Attachment File}: \texttt{\{files\_with\_feature\_values\}}\\
\textbf{\#Instruction}:\\
Based on the attached file, please analyze carefully to create a domain of knowledge in the following format:

domain\_knowledge = """
The classification of fatigue and non-fatigue is determined by evaluating specific features across three segments (Seg1, Seg2, and Seg3). The key features and their thresholds are described below:

Segment 1 (Seg1):
\begin{itemize}
    \item Fatigue
    \begin{itemize}
        \item Mean Acceleration:
        \item Standard Deviation:
        \item Energy in Low Band:
        \item Skewness:
        \item Kurtosis:
    \end{itemize}
    \item Non-Fatigue:
    \begin{itemize}
        \item Mean Acceleration:
        \item Standard Deviation:
        \item Energy in Low Band:
        \item Skewness:
        \item Kurtosis:
    \end{itemize}
\end{itemize}

Segment 2 (Seg2):
\begin{itemize}
    \item Fatigue
    \begin{itemize}
        \item Energy in Low Band:
        \item Skewness:
    \end{itemize}
    \item Non-Fatigue:
    \begin{itemize}
        \item Energy in Low Band:
        \item Skewness:
    \end{itemize}
\end{itemize}

Segment 3 (Seg3):
\begin{itemize}
    \item Fatigue
    \begin{itemize}
        \item Standard Deviation:
        \item Energy in Low Band:
    \end{itemize}
    \item Non-Fatigue:
    \begin{itemize}
        \item Standard Deviation:
        \item Energy in Low Band:
    \end{itemize}
\end{itemize}

Conclusion:
\begin{itemize}
    \item Fatigue: Key features like Mean Acceleration, Standard Deviation, and Energy in Low Band tend to have lower values, supported by asymmetrical and sharp signal distributions (Skewness and Kurtosis).
  \item Non-Fatigue: Features show higher values with stable and flat signal distributions, indicating more intense and consistent energy patterns.
\end{itemize}

Key Principles:
\begin{enumerate}
    \item Evaluate each feature in all segments against the thresholds defined above.
    \item If any feature in any segment clearly meets the fatigue criteria, classify the label as "fatigue".
    \item If all features in all segments align with non-fatigue criteria, classify the label as "non-fatigue".
    \item Use a chain-of-thought process to evaluate all features collectively, but return only the final label: "fatigue" or "non-fatigue".
\end{enumerate}
"""

\#\textbf{Question:}\\
Please create domain knowledge according to the format.
\end{tcolorbox}
\captionof{figure}{\textbf{Prompting design for acquiring domain knowledge from sensor data feature analysis}. The design leverages sensor data features to extract domain-specific insights, guiding the LLM through targeted instructions, contextual information, and structured queries for effective knowledge acquisition.}
\label{fig:domain-know-prompt}

\tcbset{colback=white, colframe=black, fonttitle=\bfseries}
\begin{tcolorbox}[title=Example of LLM Response Result, breakable]
\textbf{\#LLM Answer:}\\
domain\_knowledge = """
The classification of fatigue and non-fatigue is determined by evaluating specific features across three segments (Seg1, Seg2, and Seg3). The key features and their thresholds are described below:

Segment 1 (Seg1):
\begin{itemize}
    \item Fatigue:
    \begin{itemize}
        \item Mean Acceleration: Values above 0.44 are indicative of fatigue.
        \item Standard Deviation: Values above 0.28 suggest less variability, consistent with fatigue.
        \item Energy in Low Band: Values above 650 reflect high energy expenditure, associated with fatigue.
        \item Skewness: Values above 0.75 highlight asymmetrical signal distributions often seen in fatigue patterns.
        \item Kurtosis: Values near or above -0.05 indicate sharper signal distributions, consistent with fatigue.
    \end{itemize}
    \item Non-Fatigue:
    \begin{itemize}
        \item Mean Acceleration: Values below 0.40 indicate lower activity, typical of non-fatigue.
        \item Standard Deviation: Values below 0.27 reflect greater variability in motion.
        \item Energy in Low Band: Values below 600 signify lower energy expenditure.
        \item Skewness: Values below 0.70 align with stable signal distributions.
        \item Kurtosis: Values below -0.10 suggest flatter distributions, typical of non-fatigue.
    \end{itemize}
\end{itemize}

Segment 2 (Seg2):
\begin{itemize}
    \item Fatigue:
    \begin{itemize}
        \item Energy in Low Band: Values above 700 reflect significant energy levels.
        \item Skewness: Values above 1.5 indicate highly skewed signal distributions, consistent with fatigue.
    \end{itemize}
    \item Non-Fatigue:
    \begin{itemize}
        \item Energy in Low Band: Values below 600 signify reduced energy activity.
        \item Skewness: Values below 1.2 represent stable signal distributions.
    \end{itemize}
\end{itemize}

Segment 3 (Seg3):
\begin{itemize}
    \item Fatigue:
    \begin{itemize}
        \item Standard Deviation: Values above 0.30 indicate signal stability, often linked to fatigue.
        \item Energy in Low Band: Values above 800 reflect increased energy expenditure, characteristic of fatigue.
    \end{itemize}
    \item Non-Fatigue:
    \begin{itemize}
        \item Standard Deviation: Values below 0.28 suggest greater variability in motion.
        \item Energy in Low Band: Values below 750 signify reduced energy activity.
    \end{itemize}
\end{itemize}

Conclusion:
\begin{itemize}
    \item Fatigue: Key features like Mean Acceleration, Standard Deviation, and Energy in Low Band tend to have higher values, supported by asymmetrical and sharp signal distributions (Skewness and Kurtosis).
    \item Non-Fatigue: Features show lower values with stable and flat signal distributions, indicating more consistent and balanced energy patterns.
\end{itemize}

Key Principles:
\begin{enumerate}
    \item Evaluate each feature in all segments against the thresholds defined above.
    \item If any feature in any segment clearly meets the fatigue criteria, classify the label as "fatigue".
    \item If all features in all segments align with non-fatigue criteria, classify the label as "non-fatigue".
    \item Use a chain-of-thought process to evaluate all features collectively, but return only the final label: "fatigue" or "non-fatigue".
\end{enumerate}
"""
\end{tcolorbox}

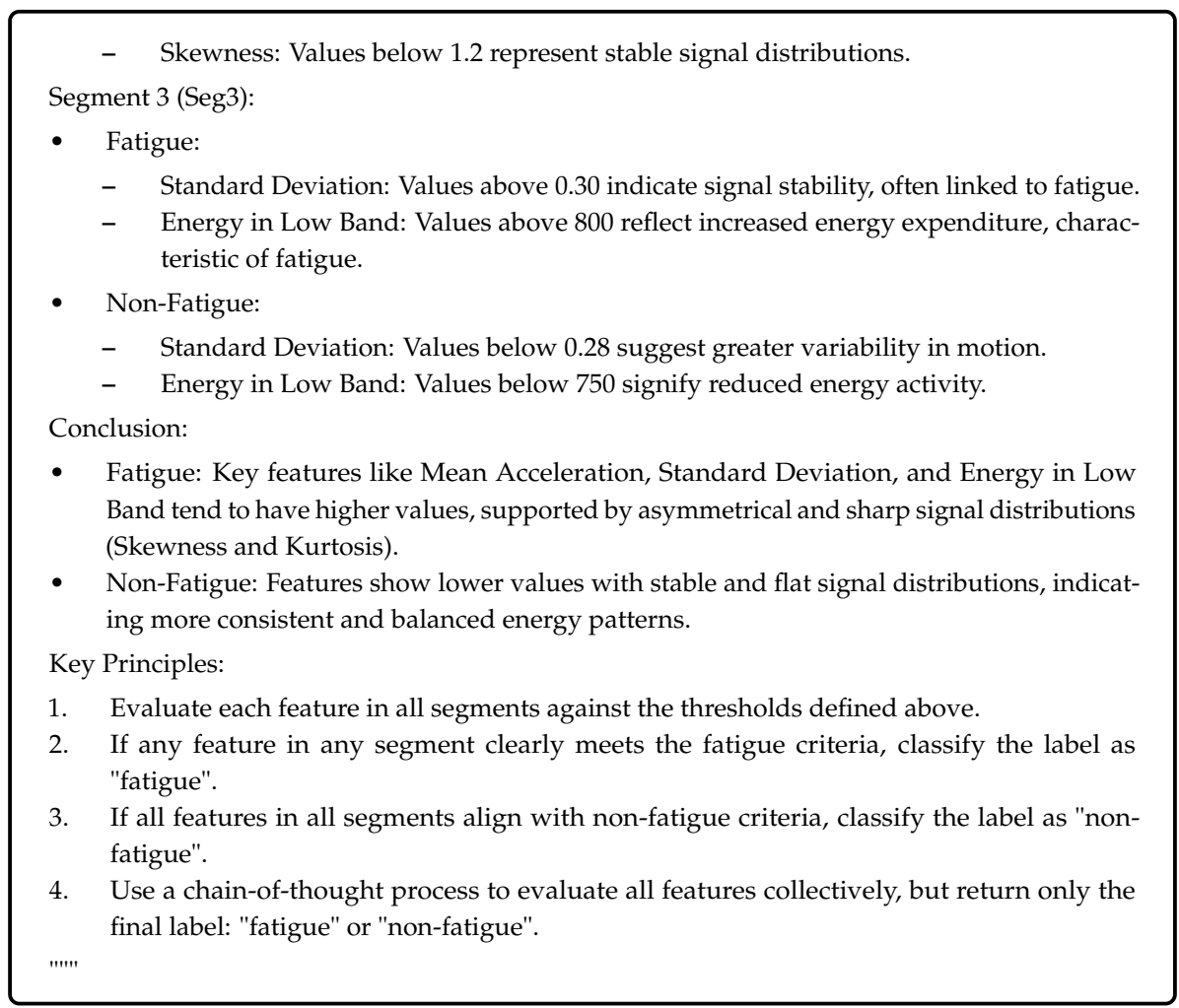
\captionof{figure}{\textbf{Example of LLM responses generating domain knowledge}. The responses are utilized to establish domain-specific insights, which are subsequently applied in LLM scoring and few-shot prompting to enhance performance in the target task.}
\label{fig:dom-know-res}

\subsection[\appendixname~\thesubsection]{}
\label{apd:few_shot}
In this section, we outline the design of a few-shot prompting strategy for fatigue prediction, leveraging the GPT-4o-mini model with a temperature setting of 0.3. Our approach integrates several key components: domain knowledge, instructions, context, and questions. Each element is critical in crafting effective prompts tailored to the task. For the context component, we employ a $top_{K}$ selection process to identify the most relevant examples based on evaluation results from LLM Scoring. These examples are then filtered down to two, forming a 2-shot prompting setup. To ensure balance and enhance the model's interpretability, we carefully select one example labeled as “fatigue” and another labeled as “non-fatigue.” This approach provides a clear distinction between the two states, aiding the model in better understanding the task.

In cases where the $top_{K}$ examples lack representation from both labels, meaning all examples belong to either "fatigue" or "non-fatigue," we construct the 2-shot prompt using examples with the same label. This adaptive strategy ensures the robustness of our method, even in scenarios with label imbalance. Fig.~\ref{fig:few-prompting} illustrates additional details on the design of our few-shot prompting approach.
\tcbset{colback=white, colframe=black, fonttitle=\bfseries}
\begin{tcolorbox}[title=Few-shot Prompting, breakable]
\textbf{\#Domain Knowledge}: \texttt{\{domain\_knowledge\}}\\
\textbf{\#Instruction}: Perform step-by-step numeric comparison internally, don't reveal it.\\
\textbf{\#Context}:\\
\textit{Example 1 | Subj \{id\}, Label=\{label\_example\_1\}, Score=\{score\_example\_1\}}\\
Reason: \texttt{\{reason\_example\_1\}}

\noindent Segment 1: Features such as Mean \texttt{\{mean\}}, Std \texttt{\{standard\_deviation\}}, Max \texttt{\{max\}}, Min \texttt{\{min\}}, and Peak-to-Peak \texttt{\{peak-to-peak\}} provide key insights into activity levels. RMS \texttt{\{rms\}} is consistent with \texttt{\{label\_example\_1\}} thresholds. Skew \texttt{\{skewness\}} and Kurtosis \texttt{\{kurtosis\}} support the \texttt{\{label\_example\_1\}} classification. Dominant Frequency \texttt{\{dominant\_frequency\}} and Energy in Low Band \texttt{\{energy\_low\_band\}} suggest high activity levels.

\noindent Segment 2: Features such as Mean \texttt{\{mean\}}, Std \texttt{\{standard\_deviation\}}, Max \texttt{\{max\}}, Min \texttt{\{min\}}, and Peak-to-Peak \texttt{\{peak-to-peak\}} provide key insights into activity levels. RMS \texttt{\{rms\}} is consistent with \texttt{\{label\_example\_1\}} thresholds. Skew \texttt{\{skewness\}} and Kurtosis \texttt{\{kurtosis\}} support the \texttt{\{label\_example\_1\}} classification. Dominant Frequency \texttt{\{dominant\_frequency\}} and Energy in Low Band \texttt{\{energy\_low\_band\}} suggest high activity levels.

\noindent Segment 3: Features such as Mean \texttt{\{mean\}}, Std \texttt{\{standard\_deviation\}}, Max \texttt{\{max\}}, Min \texttt{\{min\}}, and Peak-to-Peak \texttt{\{peak-to-peak\}} provide key insights into activity levels. RMS \texttt{\{rms\}} is consistent with \texttt{\{label\_example\_1\}} thresholds. Skew \texttt{\{skewness\}} and Kurtosis \texttt{\{kurtosis\}} support the \texttt{\{label\_example\_1\}} classification. Dominant Frequency \texttt{\{dominant\_frequency\}} and Energy in Low Band \texttt{\{energy\_low\_band\}} suggest high activity levels.

\noindent Conclusion: The overall pattern in features across segments supports the classification of \texttt{\{label\_example\_1\}}.\\
Label: \texttt{\{label\_example\_1\}}.
\vspace{2mm}\\
\textit{Example 2 | Subj \{id\}, Label=\{label\_example\_2\}, Score=\{score\_example\_2\}}\\
Reason: \texttt{\{reason\_example\_2\}}

\noindent Segment 1: Features such as Mean \texttt{\{mean\}}, Std \texttt{\{standard\_deviation\}}, Max \texttt{\{max\}}, Min \texttt{\{min\}}, and Peak-to-Peak \texttt{\{peak-to-peak\}} provide key insights into activity levels. RMS \texttt{\{rms\}} is consistent with \texttt{\{label\_example\_2\}} thresholds. Skew \texttt{\{skewness\}} and Kurtosis \texttt{\{kurtosis\}} support the \texttt{\{label\_example\_2\}} classification. Dominant Frequency \texttt{\{dominant\_frequency\}} and Energy in Low Band \texttt{\{energy\_low\_band\}} suggest high activity levels.

\noindent Segment 2: Features such as Mean \texttt{\{mean\}}, Std \texttt{\{standard\_deviation\}}, Max \texttt{\{max\}}, Min \texttt{\{min\}}, and Peak-to-Peak \texttt{\{peak-to-peak\}} provide key insights into activity levels. RMS \texttt{\{rms\}} is consistent with \texttt{\{label\_example\_2\}} thresholds. Skew \texttt{\{skewness\}} and Kurtosis \texttt{\{kurtosis\}} support the \texttt{\{label\_example\_2\}} classification. Dominant Frequency \texttt{\{dominant\_frequency\}} and Energy in Low Band \texttt{\{energy\_low\_band\}} suggest high activity levels.

\noindent Segment 3: Features such as Mean \texttt{\{mean\}}, Std \texttt{\{standard\_deviation\}}, Max \texttt{\{max\}}, Min \texttt{\{min\}}, and Peak-to-Peak \texttt{\{peak-to-peak\}} provide key insights into activity levels. RMS \texttt{\{rms\}} is consistent with \texttt{\{label\_example\_2\}} thresholds. Skew \texttt{\{skewness\}} and Kurtosis \texttt{\{kurtosis\}} support the \texttt{\{label\_example\_2\}} classification. Dominant Frequency \texttt{\{dominant\_frequency\}} and Energy in Low Band \texttt{\{energy\_low\_band\}} suggest high activity levels.

\noindent Conclusion: The overall pattern in features across segments supports the classification of \texttt{\{label\_example\_2\}}.\\
Label: \texttt{\{label\_example\_2\}}.
\vspace{2mm}\\
\textit{New Data | Subject \{id\}, Unlabeled}\\
\noindent Segment 1: Features such as Mean \texttt{\{mean\}}, Std \texttt{\{standard\_deviation\}}, Max \texttt{\{max\}}, Min \texttt{\{min\}}, and Peak-to-Peak \texttt{\{peak-to-peak\}} provide key insights into activity levels. RMS \texttt{\{rms\}} indicates overall activity magnitude. Skew \texttt{\{skewness\}} and Kurtosis \texttt{\{kurtosis\}} reveal distribution characteristics of the signal. Dominant Frequency \texttt{\{dominant\_frequency\}} and Energy in Low Band \texttt{\{energy\_low\_band\}} will be evaluated to determine their relevance to the classification.

\noindent Segment 2: Features such as Mean \texttt{\{mean\}}, Std \texttt{\{standard\_deviation\}}, Max \texttt{\{max\}}, Min \texttt{\{min\}}, and Peak-to-Peak \texttt{\{peak-to-peak\}} provide key insights into activity levels. RMS \texttt{\{rms\}} indicates overall activity magnitude. Skew \texttt{\{skewness\}} and Kurtosis \texttt{\{kurtosis\}} reveal distribution characteristics of the signal. Dominant Frequency \texttt{\{dominant\_frequency\}} and Energy in Low Band \texttt{\{energy\_low\_band\}} will be evaluated to determine their relevance to the classification.

\noindent Segment 3: Features such as Mean \texttt{\{mean\}}, Std \texttt{\{standard\_deviation\}}, Max \texttt{\{max\}}, Min \texttt{\{min\}}, and Peak-to-Peak \texttt{\{peak-to-peak\}} provide key insights into activity levels. RMS \texttt{\{rms\}} indicates overall activity magnitude. Skew \texttt{\{skewness\}} and Kurtosis \texttt{\{kurtosis\}} reveal distribution characteristics of the signal. Dominant Frequency \texttt{\{dominant\_frequency\}} and Energy in Low Band \texttt{\{energy\_low\_band\}} will be evaluated to determine their relevance to the classification.
\vspace{2mm}\\
\#\textbf{Task:}\\
Based on the provided examples and the new data:
\begin{enumerate}
    \item Compare the new data's features with the provided examples.
    \item Evaluate which example the new data most closely matches.
    \item Based on the closeness and domain knowledge, classify the new data as ‘fatigue’ or ‘non-fatigue’.
    \item Use reasoning internally to justify your classification, but return only the final label: ‘fatigue’ or ‘non-fatigue’.
\end{enumerate}

\#\textbf{Question:}\\
Please classify the new data using the provided examples and domain knowledge. Return only ‘fatigue’ or ‘non-fatigue’ as the final label.
\end{tcolorbox}
\captionof{figure}{\textbf{Prompting design for Few-shot Prompting}. The framework consists of multiple components: (a) Domain knowledge to define threshold values for optimizing performance, (b) Instructions to streamline the LLM's focus, (c) Context to supply essential background information, and (d) Questions to facilitate precise and relevant responses.}
\label{fig:few-prompting}
% \begin{table}[H] 
% \caption{This is a table caption.\label{tab5}}
% %\newcolumntype{C}{>{\centering\arraybackslash}X}
% \begin{tabularx}{\textwidth}{CCC}
% \toprule
% \textbf{Title 1}	& \textbf{Title 2}	& \textbf{Title 3}\\
% \midrule
% Entry 1		& Data			& Data\\
% Entry 2		& Data			& Data\\
% \bottomrule
% \end{tabularx}
% \end{table}

\subsection[\appendixname~\thesubsection]{}
\label{apd:conf_mat}
This section presents the confusion matrix results for the baseline methods and our proposed approach, providing a detailed model performance analysis. The confusion matrix is a critical tool for evaluating model behavior, offering insights into the distribution of predictions through metrics such as True Positives (TP), True Negatives (TN), False Positives (FP), and False Negatives (FN). These values are integral to calculating the macro F1-score, a key evaluation metric that captures the balance between precision and recall across all classes.

\begin{figure}[H]
  \centering
  \begin{tabular}{@{}c@{}}
    \includegraphics[width=.2\linewidth]{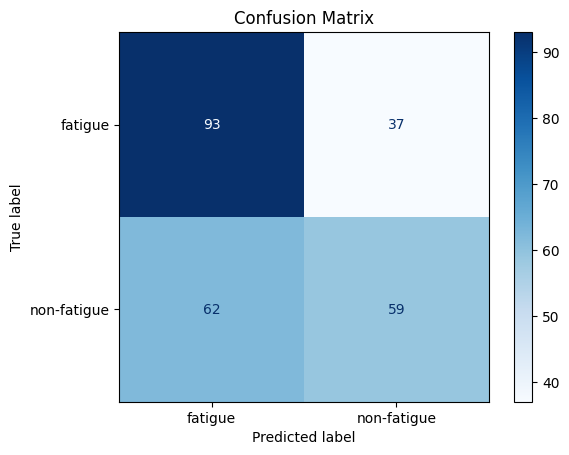} \\[\abovecaptionskip]
    \small (a) User-ID 4
  \end{tabular}
  \begin{tabular}{@{}c@{}}
    \includegraphics[width=.2\linewidth]{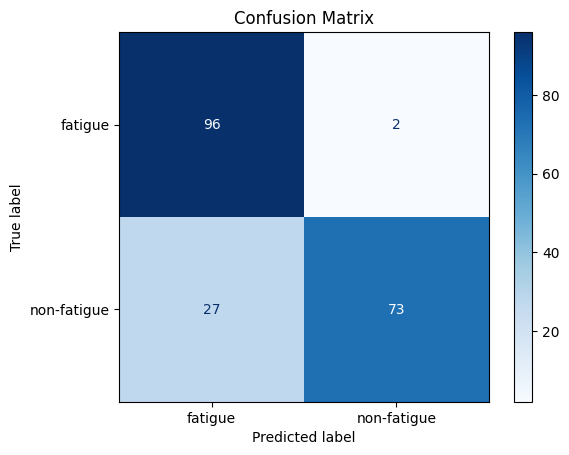} \\[\abovecaptionskip]
    \small (b) User-ID 5
  \end{tabular}
    \vspace{\floatsep}
  \begin{tabular}{@{}c@{}}
    \includegraphics[width=.2\linewidth]{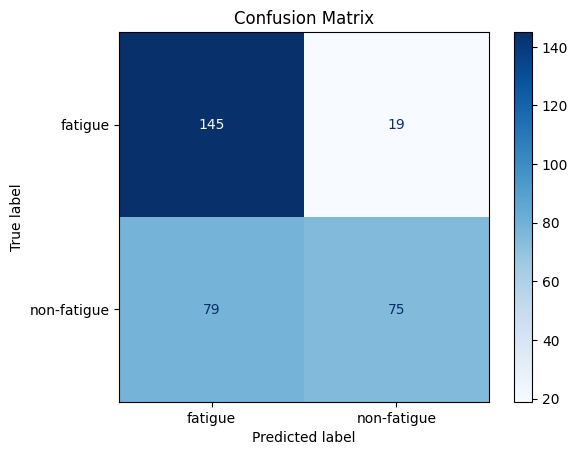} \\[\abovecaptionskip]
    \small (c) User-ID 6
  \end{tabular}
  \begin{tabular}{@{}c@{}}
    \includegraphics[width=.2\linewidth]{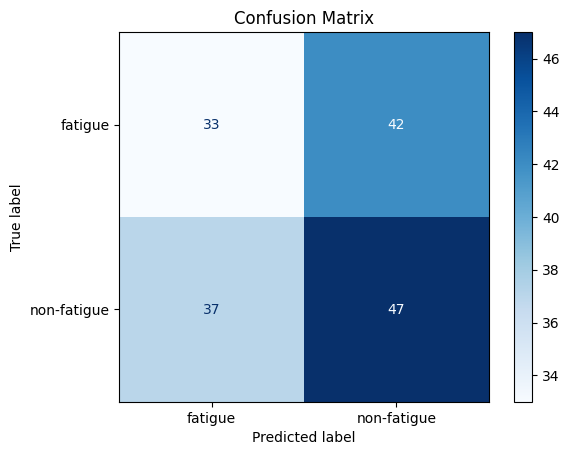} \\[\abovecaptionskip]
    \small (d) User-ID 7
  \end{tabular}
   \begin{tabular}{@{}c@{}}
    \includegraphics[width=.2\linewidth]{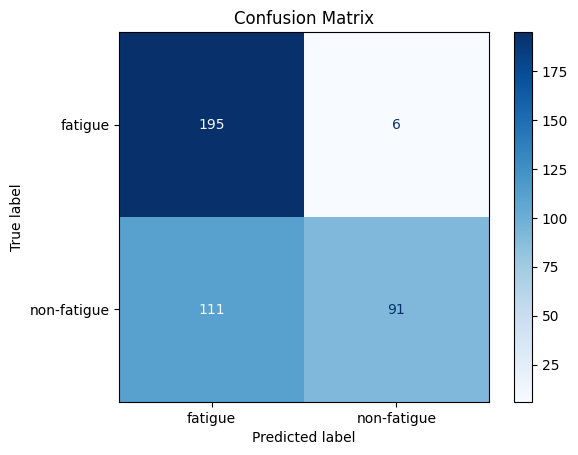} \\[\abovecaptionskip]
    \small (e) User-ID 8
  \end{tabular}
   \begin{tabular}{@{}c@{}}
    \includegraphics[width=.2\linewidth]{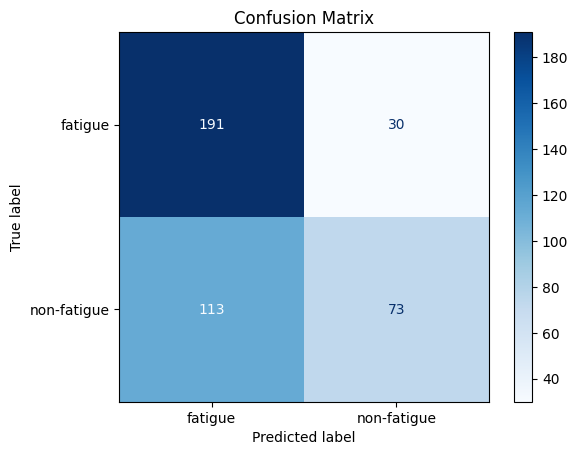} \\[\abovecaptionskip]
    \small (f) User-ID 9
  \end{tabular}
   \begin{tabular}{@{}c@{}}
    \includegraphics[width=.2\linewidth]{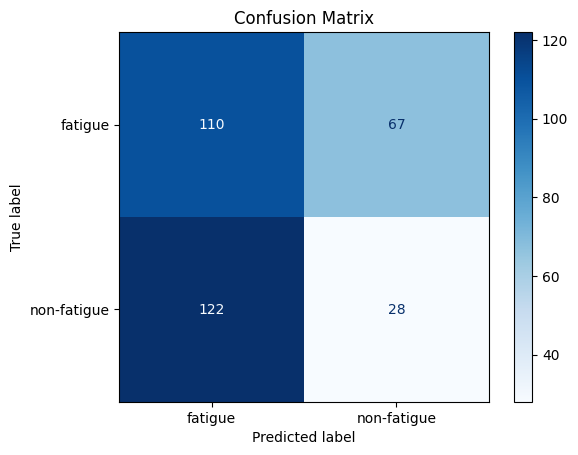} \\[\abovecaptionskip]
    \small (g) User-ID 10
  \end{tabular}
   \begin{tabular}{@{}c@{}}
    \includegraphics[width=.2\linewidth]{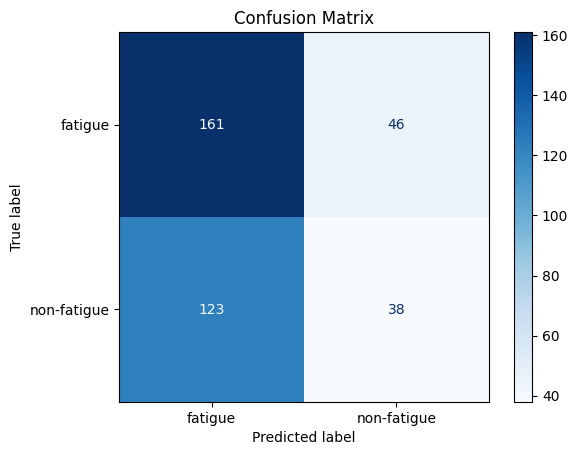} \\[\abovecaptionskip]
    \small (h) User-ID 11
  \end{tabular}
   \begin{tabular}{@{}c@{}}
    \includegraphics[width=.2\linewidth]{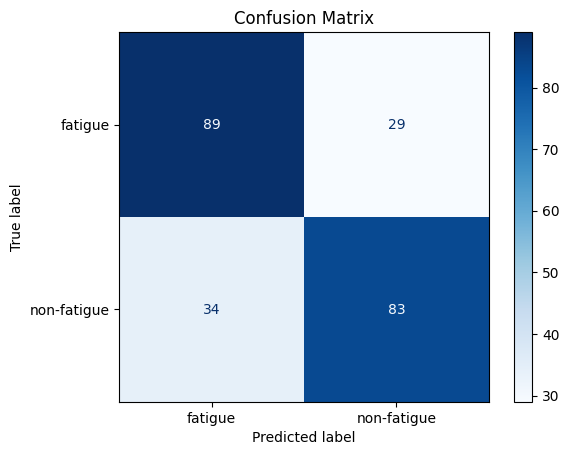} \\[\abovecaptionskip]
    \small (i) User-ID 12
  \end{tabular}
   \begin{tabular}{@{}c@{}}
    \includegraphics[width=.2\linewidth]{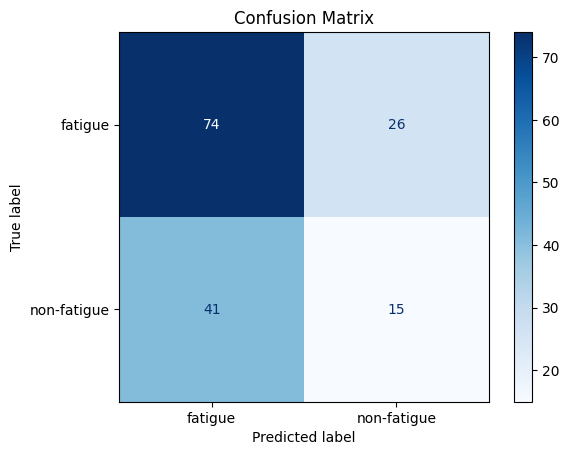} \\[\abovecaptionskip]
    \small (j) User-ID 13
  \end{tabular}
   \begin{tabular}{@{}c@{}}
    \includegraphics[width=.2\linewidth]{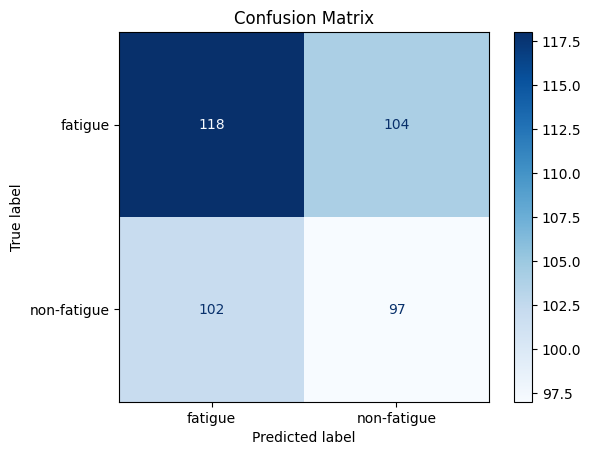} \\[\abovecaptionskip]
    \small (k) User-ID 14
  \end{tabular}
  \begin{tabular}{@{}c@{}}
    \includegraphics[width=.2\linewidth]{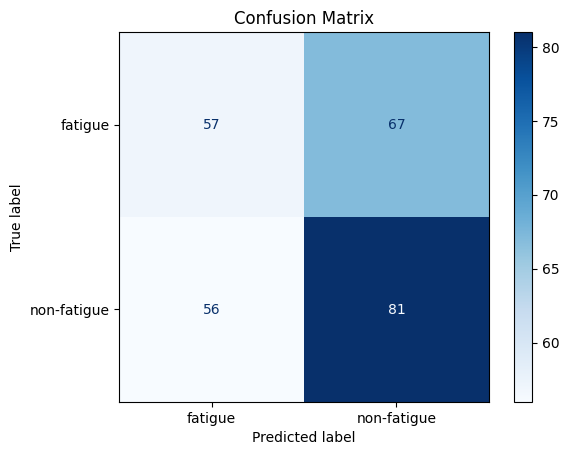} \\[\abovecaptionskip]
    \small (l) User-ID 15
  \end{tabular}
  \begin{tabular}{@{}c@{}}
    \includegraphics[width=.2\linewidth]{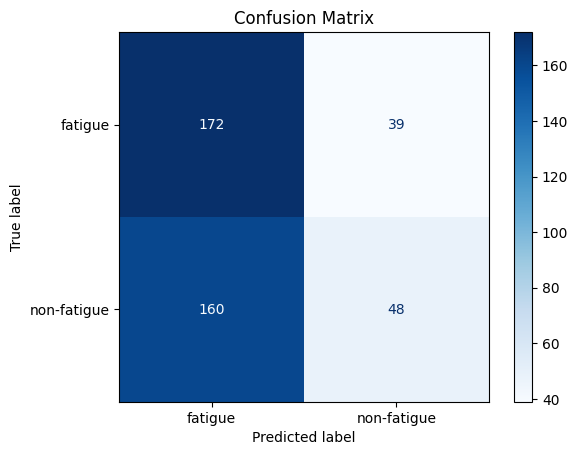} \\[\abovecaptionskip]
    \small (m) User-ID 17
  \end{tabular}
  \begin{tabular}{@{}c@{}}
    \includegraphics[width=.2\linewidth]{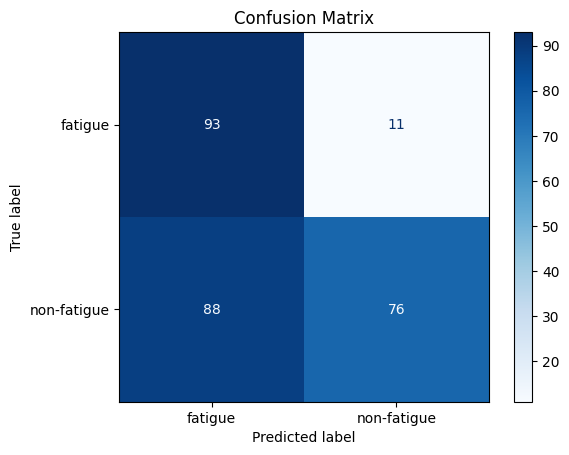} \\[\abovecaptionskip]
    \small (n) User-ID 18
  \end{tabular}
  \begin{tabular}{@{}c@{}}
    \includegraphics[width=.2\linewidth]{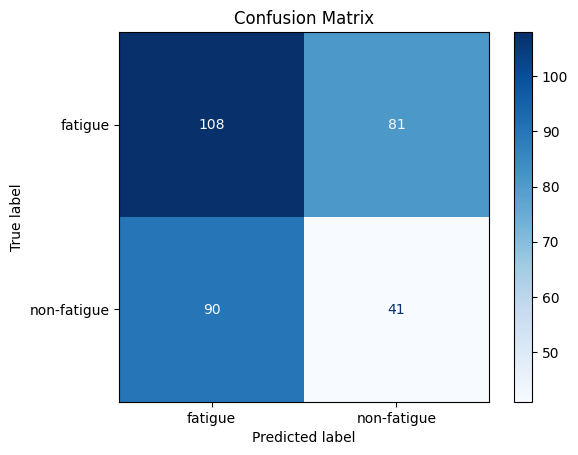} \\[\abovecaptionskip]
    \small (o) User-ID 19
  \end{tabular}
  \begin{tabular}{@{}c@{}}
    \includegraphics[width=.2\linewidth]{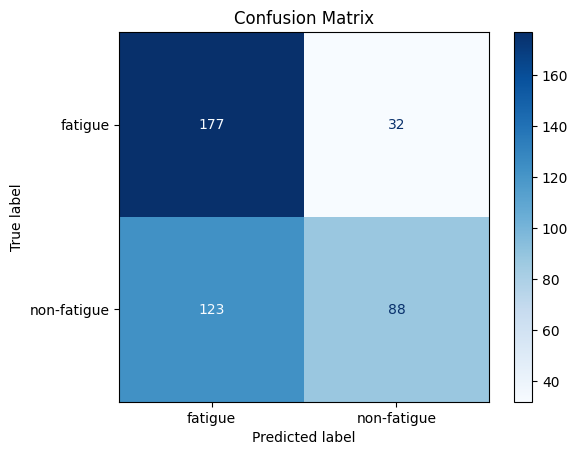} \\[\abovecaptionskip]
    \small (p) User-ID 20
  \end{tabular}
  \begin{tabular}{@{}c@{}}
    \includegraphics[width=.2\linewidth]{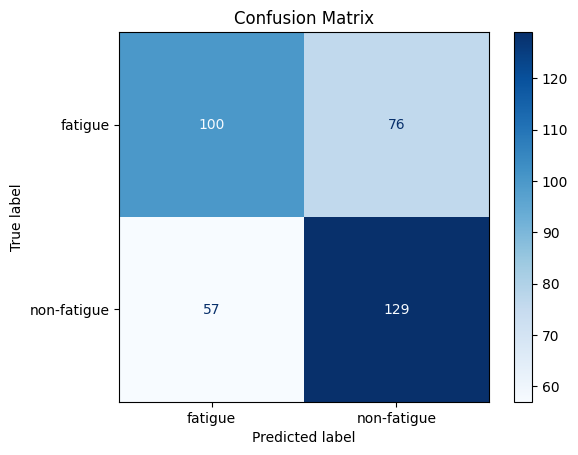} \\[\abovecaptionskip]
    \small (q) User-ID 21
  \end{tabular}
  \begin{tabular}{@{}c@{}}
    \includegraphics[width=.2\linewidth]{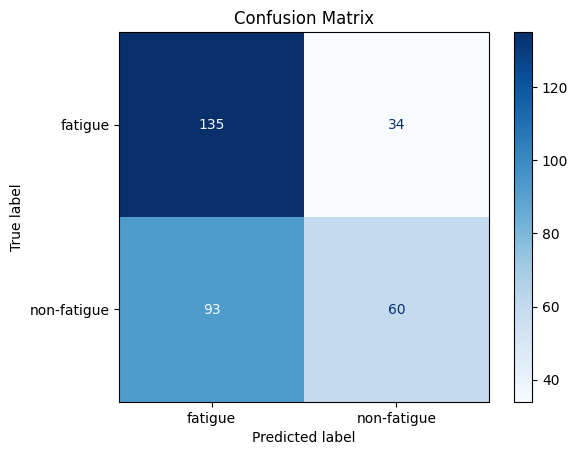} \\[\abovecaptionskip]
    \small (r) User-ID 22
  \end{tabular}
  \begin{tabular}{@{}c@{}}
    \includegraphics[width=.2\linewidth]{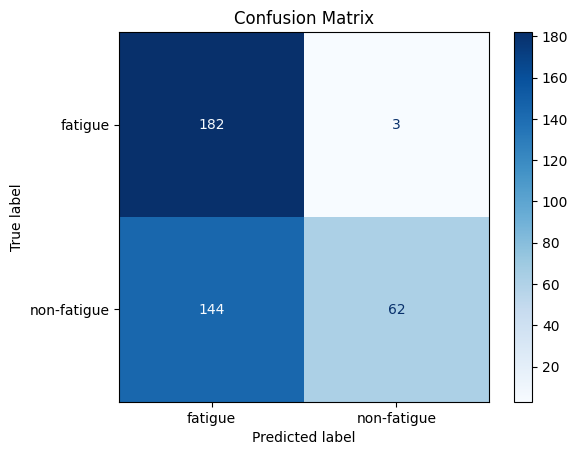} \\[\abovecaptionskip]
    \small (s) User-ID 23
  \end{tabular}
\caption{\textbf{Confusion Matrix for Random Approach}.}
\label{fig:random}
\end{figure}

\begin{figure}[H]
  \centering
  \begin{tabular}{@{}c@{}}
    \includegraphics[width=.23\linewidth]{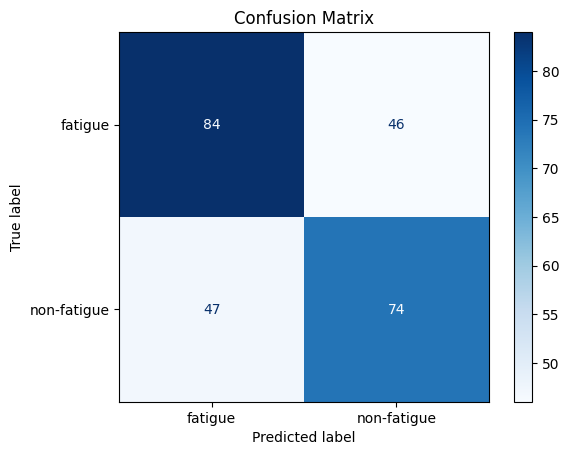} \\[\abovecaptionskip]
    \small (a) User-ID 4
  \end{tabular}
  \begin{tabular}{@{}c@{}}
    \includegraphics[width=.23\linewidth]{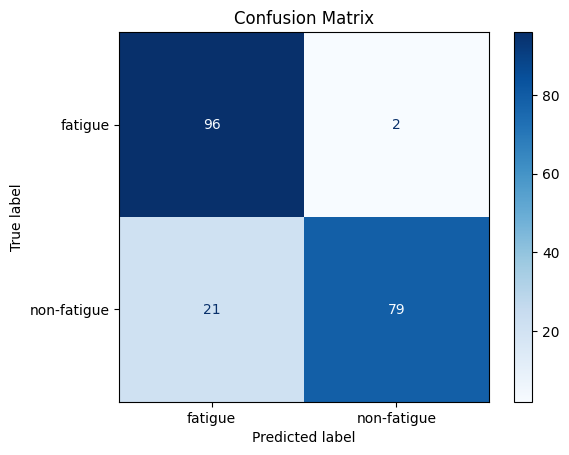} \\[\abovecaptionskip]
    \small (b) User-ID 5
  \end{tabular}
    \vspace{\floatsep}
  \begin{tabular}{@{}c@{}}
    \includegraphics[width=.23\linewidth]{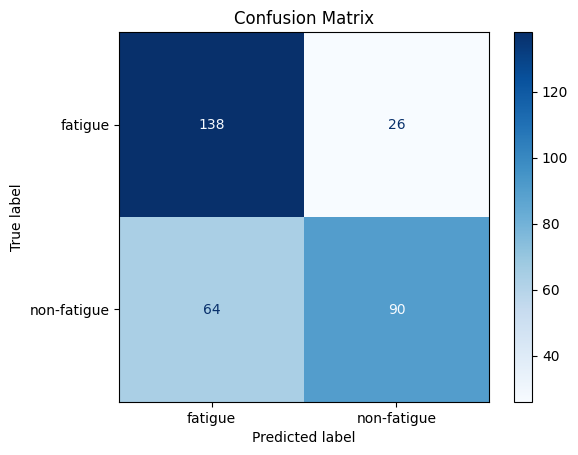} \\[\abovecaptionskip]
    \small (c) User-ID 6
  \end{tabular}
  \begin{tabular}{@{}c@{}}
    \includegraphics[width=.23\linewidth]{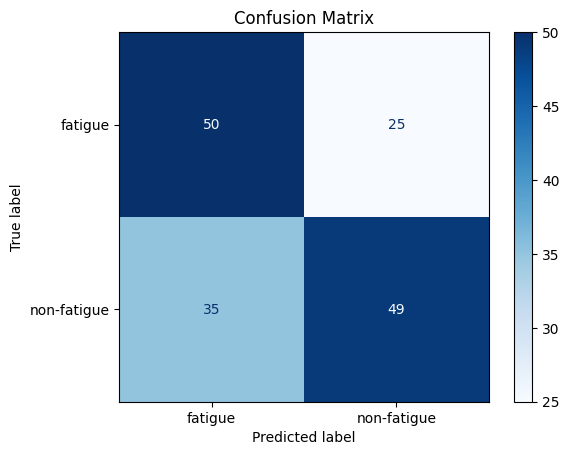} \\[\abovecaptionskip]
    \small (d) User-ID 7
  \end{tabular}
   \begin{tabular}{@{}c@{}}
    \includegraphics[width=.23\linewidth]{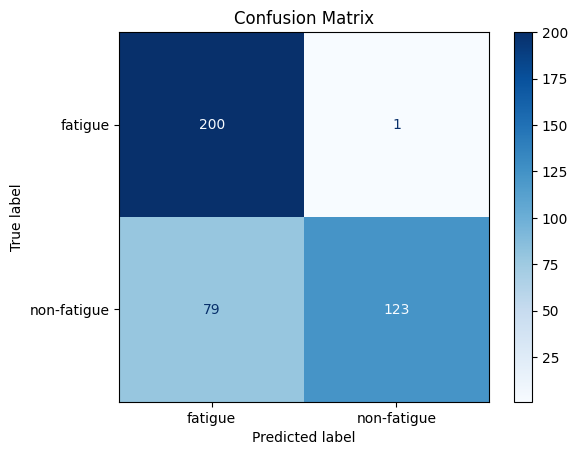} \\[\abovecaptionskip]
    \small (e) User-ID 8
  \end{tabular}
   \begin{tabular}{@{}c@{}}
    \includegraphics[width=.23\linewidth]{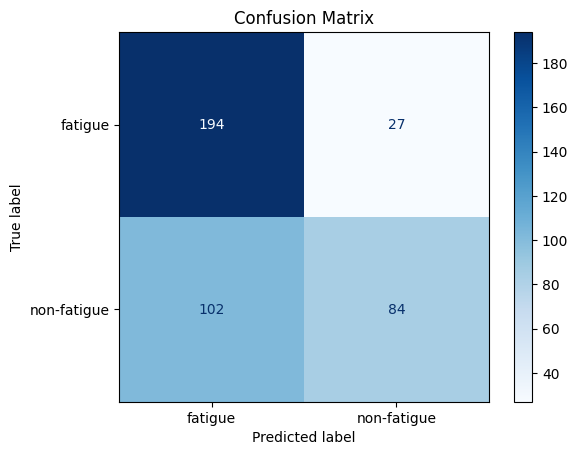} \\[\abovecaptionskip]
    \small (f) User-ID 9
  \end{tabular}
   \begin{tabular}{@{}c@{}}
    \includegraphics[width=.23\linewidth]{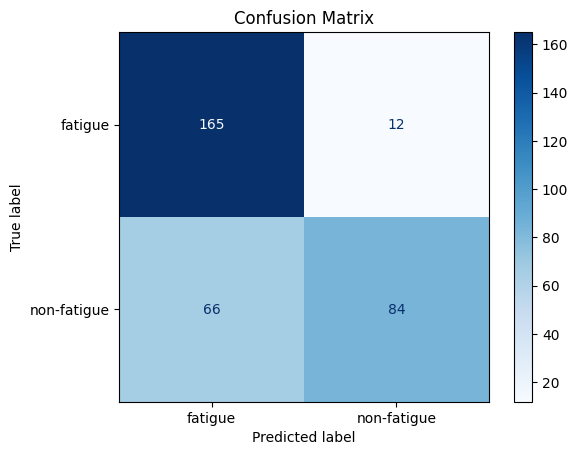} \\[\abovecaptionskip]
    \small (g) User-ID 10
  \end{tabular}
   \begin{tabular}{@{}c@{}}
    \includegraphics[width=.23\linewidth]{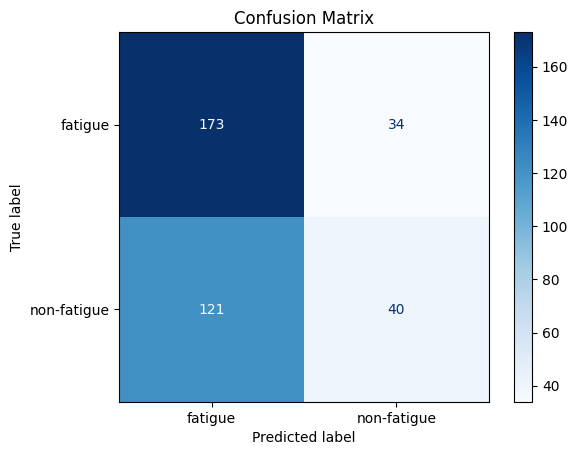} \\[\abovecaptionskip]
    \small (h) User-ID 11
  \end{tabular}
   \begin{tabular}{@{}c@{}}
    \includegraphics[width=.23\linewidth]{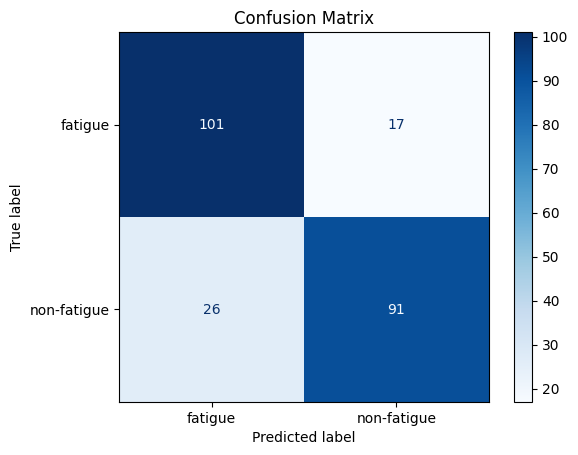} \\[\abovecaptionskip]
    \small (i) User-ID 12
  \end{tabular}
   \begin{tabular}{@{}c@{}}
    \includegraphics[width=.23\linewidth]{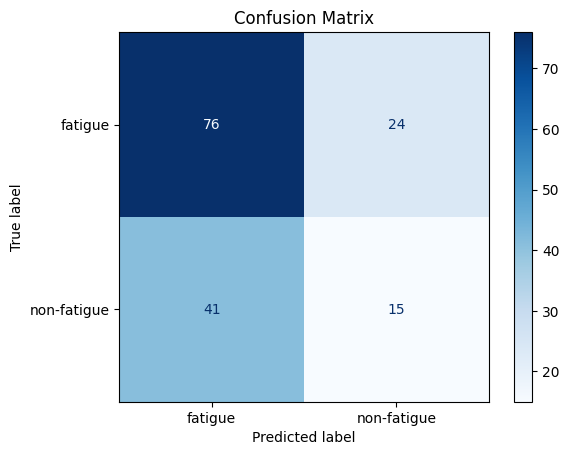} \\[\abovecaptionskip]
    \small (j) User-ID 13
  \end{tabular}
   \begin{tabular}{@{}c@{}}
    \includegraphics[width=.23\linewidth]{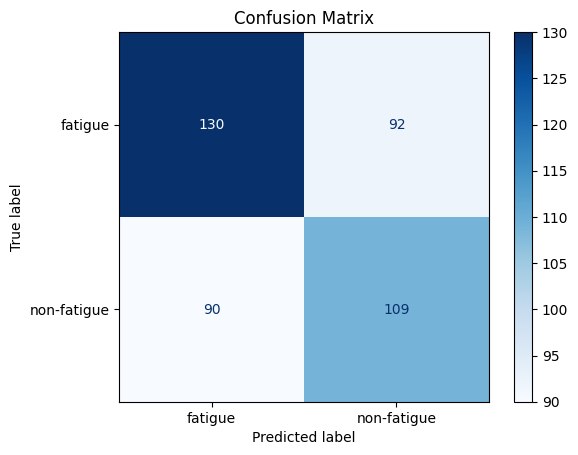} \\[\abovecaptionskip]
    \small (k) User-ID 14
  \end{tabular}
  \begin{tabular}{@{}c@{}}
    \includegraphics[width=.23\linewidth]{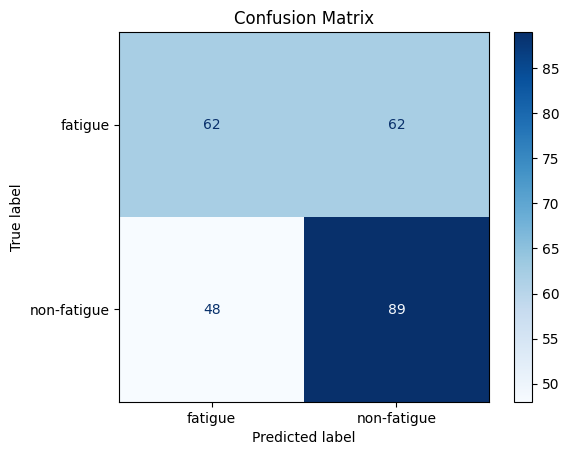} \\[\abovecaptionskip]
    \small (l) User-ID 15
  \end{tabular}
  \begin{tabular}{@{}c@{}}
    \includegraphics[width=.23\linewidth]{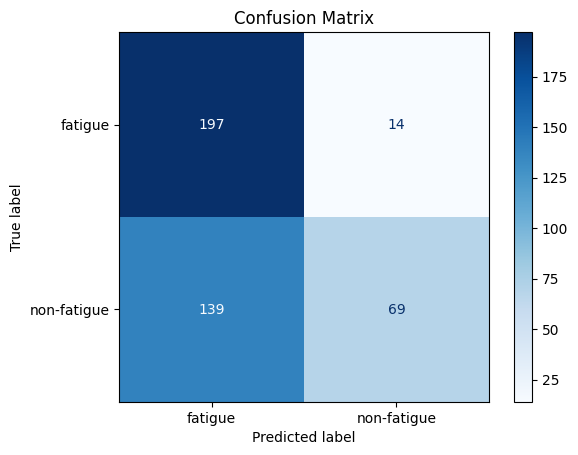} \\[\abovecaptionskip]
    \small (m) User-ID 17
  \end{tabular}
  \begin{tabular}{@{}c@{}}
    \includegraphics[width=.23\linewidth]{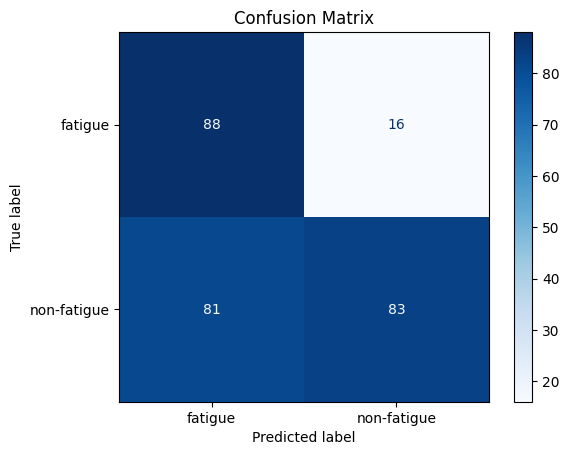} \\[\abovecaptionskip]
    \small (n) User-ID 18
  \end{tabular}
  \begin{tabular}{@{}c@{}}
    \includegraphics[width=.23\linewidth]{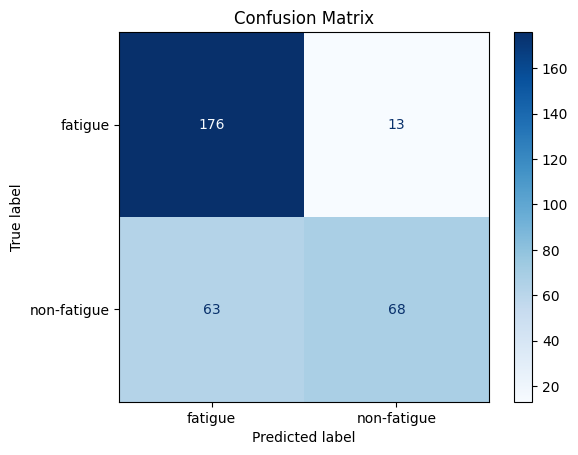} \\[\abovecaptionskip]
    \small (o) User-ID 19
  \end{tabular}
  \begin{tabular}{@{}c@{}}
    \includegraphics[width=.23\linewidth]{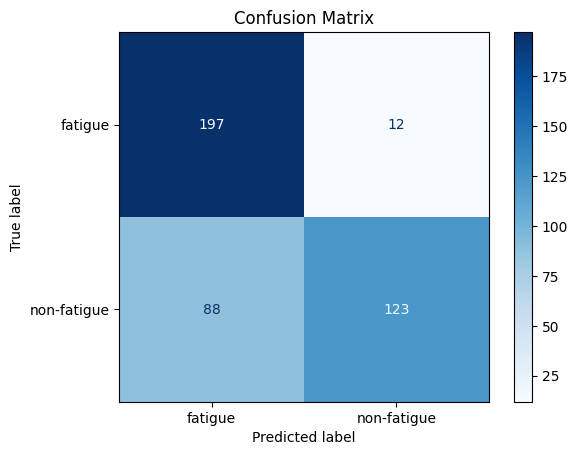} \\[\abovecaptionskip]
    \small (p) User-ID 20
  \end{tabular}
  \begin{tabular}{@{}c@{}}
    \includegraphics[width=.23\linewidth]{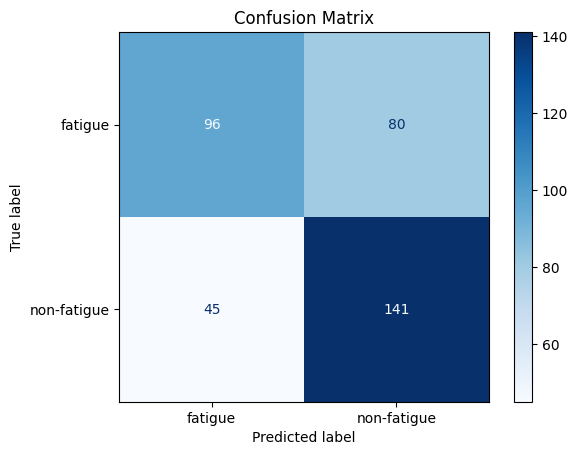} \\[\abovecaptionskip]
    \small (q) User-ID 21
  \end{tabular}
  \begin{tabular}{@{}c@{}}
    \includegraphics[width=.23\linewidth]{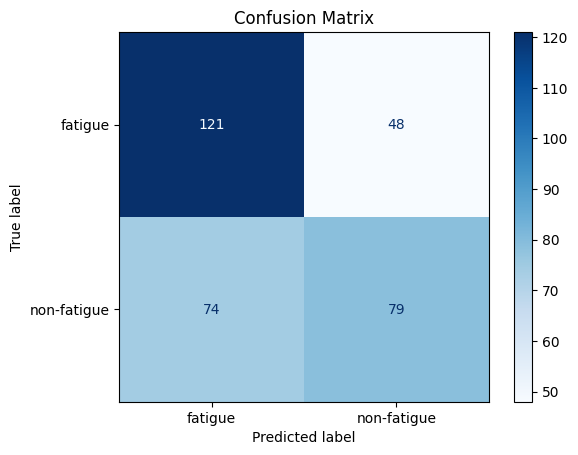} \\[\abovecaptionskip]
    \small (r) User-ID 22
  \end{tabular}
  \begin{tabular}{@{}c@{}}
    \includegraphics[width=.23\linewidth]{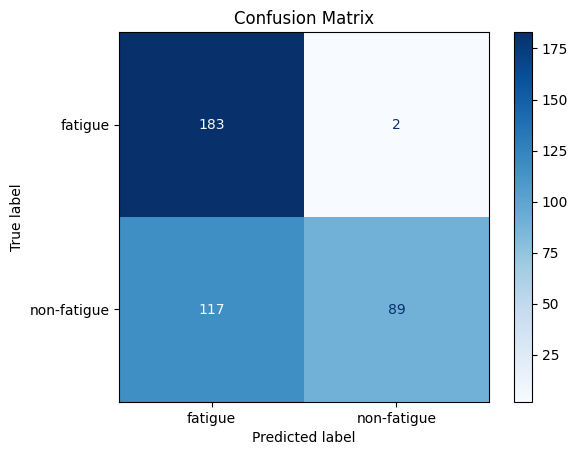} \\[\abovecaptionskip]
    \small (s) User-ID 23
  \end{tabular}
\caption{\textbf{Confusion Matrix for Distance Approach}.}
\label{fig:dist}
\end{figure}

\begin{figure}[H]
  \centering
  \begin{tabular}{@{}c@{}}
    \includegraphics[width=.23\linewidth]{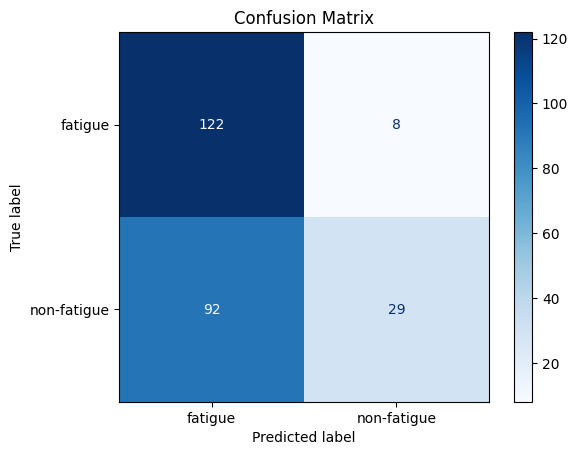} \\[\abovecaptionskip]
    \small (a) User-ID 4
  \end{tabular}
  \begin{tabular}{@{}c@{}}
    \includegraphics[width=.23\linewidth]{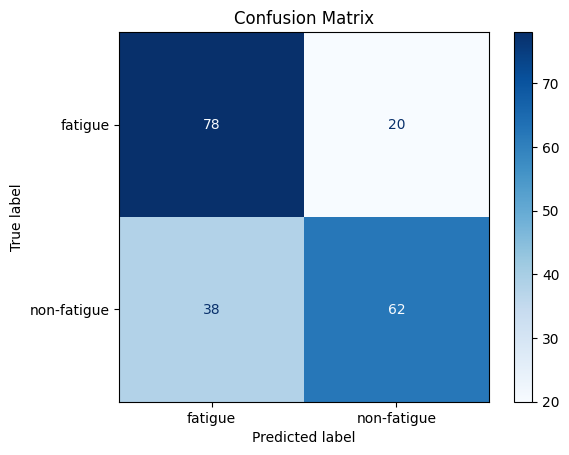} \\[\abovecaptionskip]
    \small (b) User-ID 5
  \end{tabular}
    \vspace{\floatsep}
  \begin{tabular}{@{}c@{}}
    \includegraphics[width=.23\linewidth]{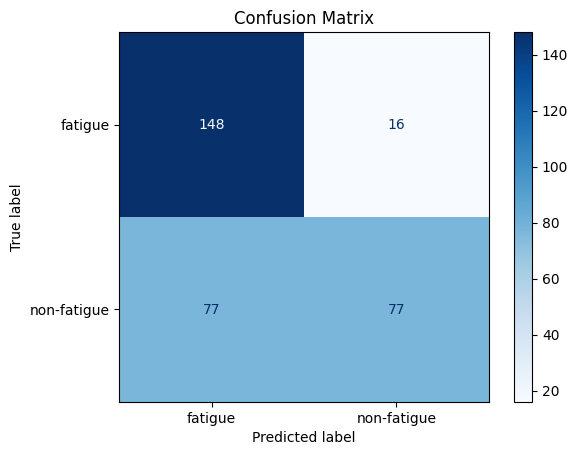} \\[\abovecaptionskip]
    \small (c) User-ID 6
  \end{tabular}
  \begin{tabular}{@{}c@{}}
    \includegraphics[width=.23\linewidth]{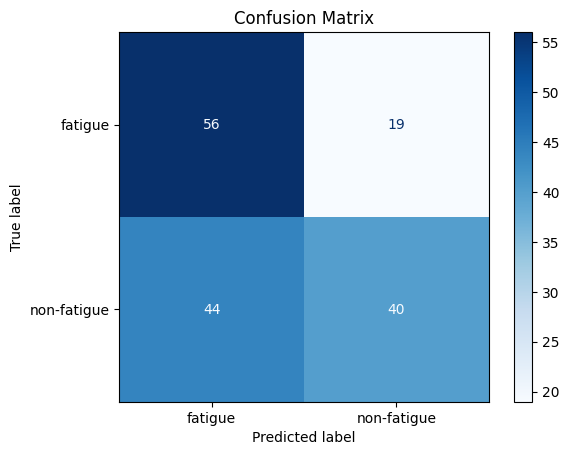} \\[\abovecaptionskip]
    \small (d) User-ID 7
  \end{tabular}
   \begin{tabular}{@{}c@{}}
    \includegraphics[width=.23\linewidth]{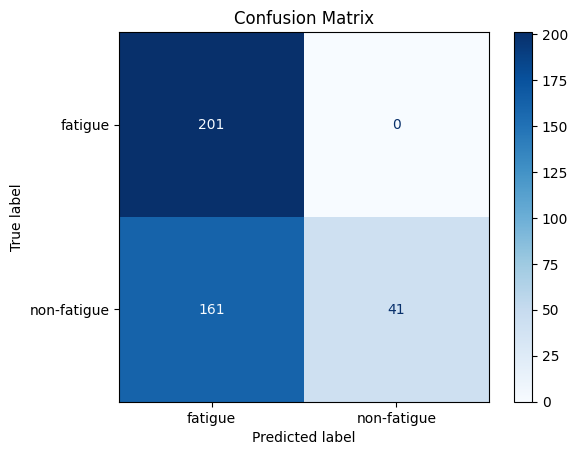} \\[\abovecaptionskip]
    \small (e) User-ID 8
  \end{tabular}
   \begin{tabular}{@{}c@{}}
    \includegraphics[width=.23\linewidth]{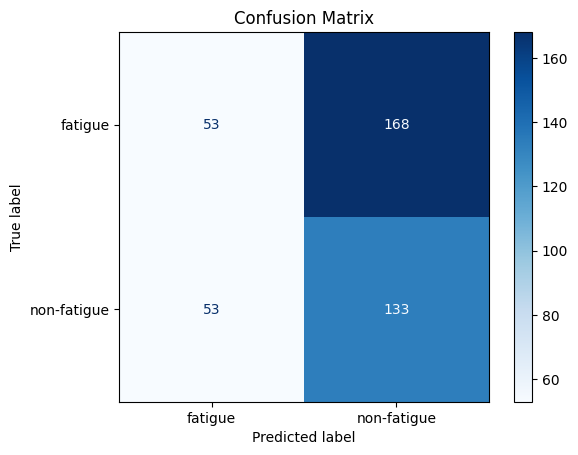} \\[\abovecaptionskip]
    \small (f) User-ID 9
  \end{tabular}
   \begin{tabular}{@{}c@{}}
    \includegraphics[width=.23\linewidth]{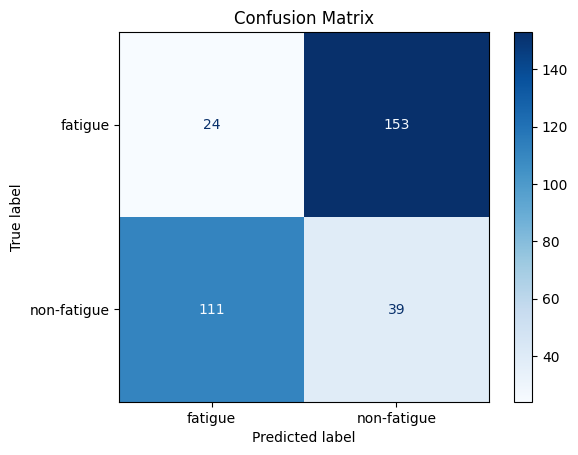} \\[\abovecaptionskip]
    \small (g) User-ID 10
  \end{tabular}
   \begin{tabular}{@{}c@{}}
    \includegraphics[width=.23\linewidth]{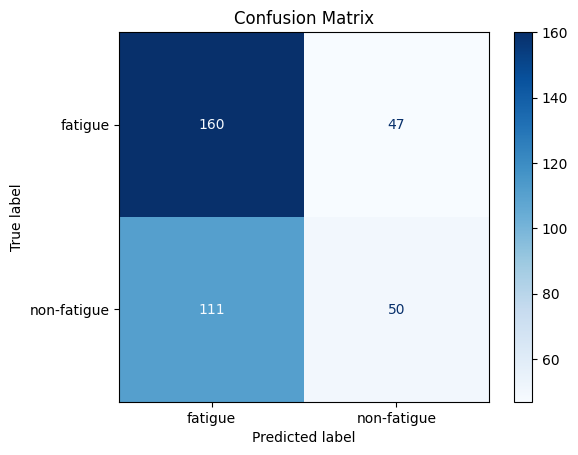} \\[\abovecaptionskip]
    \small (h) User-ID 11
  \end{tabular}
   \begin{tabular}{@{}c@{}}
    \includegraphics[width=.23\linewidth]{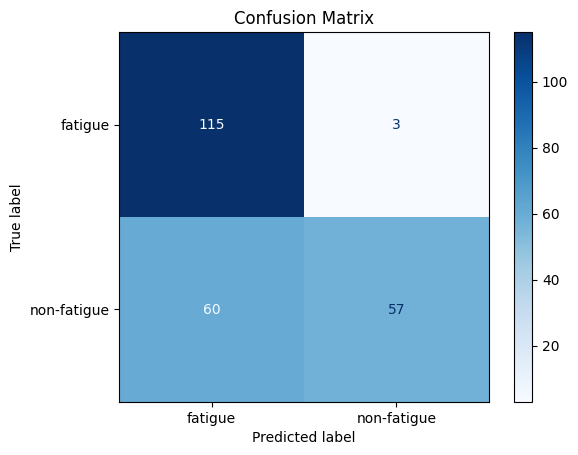} \\[\abovecaptionskip]
    \small (i) User-ID 12
  \end{tabular}
   \begin{tabular}{@{}c@{}}
    \includegraphics[width=.23\linewidth]{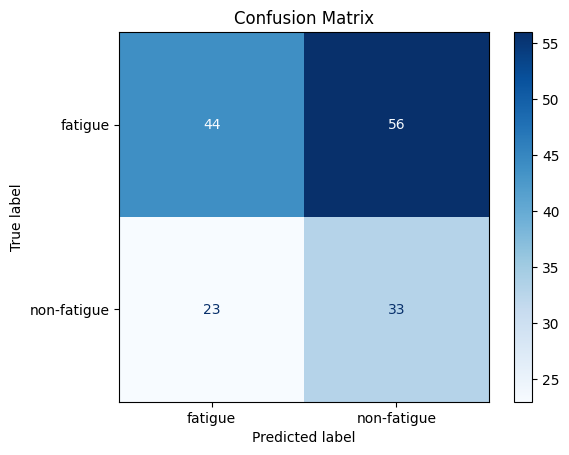} \\[\abovecaptionskip]
    \small (j) User-ID 13
  \end{tabular}
   \begin{tabular}{@{}c@{}}
    \includegraphics[width=.23\linewidth]{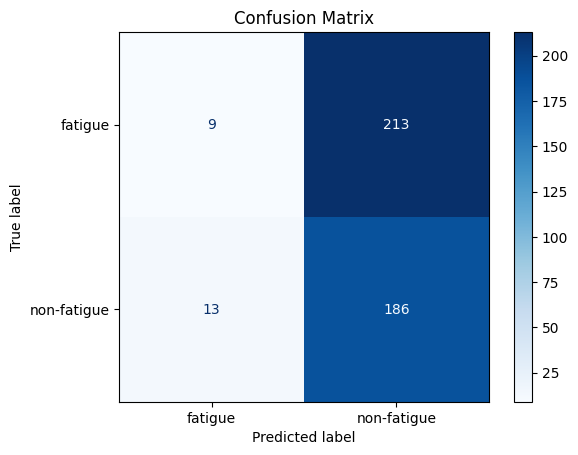} \\[\abovecaptionskip]
    \small (k) User-ID 14
  \end{tabular}
  \begin{tabular}{@{}c@{}}
    \includegraphics[width=.23\linewidth]{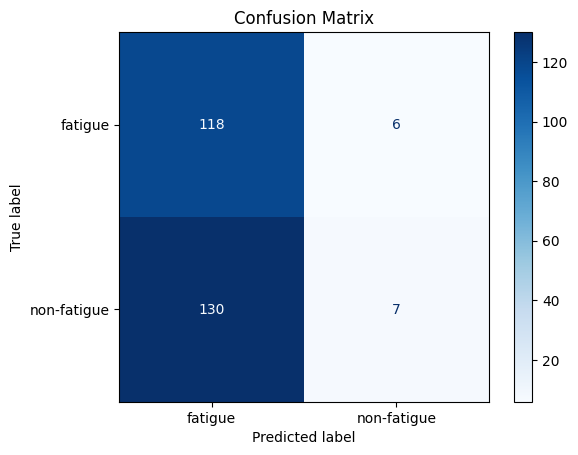} \\[\abovecaptionskip]
    \small (l) User-ID 15
  \end{tabular}
  \begin{tabular}{@{}c@{}}
    \includegraphics[width=.23\linewidth]{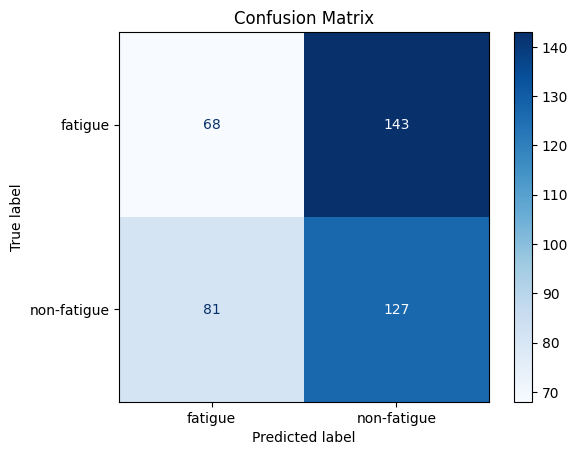} \\[\abovecaptionskip]
    \small (m) User-ID 17
  \end{tabular}
  \begin{tabular}{@{}c@{}}
    \includegraphics[width=.23\linewidth]{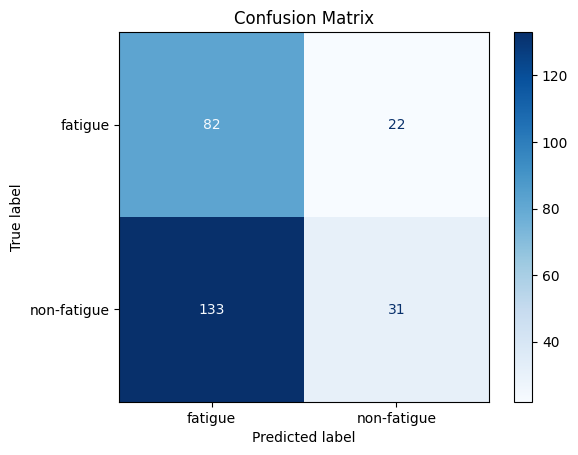} \\[\abovecaptionskip]
    \small (n) User-ID 18
  \end{tabular}
  \begin{tabular}{@{}c@{}}
    \includegraphics[width=.23\linewidth]{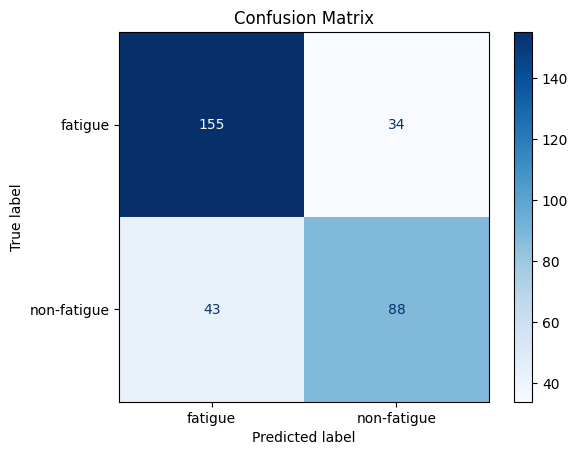} \\[\abovecaptionskip]
    \small (o) User-ID 19
  \end{tabular}
  \begin{tabular}{@{}c@{}}
    \includegraphics[width=.23\linewidth]{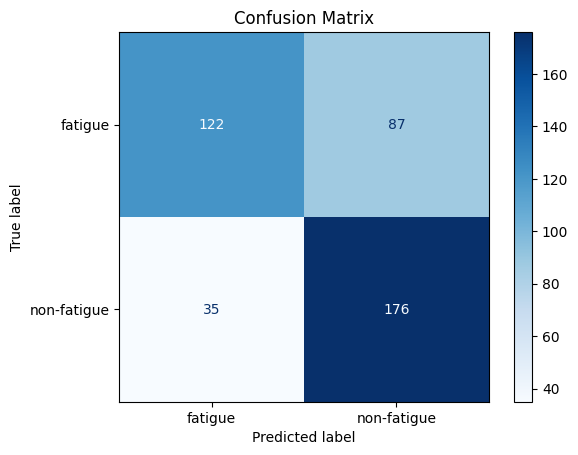} \\[\abovecaptionskip]
    \small (p) User-ID 20
  \end{tabular}
  \begin{tabular}{@{}c@{}}
    \includegraphics[width=.23\linewidth]{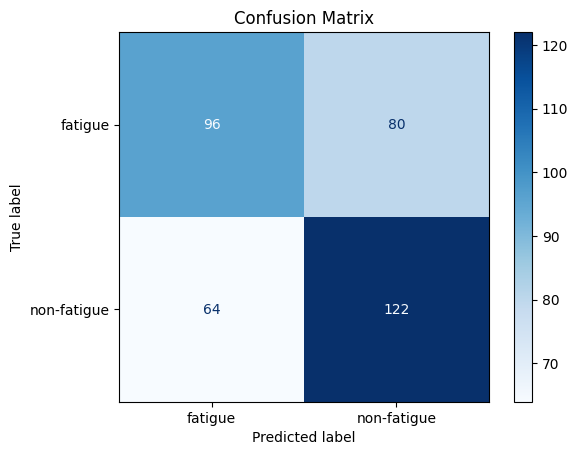} \\[\abovecaptionskip]
    \small (q) User-ID 21
  \end{tabular}
  \begin{tabular}{@{}c@{}}
    \includegraphics[width=.23\linewidth]{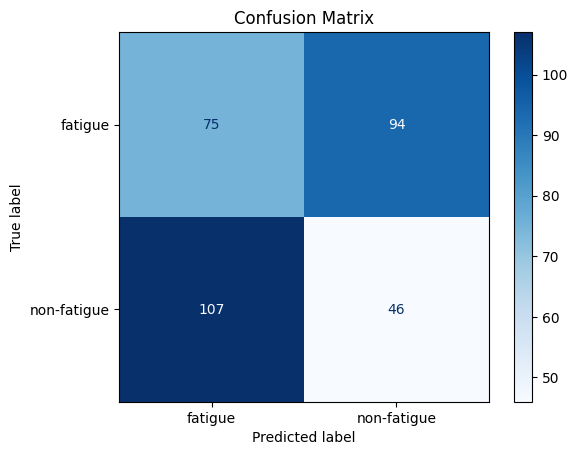} \\[\abovecaptionskip]
    \small (r) User-ID 22
  \end{tabular}
  \begin{tabular}{@{}c@{}}
    \includegraphics[width=.23\linewidth]{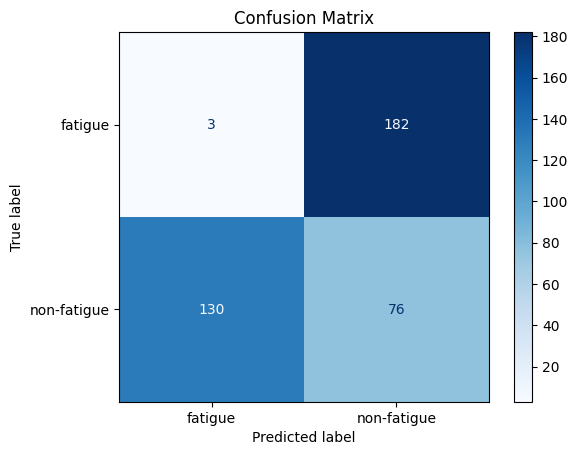} \\[\abovecaptionskip]
    \small (s) User-ID 23
  \end{tabular}
\caption{\textbf{Confusion Matrix for Traditional ML Approach}.}
\label{fig:ml}
\end{figure}

\begin{figure}[H]
  \centering
  \begin{tabular}{@{}c@{}}
    \includegraphics[width=.23\linewidth]{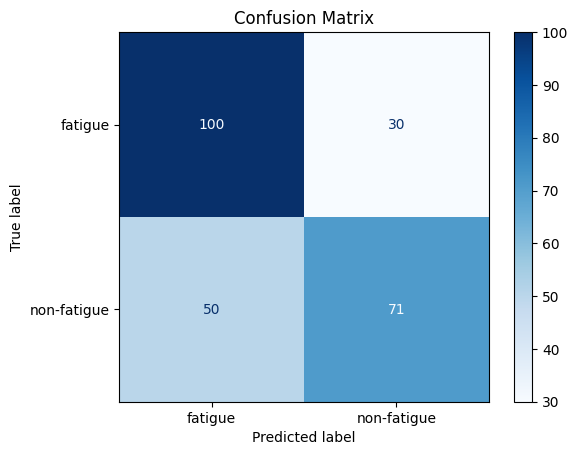} \\[\abovecaptionskip]
    \small (a) User-ID 4
  \end{tabular}
  \begin{tabular}{@{}c@{}}
    \includegraphics[width=.23\linewidth]{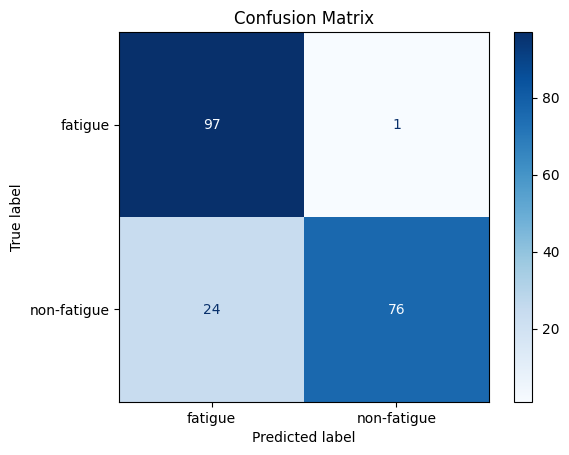} \\[\abovecaptionskip]
    \small (b) User-ID 5
  \end{tabular}
    \vspace{\floatsep}
  \begin{tabular}{@{}c@{}}
    \includegraphics[width=.23\linewidth]{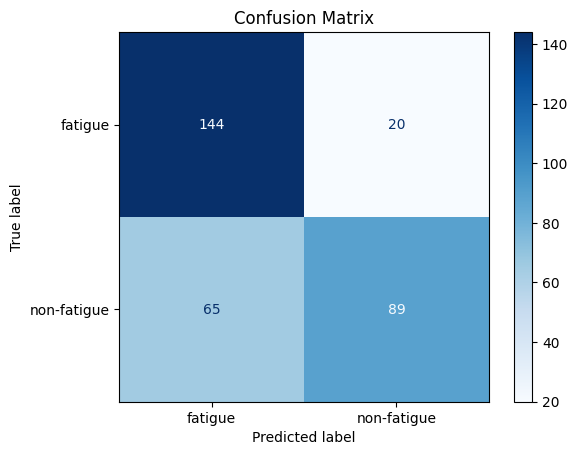} \\[\abovecaptionskip]
    \small (c) User-ID 6
  \end{tabular}
  \begin{tabular}{@{}c@{}}
    \includegraphics[width=.23\linewidth]{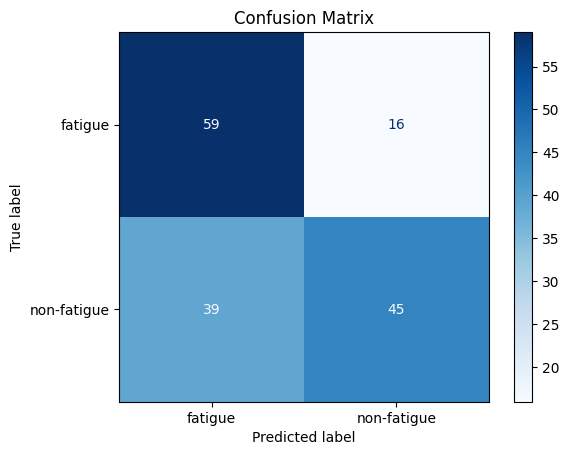} \\[\abovecaptionskip]
    \small (d) User-ID 7
  \end{tabular}
   \begin{tabular}{@{}c@{}}
    \includegraphics[width=.23\linewidth]{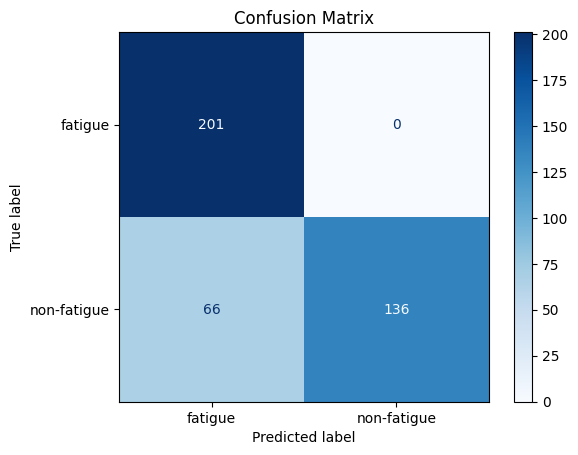} \\[\abovecaptionskip]
    \small (e) User-ID 8
  \end{tabular}
   \begin{tabular}{@{}c@{}}
    \includegraphics[width=.23\linewidth]{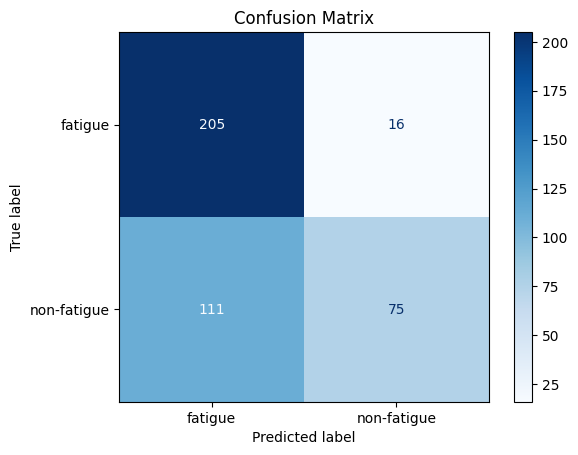} \\[\abovecaptionskip]
    \small (f) User-ID 9
  \end{tabular}
   \begin{tabular}{@{}c@{}}
    \includegraphics[width=.23\linewidth]{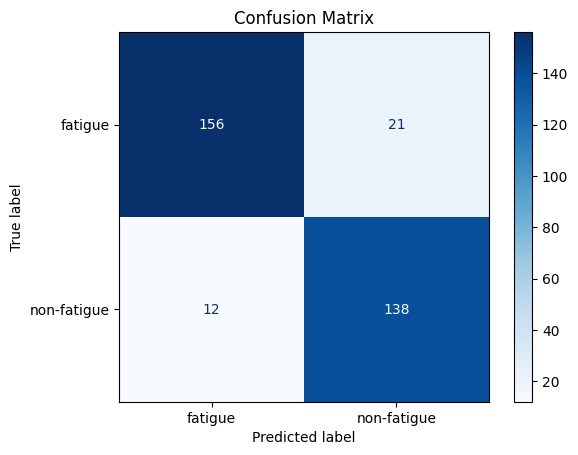} \\[\abovecaptionskip]
    \small (g) User-ID 10
  \end{tabular}
   \begin{tabular}{@{}c@{}}
    \includegraphics[width=.23\linewidth]{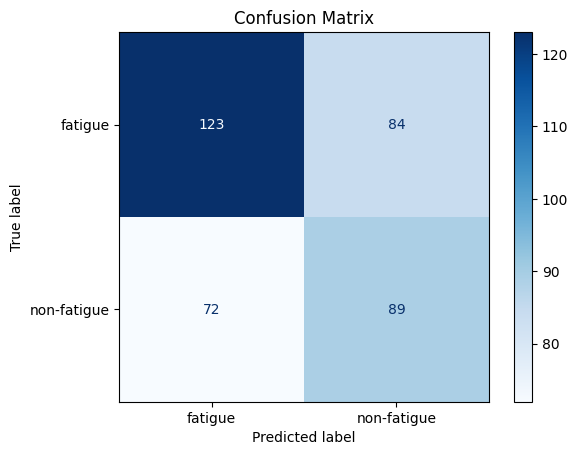} \\[\abovecaptionskip]
    \small (h) User-ID 11
  \end{tabular}
   \begin{tabular}{@{}c@{}}
    \includegraphics[width=.23\linewidth]{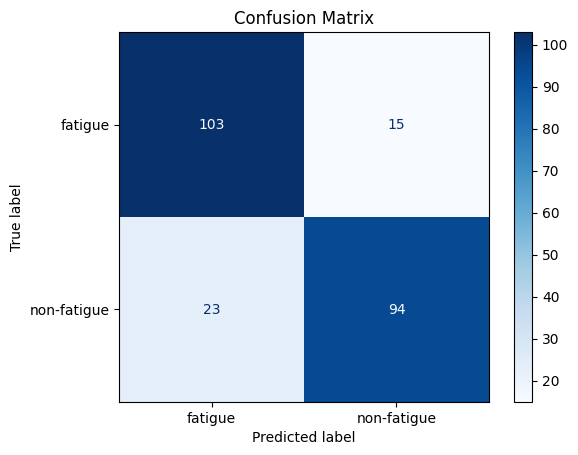} \\[\abovecaptionskip]
    \small (i) User-ID 12
  \end{tabular}
   \begin{tabular}{@{}c@{}}
    \includegraphics[width=.23\linewidth]{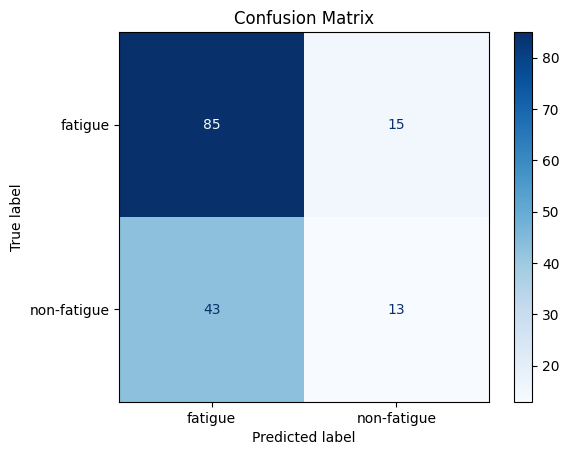} \\[\abovecaptionskip]
    \small (j) User-ID 13
  \end{tabular}
   \begin{tabular}{@{}c@{}}
    \includegraphics[width=.23\linewidth]{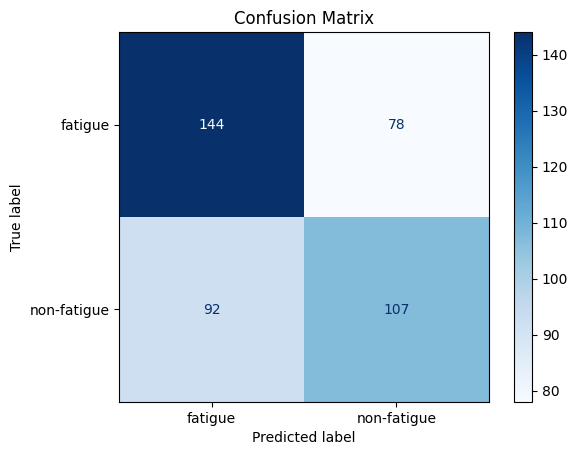} \\[\abovecaptionskip]
    \small (k) User-ID 14
  \end{tabular}
  \begin{tabular}{@{}c@{}}
    \includegraphics[width=.23\linewidth]{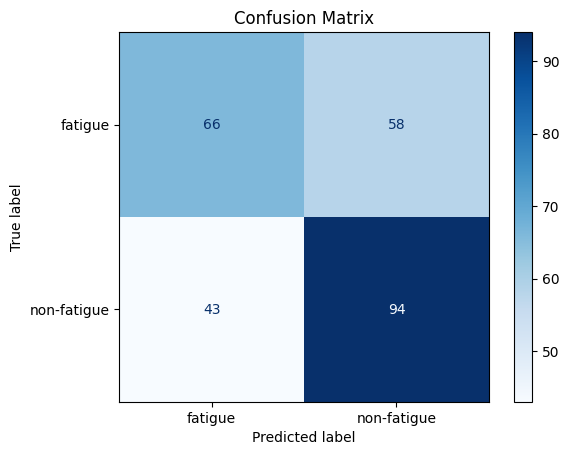} \\[\abovecaptionskip]
    \small (l) User-ID 15
  \end{tabular}
  \begin{tabular}{@{}c@{}}
    \includegraphics[width=.23\linewidth]{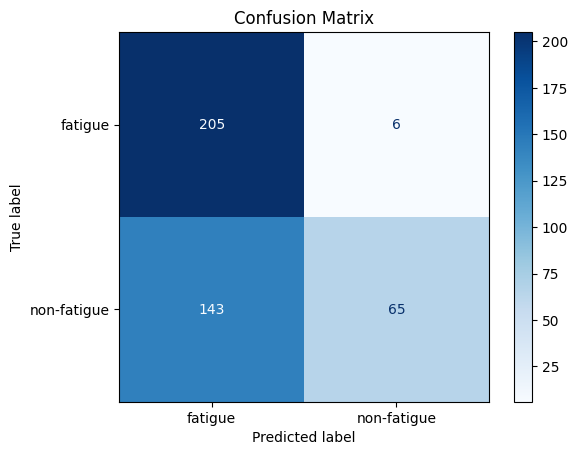} \\[\abovecaptionskip]
    \small (m) User-ID 17
  \end{tabular}
  \begin{tabular}{@{}c@{}}
    \includegraphics[width=.23\linewidth]{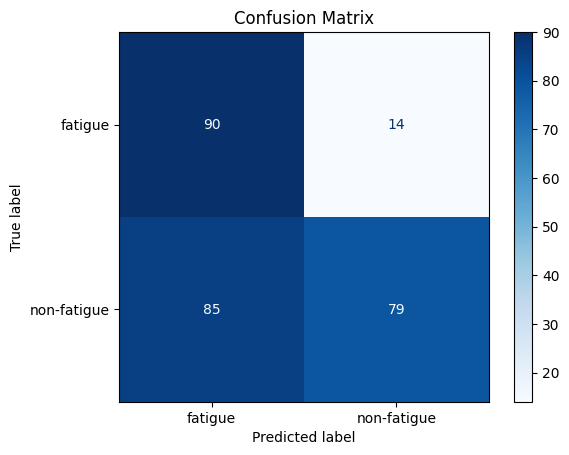} \\[\abovecaptionskip]
    \small (n) User-ID 18
  \end{tabular}
  \begin{tabular}{@{}c@{}}
    \includegraphics[width=.23\linewidth]{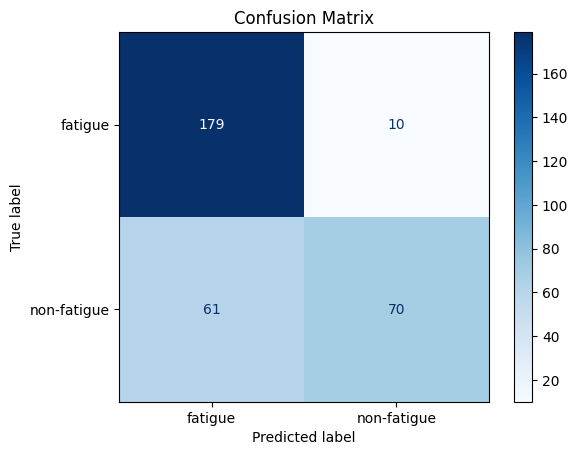} \\[\abovecaptionskip]
    \small (o) User-ID 19
  \end{tabular}
  \begin{tabular}{@{}c@{}}
    \includegraphics[width=.23\linewidth]{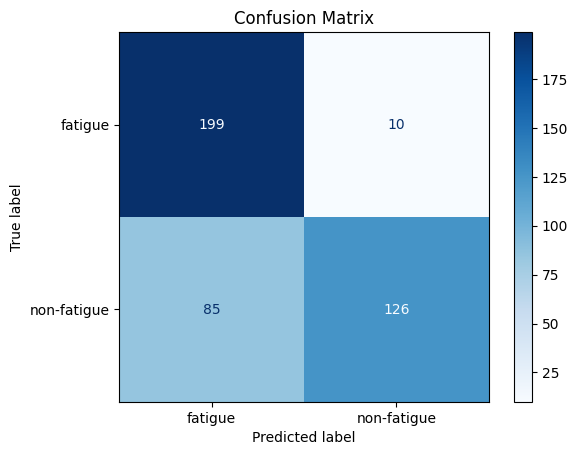} \\[\abovecaptionskip]
    \small (p) User-ID 20
  \end{tabular}
  \begin{tabular}{@{}c@{}}
    \includegraphics[width=.23\linewidth]{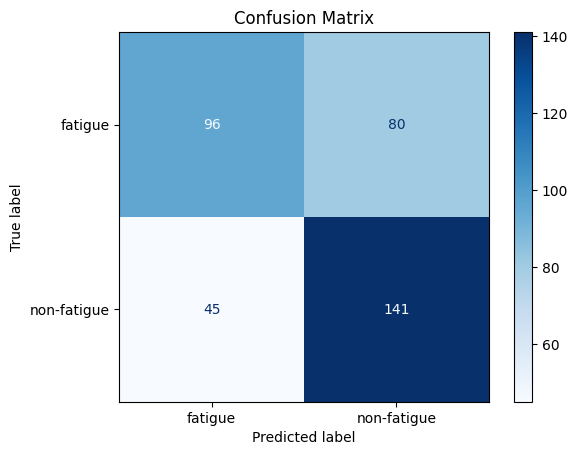} \\[\abovecaptionskip]
    \small (q) User-ID 21
  \end{tabular}
  \begin{tabular}{@{}c@{}}
    \includegraphics[width=.23\linewidth]{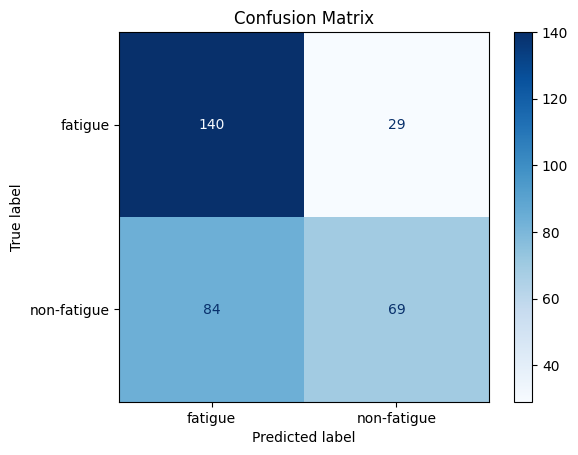} \\[\abovecaptionskip]
    \small (r) User-ID 22
  \end{tabular}
  \begin{tabular}{@{}c@{}}
    \includegraphics[width=.23\linewidth]{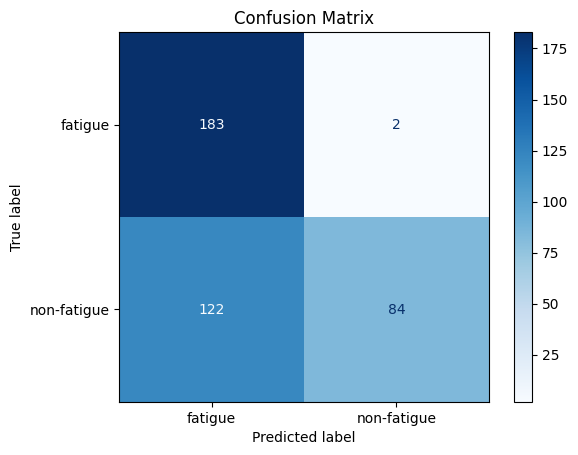} \\[\abovecaptionskip]
    \small (s) User-ID 23
  \end{tabular}
\caption{\textbf{Confusion Matrix for HED-LM with \#ParamA}.}
\label{fig:hedA}
\end{figure}

\begin{figure}[H]
  \centering
  \begin{tabular}{@{}c@{}}
    \includegraphics[width=.23\linewidth]{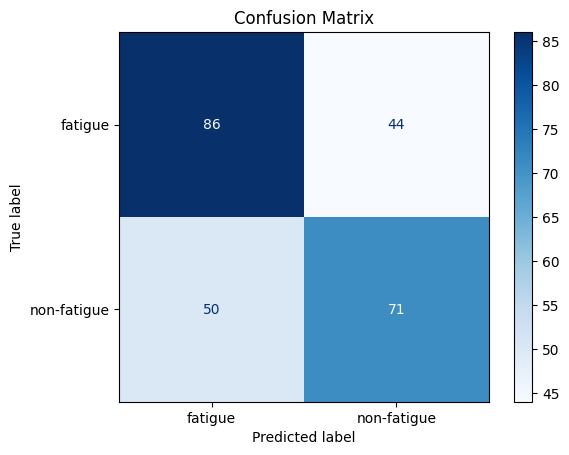} \\[\abovecaptionskip]
    \small (a) User-ID 4
  \end{tabular}
  \begin{tabular}{@{}c@{}}
    \includegraphics[width=.23\linewidth]{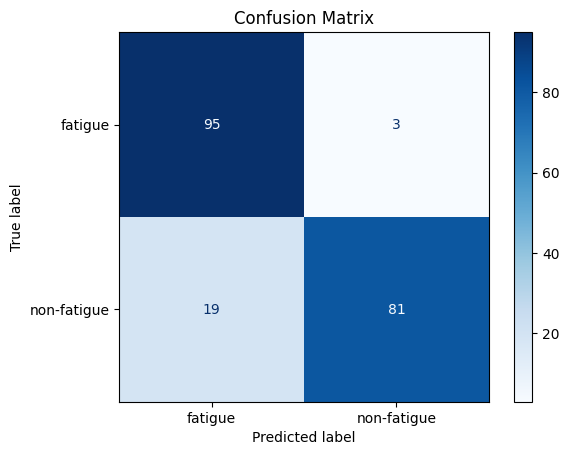} \\[\abovecaptionskip]
    \small (b) User-ID 5
  \end{tabular}
    \vspace{\floatsep}
  \begin{tabular}{@{}c@{}}
    \includegraphics[width=.23\linewidth]{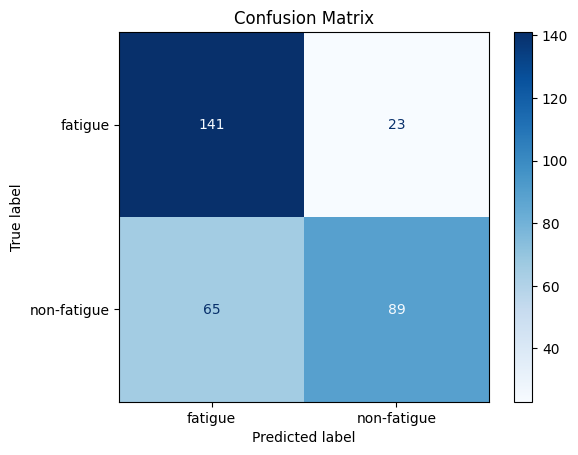} \\[\abovecaptionskip]
    \small (c) User-ID 6
  \end{tabular}
  \begin{tabular}{@{}c@{}}
    \includegraphics[width=.23\linewidth]{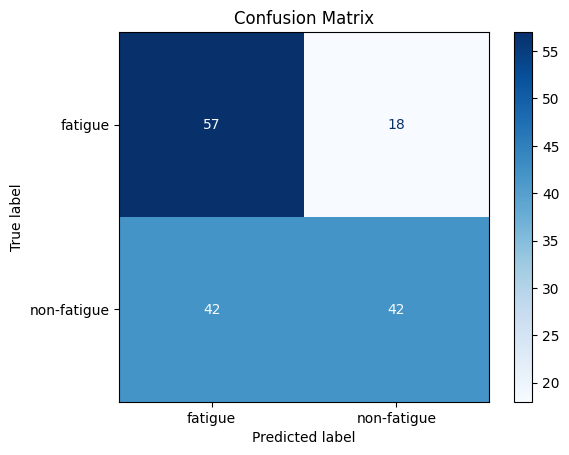} \\[\abovecaptionskip]
    \small (d) User-ID 7
  \end{tabular}
   \begin{tabular}{@{}c@{}}
    \includegraphics[width=.23\linewidth]{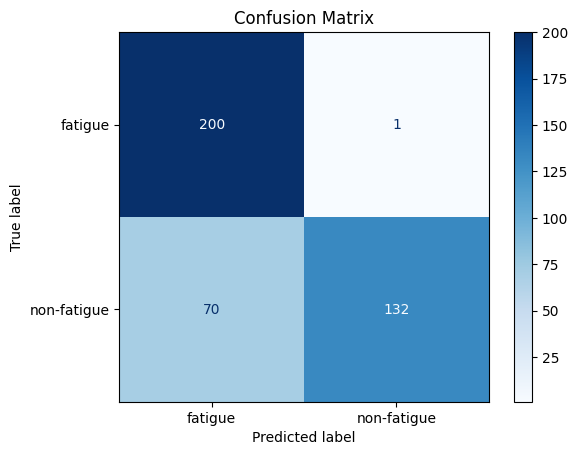} \\[\abovecaptionskip]
    \small (e) User-ID 8
  \end{tabular}
   \begin{tabular}{@{}c@{}}
    \includegraphics[width=.23\linewidth]{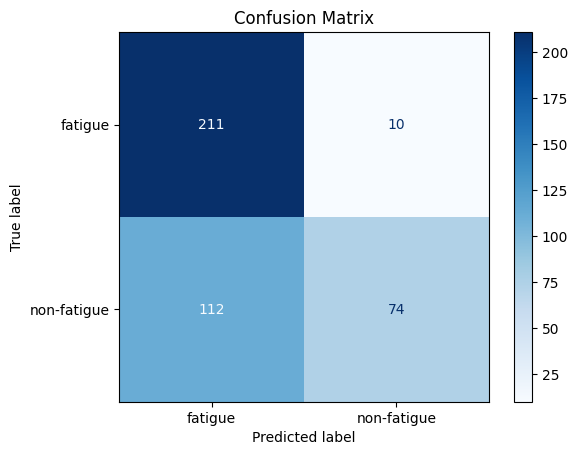} \\[\abovecaptionskip]
    \small (f) User-ID 9
  \end{tabular}
   \begin{tabular}{@{}c@{}}
    \includegraphics[width=.23\linewidth]{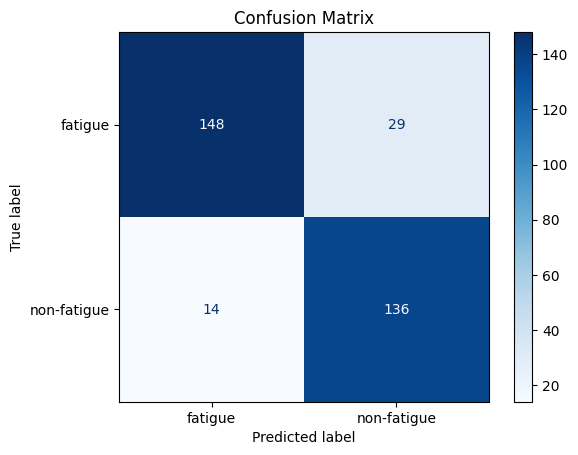} \\[\abovecaptionskip]
    \small (g) User-ID 10
  \end{tabular}
   \begin{tabular}{@{}c@{}}
    \includegraphics[width=.23\linewidth]{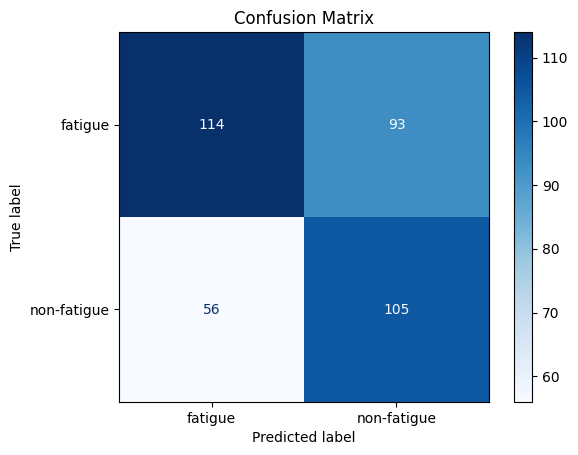} \\[\abovecaptionskip]
    \small (h) User-ID 11
  \end{tabular}
   \begin{tabular}{@{}c@{}}
    \includegraphics[width=.23\linewidth]{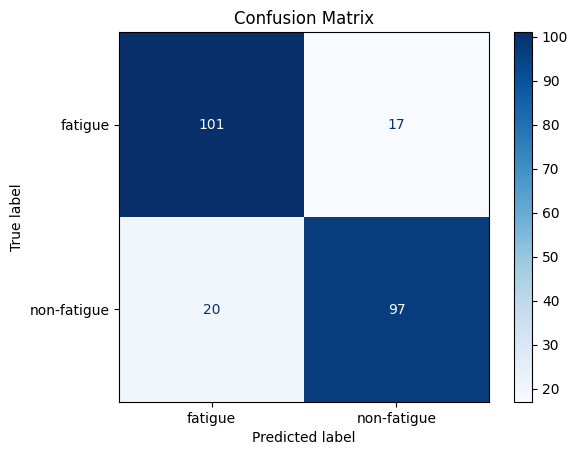} \\[\abovecaptionskip]
    \small (i) User-ID 12
  \end{tabular}
   \begin{tabular}{@{}c@{}}
    \includegraphics[width=.23\linewidth]{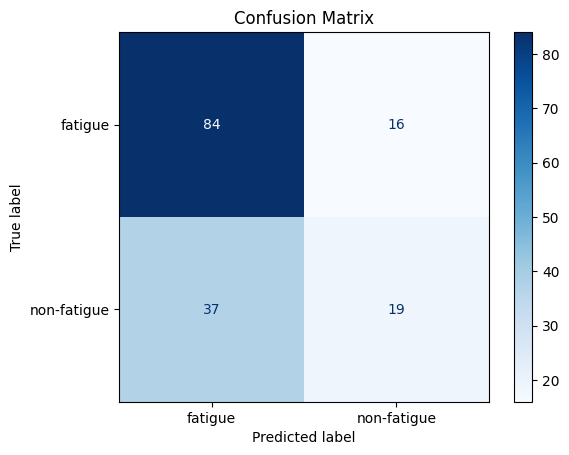} \\[\abovecaptionskip]
    \small (j) User-ID 13
  \end{tabular}
   \begin{tabular}{@{}c@{}}
    \includegraphics[width=.23\linewidth]{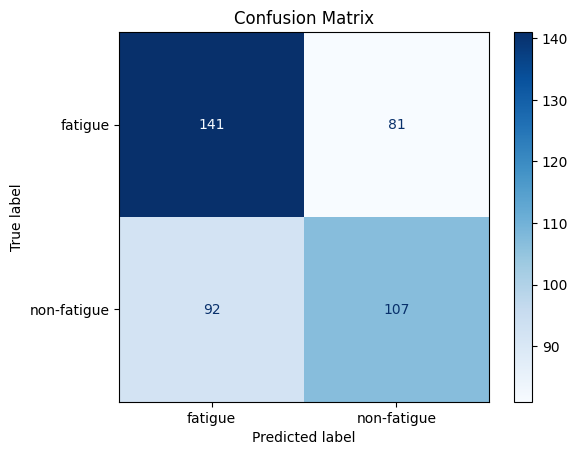} \\[\abovecaptionskip]
    \small (k) User-ID 14
  \end{tabular}
  \begin{tabular}{@{}c@{}}
    \includegraphics[width=.23\linewidth]{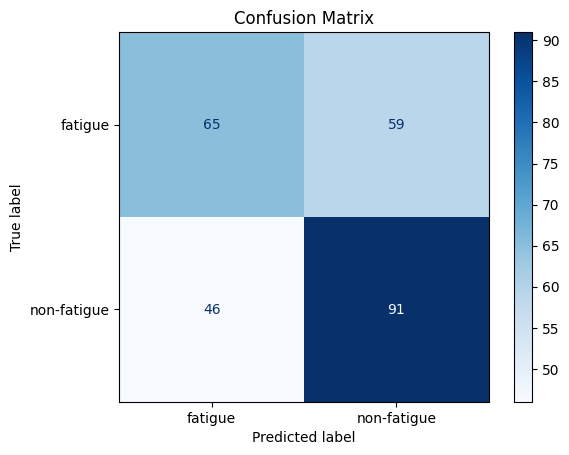} \\[\abovecaptionskip]
    \small (l) User-ID 15
  \end{tabular}
  \begin{tabular}{@{}c@{}}
    \includegraphics[width=.23\linewidth]{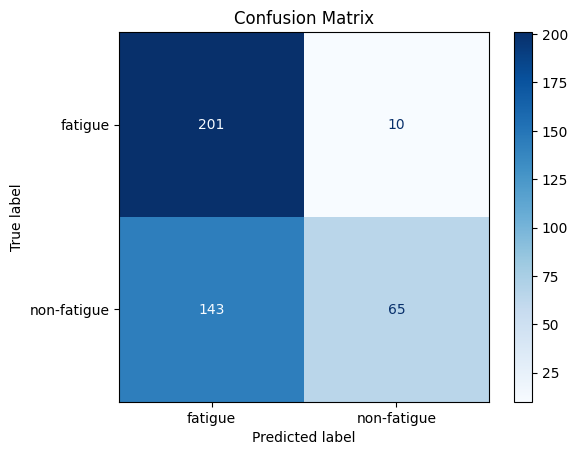} \\[\abovecaptionskip]
    \small (m) User-ID 17
  \end{tabular}
  \begin{tabular}{@{}c@{}}
    \includegraphics[width=.23\linewidth]{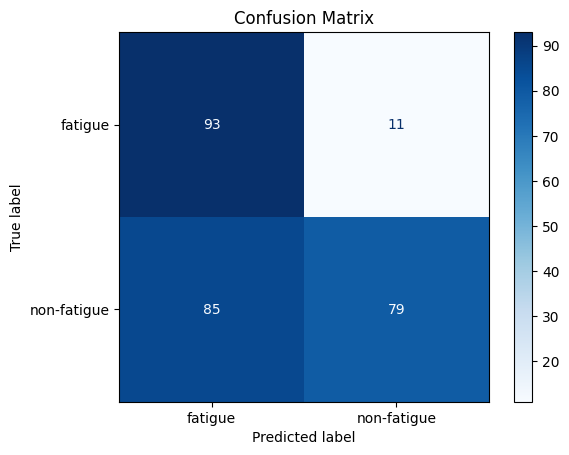} \\[\abovecaptionskip]
    \small (n) User-ID 18
  \end{tabular}
  \begin{tabular}{@{}c@{}}
    \includegraphics[width=.23\linewidth]{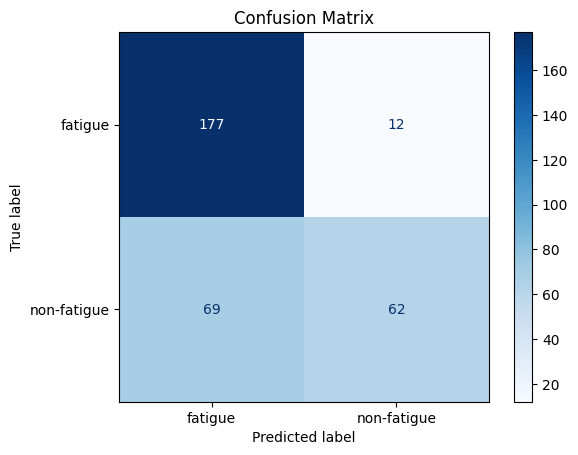} \\[\abovecaptionskip]
    \small (o) User-ID 19
  \end{tabular}
  \begin{tabular}{@{}c@{}}
    \includegraphics[width=.23\linewidth]{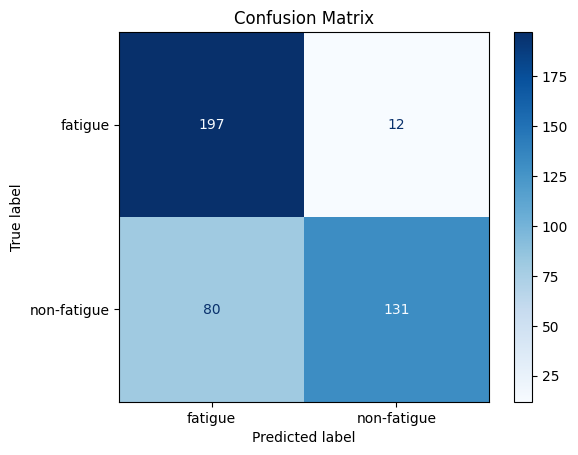} \\[\abovecaptionskip]
    \small (p) User-ID 20
  \end{tabular}
  \begin{tabular}{@{}c@{}}
    \includegraphics[width=.23\linewidth]{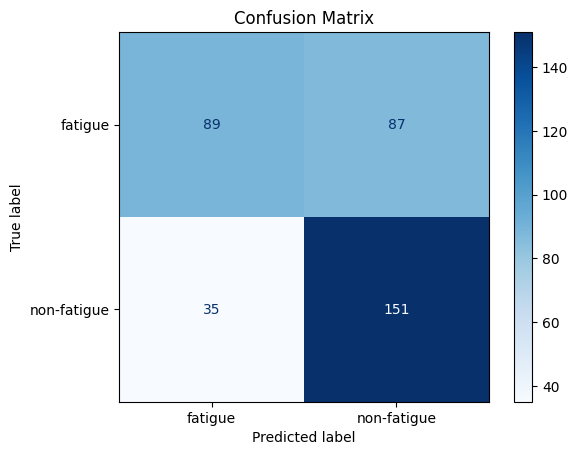} \\[\abovecaptionskip]
    \small (q) User-ID 21
  \end{tabular}
  \begin{tabular}{@{}c@{}}
    \includegraphics[width=.23\linewidth]{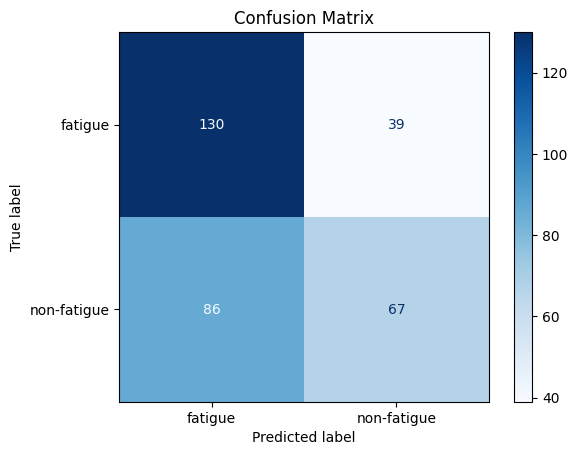} \\[\abovecaptionskip]
    \small (r) User-ID 22
  \end{tabular}
  \begin{tabular}{@{}c@{}}
    \includegraphics[width=.23\linewidth]{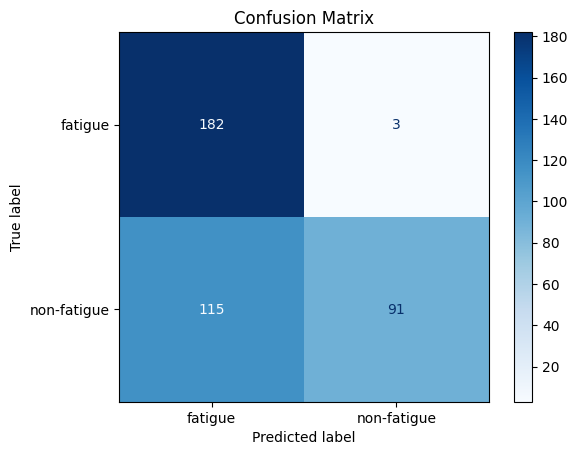} \\[\abovecaptionskip]
    \small (s) User-ID 23
  \end{tabular}
\caption{\textbf{Confusion Matrix for HED-LM with \#ParamB}.}
\label{fig:hedB}
\end{figure}

\begin{figure}[H]
  \centering
  \begin{tabular}{@{}c@{}}
    \includegraphics[width=.495\linewidth]{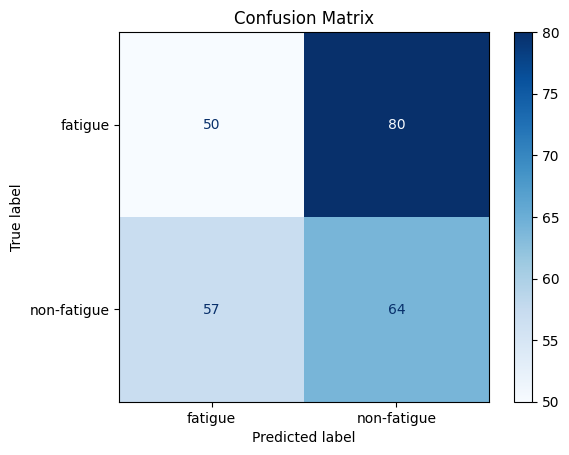} \\[\abovecaptionskip]
    \small (a) Random Approach
  \end{tabular}
  \begin{tabular}{@{}c@{}}
    \includegraphics[width=.495\linewidth]{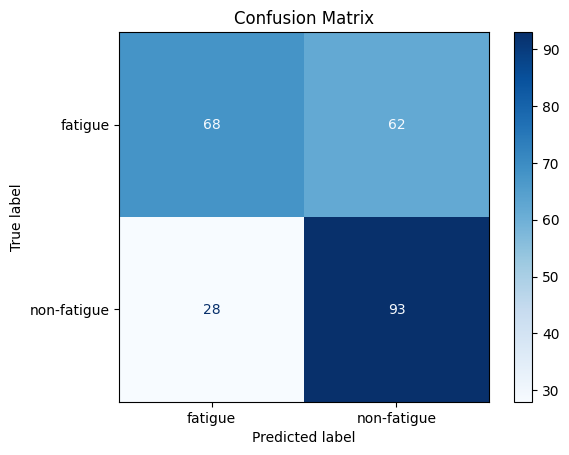} \\[\abovecaptionskip]
    \small (b) Distance Approach
  \end{tabular}
    \vspace{\floatsep}
  \begin{tabular}{@{}c@{}}
    \includegraphics[width=.495\linewidth]{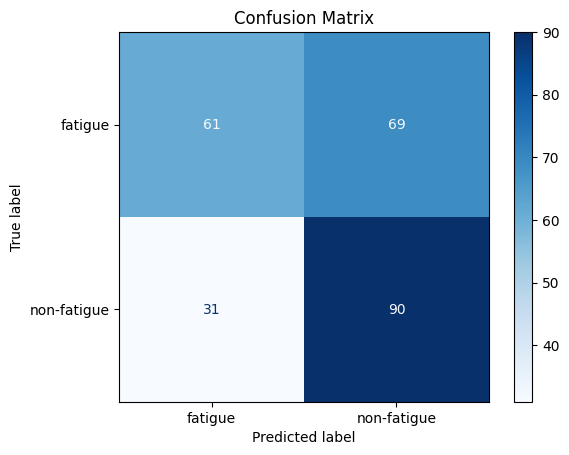} \\[\abovecaptionskip]
    \small (c) HED-LM with \#ParamA
  \end{tabular}
  \begin{tabular}{@{}c@{}}
    \includegraphics[width=.495\linewidth]{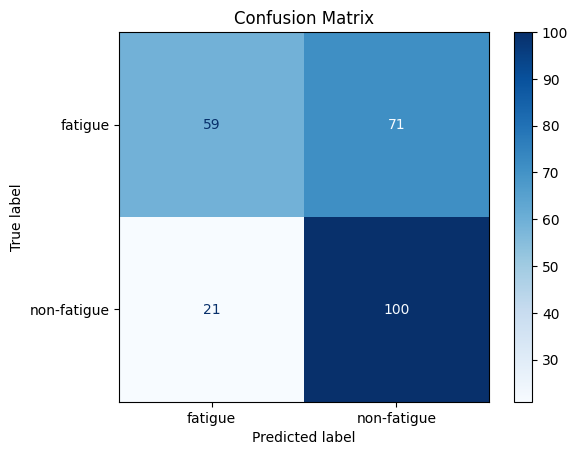} \\[\abovecaptionskip]
    \small (d) HED-LM with \#ParamB
  \end{tabular}
\caption{\textbf{Confusion matrix of User-ID 4 without domain knowledge}.}
\label{fig:wo-dom-4}
\end{figure}

By comparing the confusion matrices, we aim to identify whether our proposed approach effectively addresses the limitations of the baselines. Precisely, we assess improvements in handling minority classes, reducing misclassification errors, and achieving a more balanced distribution of predictions across different classes. Fig.~\ref{fig:random}, \ref{fig:dist}, and \ref{fig:ml} illustrate the confusion matrices for the random, distance-based, and traditional machine learning approaches for individual participants. In contrast, Fig.~
\ref{fig:hedA} and \ref{fig:hedB} highlights the confusion matrices for our two parameterized approaches, evaluated under the same conditions. These comparisons demonstrate the effectiveness of our method in overcoming baseline weaknesses and achieving a more robust and balanced predictive performance. Meanwhile, Fig.~\ref{fig:wo-dom-4} shows the confusion matrix of User-ID 4 performance without domain knowledge.

%%%%%%%%%%%%%%%%%%%%%%%%%%%%%%%%%%%%%%%%%%
\isPreprints{}{% This command is only used for ``preprints''.
%\begin{adjustwidth}{-\extralength}{0cm}
} % If the paper is ``preprints'', please uncomment this parenthesis.
%\printendnotes[custom] % Un-comment to print a list of endnotes

\reftitle{References}

% Please provide either the correct journal abbreviation (e.g. according to the “List of Title Word Abbreviations” http://www.issn.org/services/online-services/access-to-the-ltwa/) or the full name of the journal.
% Citations and References in Supplementary files are permitted provided that they also appear in the reference list here. 

%=====================================
% References, variant A: external bibliography
%=====================================
\bibliography{reference}
\PublishersNote{}
%\isPreprints{}{% This command is only used for ``preprints''.
%\end{adjustwidth}
%} % If the paper is ``preprints'', please uncomment this parenthesis.
\end{document}